\documentclass[final,5p,times]{elsarticle}

\usepackage{amsfonts}
\usepackage[tight,footnotesize]{subfigure}
\usepackage[toc,page]{appendix}
\usepackage{array,multirow}
\usepackage{amssymb}
\usepackage{amsmath}
\usepackage{amsthm}
\usepackage{mathrsfs}
\usepackage{latexsym}
\usepackage{epsf}
\usepackage{psfig}
\usepackage{url}
\usepackage{algorithm}
\usepackage{algpseudocode}
\usepackage{comment}
\usepackage{tabularx}
\usepackage{multirow}
\usepackage{booktabs}
\usepackage{subfig}
\usepackage[font=small]{caption}
\usepackage{etoolbox}
\makeatletter
\patchcmd{\ps@pprintTitle}
  {Preprint submitted to}
  {Preprint accepted by}
  {}{}
\makeatother

\usepackage{color}

\journal{Neural Networks}

\begin{document}

\begin{frontmatter}

\title{Understanding Autoencoders with Information Theoretic Concepts}



\author[label1]{Shujian Yu}
\author[label1]{Jos\'{e} C. Pr\'{i}ncipe\fnref{mark1}}

\fntext[mark1]{Corresponding author. Tel.: +001 (352) 392-2662; email: principe@cnel.ufl.edu.}

\address[label1]{Computational NeuroEngineering Laboratory,\\ Department of Electrical and Computer Engineering, \\ University of Florida, Gainesville, FL 32611, USA}

\begin{abstract}
Despite their great success in practical applications, there is still a lack of theoretical and systematic methods to analyze deep neural networks. In this paper, we illustrate an advanced information theoretic methodology to understand the dynamics of learning and the design of autoencoders, a special type of deep learning architectures that resembles a communication channel. By generalizing the information plane to any cost function, and inspecting the roles and dynamics of different layers using layer-wise information quantities, we emphasize the role that mutual information plays in quantifying learning from data. We further suggest and also experimentally validate, for mean square error training, three fundamental properties regarding the layer-wise flow of information and intrinsic dimensionality of the bottleneck layer, using respectively the data processing inequality and the identification of a bifurcation point in the information plane that is controlled by the given data. Our observations have direct impact on the optimal design of autoencoders, the design of alternative feedforward training methods, and even in the problem of generalization.
\end{abstract}


\end{frontmatter}

\graphicspath{{figures/}}

\section{Introduction}\label{sec:introduction}

Deep neural networks (DNNs) have drawn significant interest from the machine learning community, especially due to their recent empirical success in various applications such as image recognition \cite{krizhevsky2012imagenet}, speech recognition \cite{graves2013speech}, natural language processing \cite{mesnil2013investigation}. Despite the overwhelming advantages achieved by deep neural networks over the classical machine learning models, the theoretical and systematic understanding of deep neural networks still remains limited and unsatisfactory. Consequently, deep models themselves are typically regarded as ``black boxes" \cite{alain2016understanding}.

This is an unfortunate terminology that the second author has disputed since the late $90$'s \cite{principe2000neural}. In fact, most neural architectures are homogeneous in terms of processing elements (PEs), e.g., sigmoid nonlinearities. Therefore no matter if they are used in the first layer, in the middle layer or the output layer they always perform the same function: they create ridge functions~\cite{light1992ridge} in the space spanned by the previous layer outputs, i.e., training will only control the steering of the ridge, while the bias controls the aggregation of the different partitions~\cite{minsky2017perceptrons,principe2015universal}. Moreover, it is also possible to provide geometric interpretations to the projections, extending well known work of Kolmogorov for optimal filtering in linear spaces \cite{kolmogorov1939interpolation}. What has been missing is a framework that can provide an assessment of the quality of the different projections learned during training besides the quantification of the ``external" error.

More recently, there has been a growing interest in understanding deep neural networks using information theory. Information theoretic learning (ITL) \cite{principe2010information} has been successfully applied to various machine learning applications by providing more robust cost or objective functions,
but its role can be extended to create a framework to help optimally design deep learning architectures, as explained in this paper. Recently, Tishby proposed the \emph{Information Plane} (IP) as an alternative to understand the role of learning in deep architectures \cite{shwartz2017opening}. The use of information theoretic ideas is an excellent addition because Information Theory is essentially a theory of bounds \cite{mackay2003information}. Entropy and mutual information quantify properties of data and the results of functional transformations applied to data at a sufficient abstract level that can lead to an optimal performance as illustrated by Stratonovich's three variational problems \cite{stratonovich1965value}. These recent works demonstrate the potential that various information theory concepts hold to open the ``black box" of DNNs.	

As an application we will concentrate on the design of stacked autoencoders (SAE), a fully unsupervised deep architecture. Autoencoders have a remarkable similarity with a transmission channel \cite{yu2017autoencoders} and so they are a good choice to evaluate the appropriateness of using ITL in understanding the architectures and the dynamics of learning in DNNs. We are interested in unveiling the role of the layer-wise mutual information during the autoencoder training phase, and investigating how its dynamics through learning relate to different information theoretic concepts (e.g., different data processing inequalities). We propose to do this for arbitrary topologies using empirical estimators of Renyi's mutual information, as explained in \cite{giraldo2015measures}. Moreover, we are also interested in how to use our observations to benefit the design and implementation of DNNs, such as optimizing a neural network topology or training a neural network in a feedforward greedy-layer manner, as an alternative to the standard backpropagation.

The rest of this paper is organized as follows. In Section \ref{section2}, we briefly introduce background and related works, including a review of the geometric projection view of multilayer systems, elements of Renyi's entropy and their matrix-based functional as well as previous works on understanding DNNs. Following this, we suggest three fundamental properties associated with the layer-wise mutual information and also give our reasoning in Section \ref{section3}. We then carry out experiments on three real-world datasets to validate these properties in section \ref{experiments}. An experimental interpretation is also presented. We conclude this paper and present our insights in Section \ref{conclusions}.

To summarize, our main contributions are threefold:
\begin{itemize}
\item Instead of using the basic Shannon or Renyi's definitions on mutual information which require precise PDF estimation in high-dimensional space, we suggest using the recently proposed matrix-based Renyi's $\alpha$-entropy functional~\cite{giraldo2015measures} to estimate information quantities in DNNs. We demonstrate that this class of estimators can be used in high dimensions ($\sim1,000$) and preserve the theoretical expectations of the Data Processing Inequality.
    The new estimators compute the entropy and mutual information in the reproducing kernel Hilbert spaces (RKHS) and avoid explicit PDF estimation, thus making information flow estimation simple and practical as required to analyze learning dynamics of DNNs.
\item Benefiting from the simple yet precise estimator, we suggest three fundamental properties associated with information flow in SAEs, which provides insights in the dynamics of DNNs from an information theory perspective.
\item The new estimators and our observations also illustrate practical implications on the problem of architecture design/selection, generalizability, and other critical issues in deep learning communities. Moreover, the proposed methodologies can be extended to other DNN architectures much more complex than MLP or SAEs, like the CNNs~\cite{yu2018understanding}.
\end{itemize}

The abbreviations and variables mentioned in this paper are summarized in Table~\ref{Tab:Nomenclature}.

\section{Background and Related works} \label{section2}

In this section, we start with a review of the geometric interpretation of multilayer system mappings as well as the basic autoencoder that provides a geometric underpinning for the IP quantification. The geometric interpretation describes the basic operation of pairwise layer projections, which agrees with the pairwise mutual information in IP, hinting that we are effectively quantifying the role of projections using information theoretic quantities and opening the ``black box". After that, we give a brief introduction to Renyi's entropy and its associated matrix functional defined on the normalized eigenspectrum of the Hermitian matrix of the projected data in reproducing kernel Hilbert spaces (RKHS). Finally, we briefly review related works addressing explainability of DNNs.

\begin{table}[H]
  \small
  \centering
  \caption{Nomenclature.}\label{Tab:Nomenclature}
\begin{tabular}{ccc}\toprule
\multicolumn{2}{c}{Abbreviations}\\ \midrule
  \textbf{DNN} & deep neural network  \\
 \textbf{SAE} & stacked autoencoder \\
\textbf{MLP} & multilayer perceptron \\
\textbf{CNN} & convolutional neural network \\
\textbf{ITL} & information theoretic learning \\
\textbf{IP} & information plane \\
\textbf{PDF} & probability density function \\
\textbf{RKHS} & reproducing kernel Hilbert space \\
\textbf{MIPS} & multidimensional internal projection space \\
\textbf{NPD} & normalized positive definite \\
\textbf{DPI} & data processing inequality \\
\midrule
\multicolumn{2}{c}{Variables} \\\midrule
  S & number of hidden layers in the encoder or decoder \\
  K & bottleneck layer size or number of neurons in the bottleneck layer \\
  D & intrinsic dimensionality of given data $\mathbf{X}$ \\
  M & embedding length of a time series \\
\bottomrule
\end{tabular}
\end{table}

\subsection{A geometric perspective to nonlinear multilayer system projections} \label{section2_1}

Throughout this section, we use lowercase letters (e.g., $y$) to denote scalars, lowercase boldface letters to denote vectors (e.g., $\mathbf{x}$) and uppercase letters to denotes matrices (i.e., $\mathbf{X}$). Given a signal, the geometric analysis of the optimal filtering presented here is going to be done on a window of data of $N$ samples, instead of the conventional analysis of observing just the output of the filter in a subspace of dimension $m$ given by the number of filter parameters. The advantage of this analysis is to illustrate the issues of linear versus nonlinear projections.

\begin{figure}[!htbp]
\centering
\begin{tabular}{ccc}
\includegraphics[width=.45\textwidth]{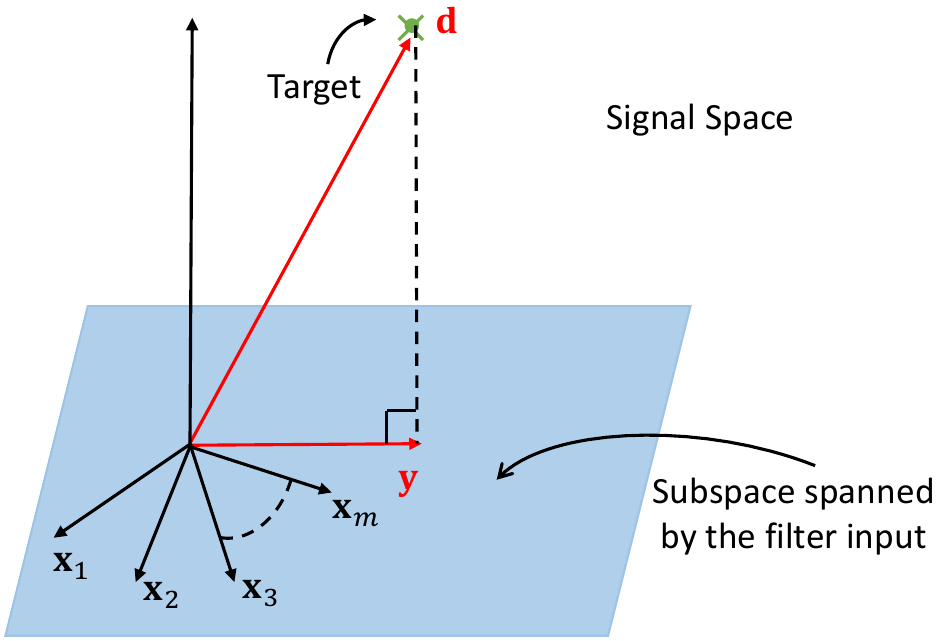}
\end{tabular}
\caption{Illustration of the problem of filtering or regression in the signal space of dimension $N$. The subspace created by the $m$ input signal vectors (assumed horizontal) is used to find the best approximation (the orthogonal projection) of the measurements $\mathbf{d}$ into this space.\vspace{-0.0cm}}
\label{fig:regression_geometry}
\end{figure}

The input signal on a window of $N\gg m$ samples can be considered as a point in a signal space of dimension $N$. Every possible signal of $N$ dimensions creates a $N$-dimensional signal space. Suppose the length of a linear filter is $m$, then we can get $m$, $N$-dimensional vectors in the full signal space. These $m$ vectors (denote $\mathbf{X}=\{\mathbf{x}_{1},\mathbf{x}_{2},...,\mathbf{x}_{m}\}$) differ only in the last $m$ samples~\cite[Chapter~3 \& 6]{haykin2014adaptive}, and define a subspace of dimension $m$ in the large signal space of dimension $N$. Because of the fact that these vectors have $(N-m)$ equal coordinates, we can take this common part of the vectors as the origin for a $m$ dimension subspace. Any point in the subspace can then be converted to a point in the $N$ dimensional space by composition of vectors. If we are just interested in the linear filter output in this subspace, it is obvious that the $N$ dimensional vector of coefficients in this filter can be reduced to an $m$ dimensional vector, by zeroing all the $(N-m)$ components of the weight vector. Denote $\mathbf{y}$ the output of $N$ samples of a linear filter with $m$ inputs, then $\mathbf{y}$ must exist in the subspace because it is a linear combination of $\mathbf{x}_{1},\mathbf{x}_{2},...,\mathbf{x}_{m}$ with filter parameters $\mathbf{w}$~\cite{haykin2014adaptive}. Suppose now that the target signal $\mathbf{d}$ is also a vector in the $N$ dimensional space. Most often the vector $\mathbf{d}$ does not lay in the subspace of dimension $m$ described above, so the problem of filtering or regression is to find the best projection in this $m$ dimensional hyperplane. From Legendre and Gauss~\cite{stigler1981gauss} we know that the optimal solution $\mathbf{w}^*=\mathbf{R}_{xx}^{-1}\mathbf{p}_{xd}$ minimizes the error power (the norm of the error vector) by approximating the cloud of measurements $\mathbf{d}$ that passes through the data, where $\mathbf{R}_{xx}=\mathbf{X}^T\mathbf{X}$ is the autocorrelation matrix and $\mathbf{p}_{xd}=\mathbf{X}^T\mathbf{d}$ is the crosscorrelation vector between $\mathbf{X}$ and $\mathbf{d}$. Geometrically this corresponds to finding the orthogonal projection of $\mathbf{d}$ into the space spanned by the inputs (see Fig. \ref{fig:regression_geometry}).

The problem with the linear solution is that the output must exist in the hyperplane spanned by the inputs, and when this is not the case (which is the norm), the optimal solution can provide an error that may be still too large to make the solution practical. This is the reason why we commonly use nonlinear mapping functions, which are not restricted anymore to provide outputs in the span of the input space, and can therefore provide smaller approximation errors. In \cite[Chapter~5 \& 10]{principe2000neural} and more recently in \cite{principe2015universal} we show that Kolmogorov interpretation can provide insights on the understanding the inner working of any multilayer perceptron (MLP), and we repeat it here for completeness.

Let $y=f(\mathbf{x})$ be a continuous function from $\mathbb{R}^m$ to $\mathbb{R}$. The goal is to approximate $y$ by a function $\hat{f}(x)$ that is built in the following way:
\begin{equation}
\hat{f}(\mathbf{x})=\phi(\sum_i\mathbf{w}_i\phi(\sum_j\mathbf{W}_{i,j}\mathbf{x}_j+\mathbf{b}_i)+b), \label{eq1}
\end{equation}
where $\phi(\cdot)$ are smooth nonlinear functions, $\mathbf{w}$ and $\mathbf{W}$ are weights, $b$ and $\mathbf{b}$ denote bias, $j$ is an index over the input dimension and $i$ is the index over the number of processing elements (PEs) $\phi$ or neurons. The number of layers in (\ref{eq1}) can be expanded and give rise to deep networks that have emerged as the big topic in neural networks. But from a function approximation perspective the single hidden layer is quite adequate as the basic topology to understand deep architectures. For simplicity, we are going to drop the external nonlinearity, yielding:
\begin{equation}
\hat{f}(\mathbf{x})=\sum_i\mathbf{w}_i\phi(\sum_j\mathbf{W}_{i,j}\mathbf{x}_j+\mathbf{b}_i)+b. \label{eq2}
\end{equation}

Let us denote $\phi_i$ as:
\begin{equation}
\phi_i(\mathbf{x})=\phi(\sum_j\mathbf{W}_{i,j}\mathbf{x}_j+\mathbf{b}_i), \label{eq3}
\end{equation}
and substituting in (\ref{eq2}) we obtain:
\begin{equation}
\hat{f}(\mathbf{x})=\sum_i\mathbf{w}_i\phi_i(\mathbf{x})+b. \label{eq4}
\end{equation}

If the same geometric interpretation of filtering or regression is used here, we see that the output of the one hidden layer machine is nothing but \emph{a projection on the space created by the bases ($\phi_i$) of the hidden layer PEs (the multidimensional internal projection space or MIPS)}. The only problem is that MIPS bases are controlled by the input data as well as by the parameters $\mathbf{W}_{i,j}$ as shown in (\ref{eq3}), so they change during learning. Moreover, because of the nonlinear PEs, the space spanned by these bases is no longer limited to the span of the $m$ input signal vectors. The MIPS can be placed anywhere to fulfill the approximation to the target function, depending on the first layer weights $\mathbf{W}_{i,j}$, which is exactly the reason why the one hidden layer machine is an universal approximator.

\begin{figure}[!htbp]
\centering
\begin{tabular}{ccc}
\subfigure[] {\includegraphics[width=.45\textwidth]{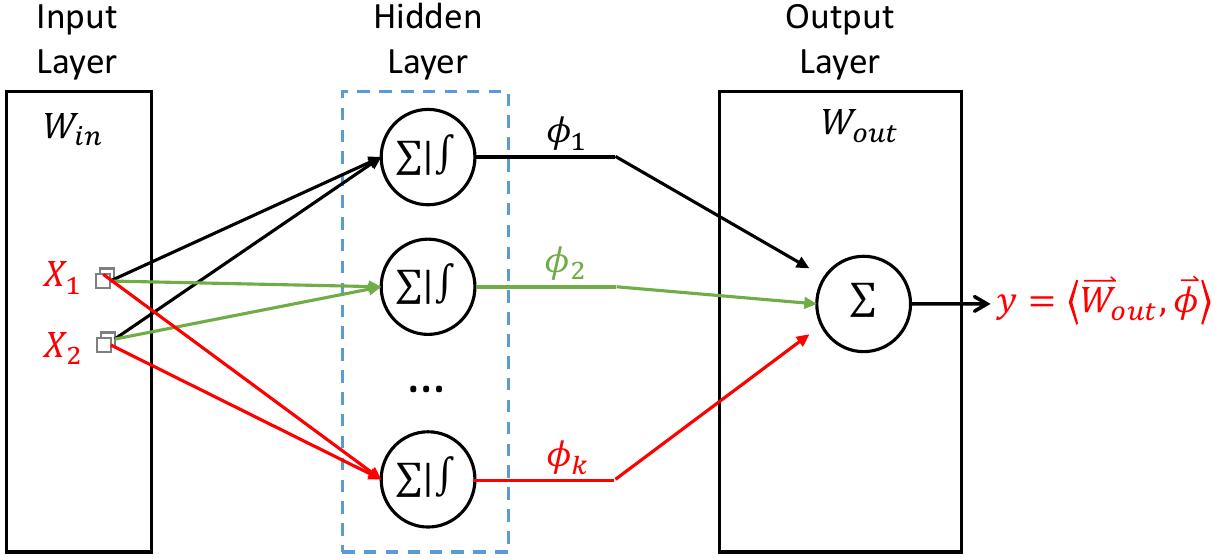}} \\
\subfigure[] {\includegraphics[width=.45\textwidth]{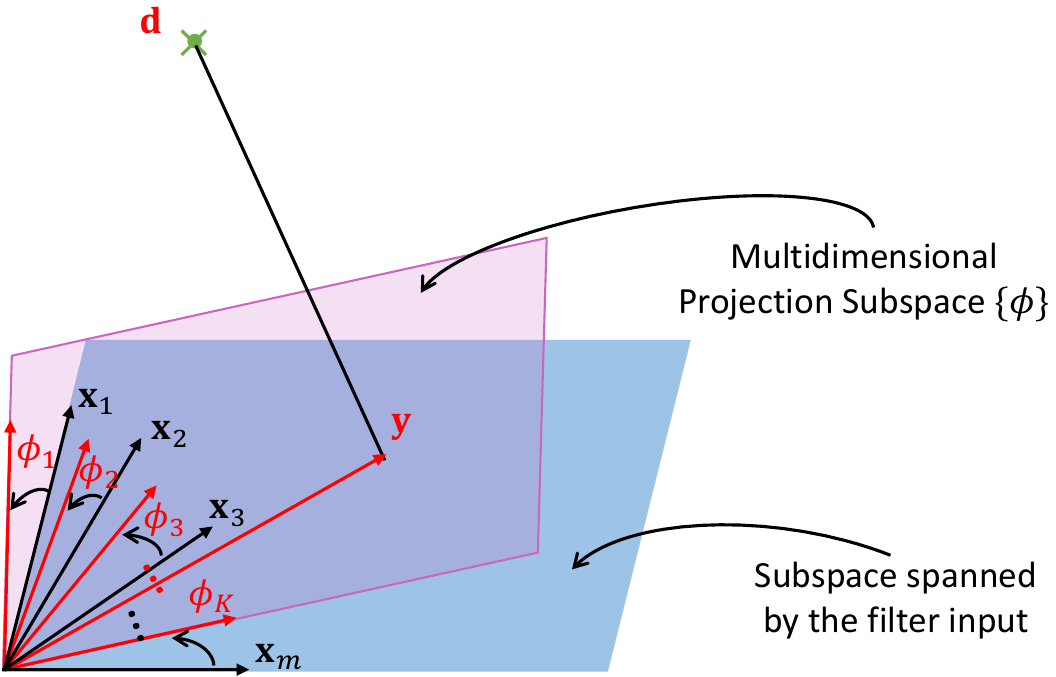}}
\end{tabular}
\caption{(a) shows a single hidden layer MLP with $k$ PEs in the hidden layer and (b) shows the interpretation of the hidden layer PE outputs as the projection space where the optimal solution of the input-output map is obtained.\vspace{-0.0cm}}
\label{fig:geometry}
\end{figure}

Since the optimization problem remains the same, we can now understand better the role of each one of the layers of our learning machine (Fig. \ref{fig:geometry}): the output weights are still finding the orthogonal projection on the MIPS subspace spanned by the $\phi_i$, and this optimization is convex in the parameters provided the output PE is linear. Moreover, the MIPS is no longer the input space and it is dynamically changing during learning, because the bases are themselves a function of the weights of the first layer parameters, which change during training. This also shows that in the beginning of training, MIPS coincides with the input data space (when the weights are started with small random values that put the sigmoid in the linear region), but progressively the mappings become much more dependent upon the goal of the processing dictated by the desired response. In a deep layer network, this perspective of using pairs of layers to understand the internal mechanism of finding representations remains essentially the same (the nonlinearity at the output becomes part of the next pair of layers). Understanding this mechanism also saves precious adaptation time, because it is obvious that the optimal projection on the top layer can only be determined when the previous MIPS stabilize, which calls for different learning rates in each layer. This is textbook material \cite{principe2000neural} that was not fully assimilated by practitioners and non-practitioners alike, who keep declaring that MLPs are black boxes, which they are not! The difficult part is to predict the effect of the bias in the inner layers, which have the ability to aggregate subsets of previously created partitions, as required for the overall mapping. However, this geometric picture still tells us little about how information flows inside the network as its parameters are being adapted; it just tells us how the mappings are implemented.

To summarize, it is obvious that the pair-wise interactions among the variables in the original data space, the MIPS and the output space play significant roles in understanding learning (or mapping) systems that include but are not limited to DNNs. From this perspective, it makes sense to infer the system properties by inspecting the interactions between input and hidden representations and the mutual information between hidden representation and output. A straightforward example comes from the domain of system identification based on mutual information criterion \cite[Chapter~6]{chen2013system}, in which either the minimum mutual information criterion or the maximum mutual information criterion consistently optimize such interaction between variables in two interesting data spaces. However, the motivation of this paper is not limited to the mutual information cost functions as is the norm in neural networks training. We believe the above analysis provides a geometric perspective for the changes in representations during learning, which can be further quantified by the information flow provided by IP in \cite{shwartz2017opening}.

\subsection{Autoencoder and its geometric perspective}

This section gives a brief introduction to the basic architecture of the autoencoder, also from a geometric perspective. The autoencoder is a special type of MLP that aims to transform inputs into outputs with the least possible amount of distortion. It consists of two modules: a feedforward encoder module that maps the input to a code vector or hidden representation $\mathbf{z}$ in the bottleneck layer and a decoder module that tends to reconstruct the input sample from $\mathbf{z}$. In this sense, the autoencoder is a supervised version of a clustering algorithm in a projected space, where the bottleneck layer PEs implement global projections.

\begin{figure}[!htbp]
\centering
\begin{tabular}{ccc}
\subfigure[] {\includegraphics[width=.45\textwidth]{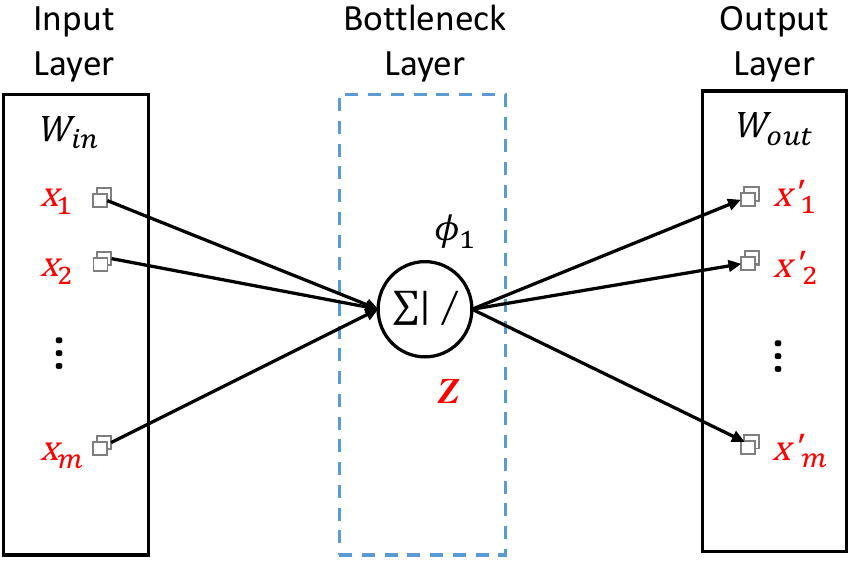}} \\
\subfigure[] {\includegraphics[width=.45\textwidth]{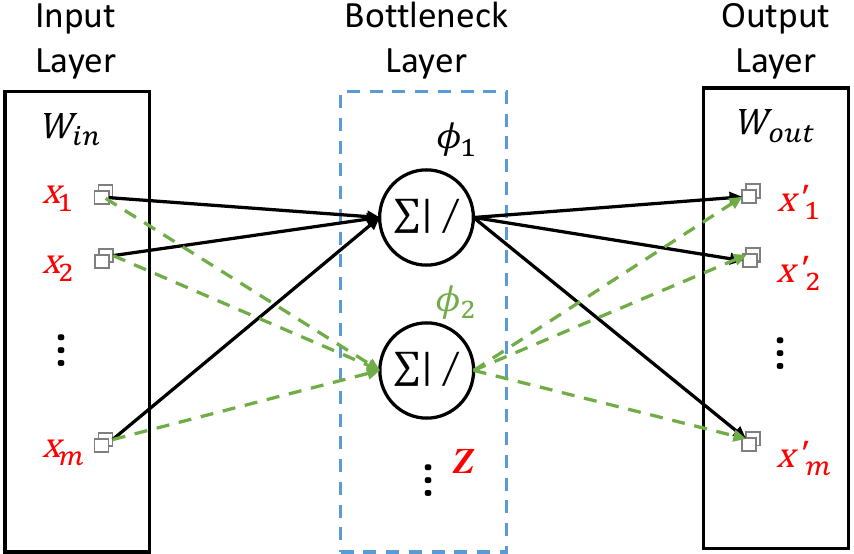}}
\end{tabular}
\caption{(a) shows a standard autoencoder with one PE. A sample $\mathbf{x}$ is compressed to one component $\mathbf{z}$ in the bottleneck layer by the encoder. The decoder reconstructs $\mathbf{x}'$ from $\mathbf{z}$. The sample $\mathbf{x}'$ is usually a noise-reduced representation of $\mathbf{x}$. The network can be extended to extract more than one component by additional PEs in the bottleneck layer as shown in (b). The code $\mathbf{z}$ lies in column space spanned by the input batch if the PEs in the bottleneck layer is linear.\vspace{-0.0cm}}
\label{fig:AE_architecture}
\end{figure}

Specifically, we are given a (mini) batch of samples in matrix $\mathbf{X}\in \mathbb{R}^{N\times m}$, where each row is an input vector. The output of autoencoder (i.e., $\mathbf{X}'$) is enforced to equal to $\mathbf{X}$ with high fidelity by minimizing the squared reconstruction error $\|\mathbf{X}-\mathbf{X}'\|^2$. For simplicity, we assume there is only one linear PE in the bottleneck layer and the weights $\mathbf{w}$ are symmetric in the encoder and decoder (see Fig. \ref{fig:AE_architecture}(a)). Over the batch, the encoder does $\mathbf{X}\mathbf{w}$ and the obtained vector lies in the column space spanned by $\mathbf{X}$ as emphasized in Section \ref{section2_1}. Because of symmetry, the decoder does $\mathbf{X}\mathbf{w}\mathbf{w}^T$. Then the objective is to minimize $\mathrm{tr}((\mathbf{X}-\mathbf{X}\mathbf{w}\mathbf{w}^T)(\mathbf{X}-\mathbf{X}\mathbf{w}\mathbf{w}^T)^T)$, where $\mathrm{tr}$ denotes trace. This is exactly the Principal Component Analysis (PCA) decomposition \cite{kokiopoulou2011trace}, and the optimal solution is given by $\mathbf{w}\propto\mathbf{e}$, the top eigenvector of the matrix $\mathbf{X}^T\mathbf{X}$ \cite{baldi1989neural}.


Similar to the regression case in Section \ref{section2_1}, we end up in the span of the input batch, defined by the column space of $\mathbf{X}$.
Of course, $k$ PEs will lead to $\mathbf{W}\rightarrow \mathbf{E}_k$, the matrix of top $k$ eigenvectors of $\mathbf{X}^T\mathbf{X}$ (see Fig. \ref{fig:AE_architecture}(b)). Actually, if we have multiple hidden layers with nonlinear PEs in both encoder and decoder, this interpretation still holds in the bottleneck (or innermost) layer. The only difference is that the eigenvectors are now embedded in the span of nonlinear MIPS, rather than that of the input batch.

\subsection{Elements of Renyi's entropy and their matrix-based functional} \label{section2_2}
In information theory, a natural extension of the well-known Shannon's entropy is Renyi's $\alpha$-entropy \cite{renyi1961measures}. For a random variable $X$ with probability density function (PDF) $f(x)$ in a finite set $\mathcal{X}$, the $\alpha$-entropy $\mathbf{H}_\alpha(X)$ is defined as:
\begin{equation}
\mathbf{H}_{\alpha}(f)=\frac{1}{1-\alpha}\log\int_\mathcal{X} dxf^\alpha(x).\label{eq5}
\end{equation}

For $\alpha=1$, (\ref{eq5}) is defined in the limit $\mathbf{H}_1(f)=\lim_{\alpha\rightarrow 1} \mathbf{H}_{\alpha}(f)$, which reduces to the Shannon (differential) entropy. It also turns out that for any real $\alpha$, the above quantity can be expressed, as function of inner products between PDFs. In particular, the $2$-order (or quadratic) entropy can be expressed as:
\begin{equation}
\mathbf{H}_2(f)=-\log\int_\mathcal{X} dxf^2(x).\label{eq6}
\end{equation}

In order to apply this expression to any PDF, ITL \cite{principe2010information} uses Parzen-window density estimation with Gaussian kernel $G_\sigma(\cdot)=\frac{1}{\sqrt{2\pi}\sigma}\exp(\frac{\|\cdot\|^2}{2\sigma^2})$ to estimate $\alpha$-norm PDF directly from data. More specifically, the estimator of Renyi's quadratic entropy is given by:
\begin{equation}
\hat{\mathbf{H}}_2(X)=-\log\big(\frac{1}{N^2}\sum_{i=1}^N\sum_{j=1}^NG_{\sigma\sqrt{2}}(x_i-x_j)\big), \label{eq7}
\end{equation}
where $X=\{x_1,x_2,...,x_N\}$ represents a realization of $f$. Note that this estimator has a free parameter (the kernel size $\sigma$) and that its scalability is constrained by its origin on kernel density estimation \cite{silverman1986density}.

Recently, a novel non-parametric estimator for the matrix based Renyi's entropy was developed under the ITL framework \cite{giraldo2015measures}. The new estimator is a smooth matrix functional on the manifold of the normalized positive definite (NPD) matrices over the real numbers, and has been shown to be effective in autoencoders \cite{giraldo2013rate}, MLPs \cite{huang2016flow} and dimensionality reduction \cite{alvarez2017kernel}.

Let $\kappa:\mathcal{X}\times\mathcal{X}\mapsto\mathbb{R}$ be a real valued positive definite kernel that is also infinitely divisible \cite{bhatia2006infinitely}. Given $X=\{x_1,x_2,...,x_N\}$, the Gram matrix $K$ obtained from evaluating a positive definite kernel $\kappa$ on all pairs of exemplars, that is $(K)_{ij}=\kappa(x_i,x_j)$, can be employed to define a quantity with properties similar to those of an entropy functional, for which the PDF of $X$ does not need to be estimated.

More specifically, a matrix-based analogue to Renyi's $\alpha$-entropy for a NPD matrix $A$ of size $N\times N$,  such that $\mathrm{tr}(A)=1$, can be given by the functional:
\begin{equation}
\mathbf{S}_\alpha(A)=\frac{1}{1-\alpha}\log_2\big[\sum_{i=1}^N\lambda_i(A)^\alpha\big], \label{eq8}
\end{equation}
where $\lambda_i(A)$ denotes the $i$-th eigenvalue of $A$, a normalized version of $K$:
\begin{equation}
A_{ij}=\frac{1}{N}\frac{K_{ij}}{\sqrt{K_{ii}K_{jj}}}. \label{eq9}
\end{equation}

Furthermore, based on the product kernel, the joint-entropy can be defined as:
\begin{equation}
\mathbf{S}_\alpha(A,B)=\mathbf{S}_\alpha\big(\frac{A\circ B}{\mathrm{tr}(A\circ B)}\big), \label{eq10}
\end{equation}
where $A\circ B$ denotes the Hadamard product between the matrices $A$ and $B$. It can be shown that if the Gram matrices $A$ and $B$ are constructed using normalized infinitely divisible kernels (based on (\ref{eq9})), such that $A_{ii}=B_{ii}=1/n$, (\ref{eq10}) is never larger than the sum of the individual entropies $\mathbf{S}_\alpha(A)$ and $\mathbf{S}_\alpha(B)$. This allows us to define the matrix notion of Renyi's mutual information:
\begin{equation}
\mathbf{I}_\alpha(A;B)=\mathbf{S}_\alpha(A)+\mathbf{S}_\alpha(B)-\mathbf{S}_\alpha(A,B). \label{eq11}
\end{equation}

As we are going to see this definition allows us to interpret mappings created by arbitrary cost functions which are the norm in deep learning. They are also readily applicable to any dataset because both the ITL estimator of Renyi's entropy and mutual information and their NPD matrix extensions can be directly applied to data. However, there is again a free parameter (the kernel size) that needs to be cross-validated or carefully tuned for the estimation \cite{silverman1986density}, and the computation complexity is high because of the intrinsic eigendecomposition.

\subsection{Previous approaches}

Current works on understanding DNNs typically fall into three categories. The first category intends to explain the mechanism of DNNs by building a strong connection with the widely acknowledged concepts or theorems from other disciplines. The authors in \cite{mehta2014exact} showed that there is an exact one-to-one mapping between the variational renormalization group (RG) in theoretical physics and stacked restricted Boltzmann machines (RBM), which suggests that stacked RBM iteratively integrate out irrelevant features in the bottom layer while retaining the most relevant ones in the upper layer. This argument was later questioned by \cite{lin2017does}, in which the authors claimed an extraordinary link between DNNs and the nature of the universe. Therefore, the essence of DNNs seems to be buried in the laws of physics. On the other hand, the authors in \cite{tishby2015deep} proposed to formulate the learning of a DNN as a trade-off between compression and prediction, i.e., the DNN learning problem can be formulated under the information bottleneck (IB) framework \cite{tishby2000information} that attempts to extract the minimal sufficient statistics of input data with respect to the target. The authors in \cite{khadivi2016flow} investigated the flow of the discrete Shannon entropy across consecutive layers in a MLP and defined a new optimization problem for training a MLP based on the IB principle. Moreover, they demonstrated numerically that a MLP can successfully learn Boolean functions (AND, OR, XOR) while achieving the minimal representation of the data. A similar work is shown in \cite{huang2016flow}, where the training of MLP is formulated with the rate-distortion function.
As a parallel work, the authors in~\cite{burgess2018understanding} interpreted the emergence of disentangled representation in $\beta$-variational autoencoders ($\beta$-VAE)~\cite{higgins2016beta} from a rate-distortion perspective.
Interestingly, some recent work demonstrated that SAEs have remarkable similarity with communication channels, thus holding the potential to lead to alternative communication system designs \cite{yu2017autoencoders,o2017introduction}.

Following \cite{tishby2015deep}, a DNN should be analyzed by measuring the information quantities that each layer's output preserves about the input with respect to the target. A new terminology of the IP\footnote{The plane of information quantities that each hidden layer $T$ preserves about the input $X$ with respect to the target $Y$, i.e., $\mathbf{I}(X;T)$ with respect to $\mathbf{I}(T;Y)$, where $\mathbf{I}$ denotes mutual information.} framework is defined thereafter in \cite{shwartz2017opening}. This paper empirically shows that the common stochastic gradient descent (SGD) optimization undergoes two separate phases: an early ``drift" phase, in which the variance of the weights' gradients is much smaller than the means of the gradients; and a later ``diffusion" phase, in which there is a rapid reversal such that the variance of the weights' gradients becomes greater than the means of the gradients. In spite of imposing a constraint not widely used in machine learning, i.e. the cost function is not necessarily an information quantity, \cite{shwartz2017opening} conjectured that each layer's inputs and outputs follow the IB framework. These results, along with explanations for the importance of network depth and the information bottleneck optimality of the layers, made \cite{shwartz2017opening} a very promising avenue to improve the understanding of DNNs. However, the results so far have not been extended to real-world scenarios involving large networks and complex datasets, which were later questioned by some counter-examples in~\cite{saxe2018on}.

On the other hand, the approaches in the second category concentrate more on the analysis of deep feature representations from a geometric perspective. The projection space perspective can benefit from, and be easily integrated with, the IP framework. In fact, $\mathbf{I}(X,T)$ quantifies the mutual information between the cloud of samples in the input space and the corresponding projected cloud of points in the MIPS, which is a very efficient way of quantifying the MIPS rotation through learning. Likewise, $\mathbf{I}(Y,T)$ measures the mutual information between the codes in the MIPS and the cloud of points formed by the desired response. But there are more examples such as \cite{bengio2013better}, in which the authors conjectured that deep layers can help extract the underlying factors of variations that define the structure of the data geometry. This hypothesis was experimentally validated by Brahma $et$ $al$.~\cite{brahma2016deep} by quantitatively defining several manifold measures. Other examples include \cite{pascanu2013number} and \cite{montufar2014number}, in which the authors demonstrated that the layer-wise composition of functions in DNNs are able to separate the input data space into exponentially more linear response regions than their shallow counterparts, thus increasing the power of computing complex and structure data.
Different from the early work, the authors of~\cite{achille2017emergence} suggested using the information stored in the weights, rather than activations or layer outputs, to understand the network optimization and representations. According to them, networks with low information in the weights realize invariant and disentangled representations. Therefore, invariance and disentanglement emerge naturally when training a network with implicit (e.g., SGD) or explicit (e.g., IB Lagrangian~\cite{tishby2000information}) regularization.
Although these results are very promising, we will show, in the later portion of this paper that there should be a limit on the number of layers, because, the deeper the neural network, the more information about the input is lost.

Different from the above two categories, the approaches in the third category attempt to directly visualize what makes DNNs arrive at a particular
classification or recognition decision. For examples, the authors in \cite{zeiler2014visualizing} use deconvolutional network (deconvnet) \cite{zeiler2011adaptive} to visualize features in higher layers of convolutional neural networks (CNNs), whereas the authors in \cite{mahendran2015understanding} suggested understanding CNN features by inverting them to measure how much information is retained in these features from a image reconstruction perspective. Other related works include \cite{yosinski2015understanding}, \cite{nguyen2016multifaceted}, etc., and the trend is to explore the hidden mechanism of different layers using an explanatory graph \cite{zhang2017interpreting}. However, these methods are typically only applicable for CNNs and fail to unveil the intrinsic properties of DNNs in the training phase. As another parallel line, several approaches have been proposed enabling one to understand and interpret the reasoning embodied in a DNN for a single test image~\cite{bach2015pixel,montavon2017explaining,samek2017evaluating}. These methods quantify the ``importance" of individual pixels with respect to the classification decision and allow a visualization in terms of a heatmap in pixel/input space.

In our perspective, DNNs are definitely not ``black boxes" as illustrated in their geometric interpretation extended with the significance of layer-wise mutual information. However, the usefulness of the IP framework in machine learning requires further analysis to relate how the processing of information through nonlinearities can achieve the task goals, and help properly design hyper-parameters of the mapper and the learning process. Along these lines, we suggest and verify three fundamental properties associated with different layer-wise mutual information, including the data processing inequality and two novel and related IPs effectively extending the IP to any pairwise layers to further understand the learning process. Moreover, it is worth noting that, our idea is motivated from a geometric interpretation, rather than strictly by the IB principle \cite{tishby2000information}, i.e. it is not necessary to be interested in mutual information cost functions as is the norm in neural network training, so we believe this motivation complements \cite{shwartz2017opening}.

\section{Understanding Autoencoders with Information Theoretic Concepts} \label{section3}

\subsection{The Data Processing Inequality (DPI) and its extensions to stacked autoencoders (SAEs)} \label{section3.1}
Before systematically interpreting SAEs operation and learning using information theoretic concepts, let us recall the basic learning mechanism (i.e., backpropagation) in any feedforward DNNs (including SAEs, MLP, etc.), the input signals are propagated from input layer to the output layer and the errors are back-propagated in the reverse direction from the output layer to the input layer through the adjoint or dual network~\cite{kuroe1993learning} \cite[Chapter~7]{krawczak2013multilayer}. Both propagations are unidirectional and only dependent upon the previous variables, hence obeying the Markov assumption and thus forming a Markov chain \cite{shwartz2017opening,luttrell1994bayesian,yu2018understanding}. Therefore, there exists two fundamental Data Processing Inequalities (DPIs) in any feedforward DNNs with $L$ hidden layers, i.e., $\mathbf{I}(X;R_1)\geq\mathbf{I}(X;R_2)\geq\dots\geq\mathbf{I}(X;R_L)$ and $\mathbf{I}(\delta_L;\delta_{L-1})\geq\mathbf{I}(\delta_L;\delta_{L-2})\geq\dots\geq\mathbf{I}(\delta_L;\delta_{1})$, where $R_1$, $R_2$, ..., $R_L$ are successive hidden layer representations from the first hidden layer to the output layer and $\delta_L$, $\delta_{L-1}$, ..., $\delta_1$ are errors from the output layer to the first hidden layer.

\begin{figure}[!htbp]
\centering
\begin{tabular}{ccc}
\subfigure[The architecture of SAE with ($S$-1) hidden layers in both encoder and decoder.] {\includegraphics[width=.45\textwidth]{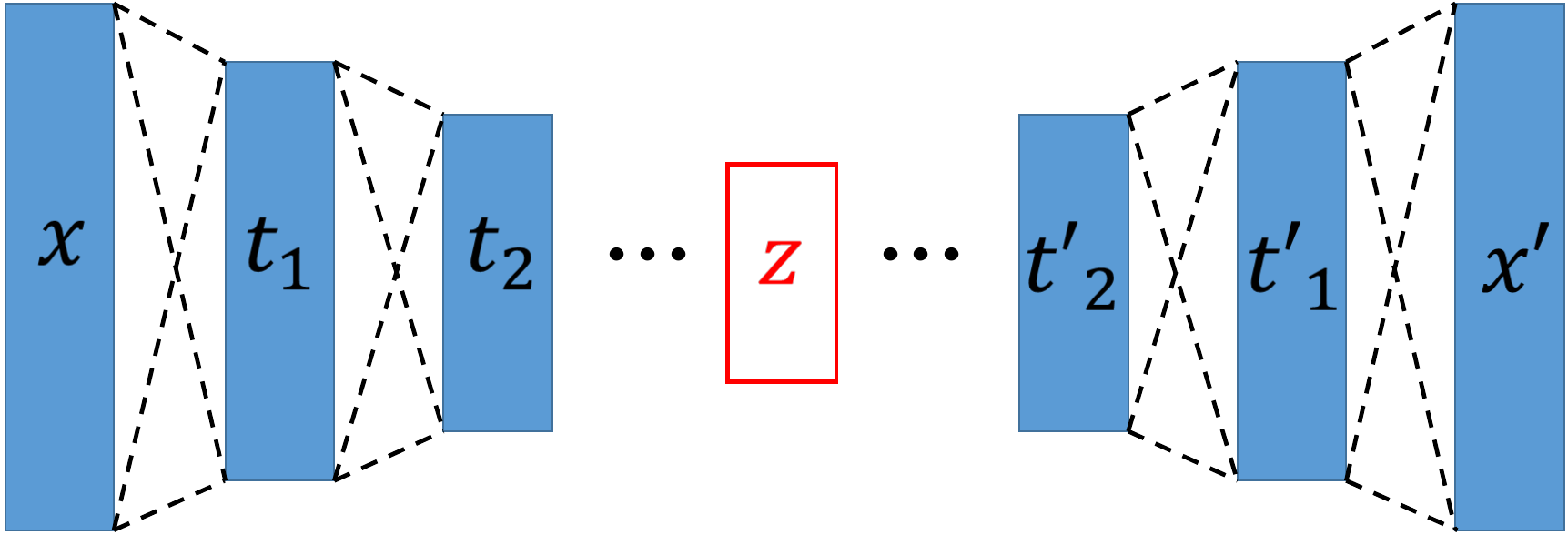}} \\
\subfigure[The graph representation of SAE with ($S$-1) hidden layers in both encoder and decoder.] {\includegraphics[width=.45\textwidth]{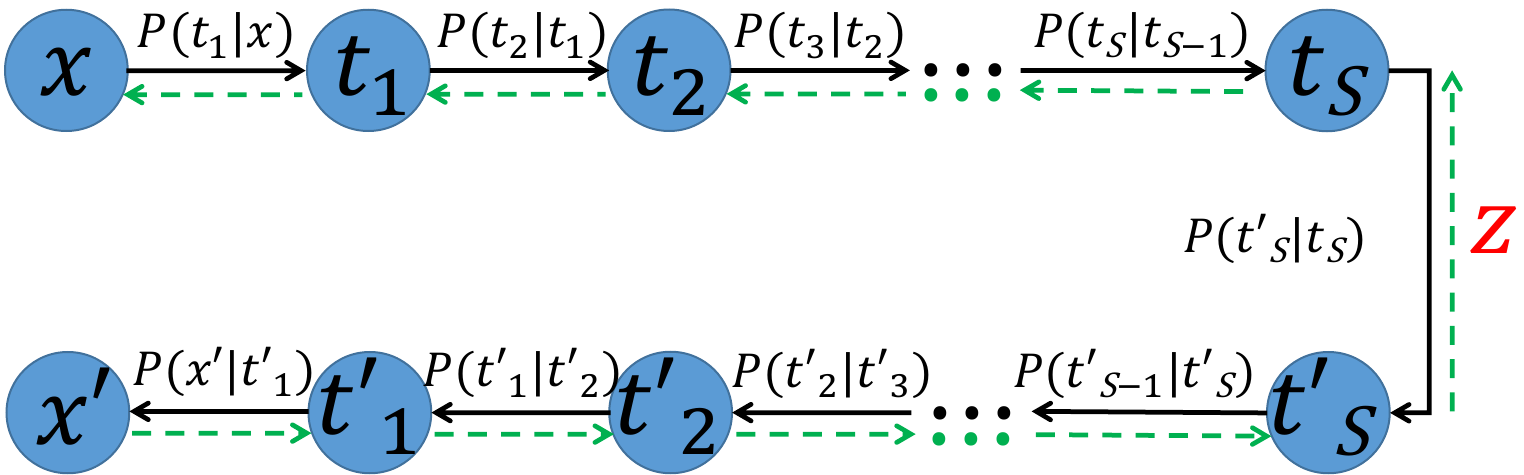}}
\end{tabular}
\caption{(a) shows a stacked autoencoder (SAE) with ($2S-1$) hidden layers and (b) shows its graph representation, where the black solid arrow denotes the direction of input feedforward propagation and the green dashed arrow denotes the direction of information flow in the error back-propagation phase. In both figures, $x$ is the observations in the input layer, $z$ is the codes in the bottleneck layer, and $x'$ is the output. The input data $x$ is encoded by probabilistic encoder $P(z|x)=\idotsint dt_1dt_2\dots dt_{s-1} P(t_1|x)P(t_2|t_1)\dots P(z|t_{s-1})$ and recovered by corresponding decoder $\tilde{P}(x'|z)=\idotsint dt'_1dt'_2\dots dt'_{s-1} \tilde{P}(x'|t'_1)\tilde{P}(t'_1|t'_2)\dots \tilde{P}(t'_{s-1}|z)$ leading to output data $x'$. Thus the SAE operates simultaneously to perform encoding and decoding. The data processing inequality associated with the mutual information states that $\mathbf{I}(X;X')\geq \mathbf{I}(T_1;T'_1)\geq\dots\geq \mathbf{I}(T_S;T'_S)$.\vspace{-0.0cm}}
\label{fig:architecture}
\end{figure}




As for the SAE, it is just a simple extension of the autoencoder (and also a special case of the MLP) that attempts to reconstruct its input. Different from the basic autoencoder or MLP, a SAE actually contains many symmetries that are not common in MLPs: first, there is an architectural symmetry between the encoder and the decoder; second, the target signal is the input. Therefore, given a basic SAE shown in Fig. \ref{fig:architecture}(a), where $X$ and $X'$ are the input and output variables respectively; $T_i$ $(1\leq i\leq S)$ denote different hidden layer representations in the encoder and $T'_i$ ($1\leq i\leq S$) denote different hidden layer representations in the decoder\footnote{All hidden layer representations mentioned in this paper refer to the values after linear/non-linear activation functions.}, it would be interesting to infer some intrinsic properties embedded in SAEs. To this end, we present the first two fundamental properties associated with SAEs:

~\\
\noindent
\textbf{Fundamental Property I:} The two fundamental DPIs in any feedforward DNNs mentioned earlier can be modified in SAEs, due to its symmetric architecture. Specifically, there exists DPIs in both encoder and decoder of SAEs, i.e., $\mathbf{I}(X;T_1)\geq\mathbf{I}(X;T_2)\geq\dots\geq\mathbf{I}(X;T_S)$ and $\mathbf{I}(X';T'_1)\geq\mathbf{I}(X';T'_2)\geq\dots\geq\mathbf{I}(X';T'_S)$.

~\\
\noindent
\textbf{Fundamental Property II:} There also exists a second type of DPI associated with the layer-wise mutual information specific to the SAEs, i.e., $\mathbf{I}(X;X')\geq \mathbf{I}(T_1;T'_1)\geq\dots\geq \mathbf{I}(T_S;T'_S)$. Note that, $\mathbf{I}(T_S;T'_S)$ reduces to $\mathbf{H}(Z)$ if a linear activation function is used in the bottleneck layer.

~\\
\noindent
\textbf{Reasoning:} On the one hand, the successive representations in the encoder should form a simple Markov chain \cite{shwartz2017opening,luttrell1994bayesian,huang2016flow}, i.e., $X\rightarrow T_1\rightarrow\dots\rightarrow T_{S-1}\rightarrow T_S$. On the other hand, the symmetric counterparts in the decoder will also form a simple Markov chain, i.e., $X'\rightarrow T'_1\rightarrow\dots\rightarrow T'_{S-1}\rightarrow T'_S$\footnote{Note that, this paper primarily aims to suggest simple information theoretic estimators to validate those possible properties. We leave a strict proof to the mentioned Markov chains as future work.}. This is because the SAE is symmetric, the decoder always ``undoes" the transformations operated by the encoder. Therefore, when the SAE is well-trained by backpropagation such that the layer-by-layer transition probabilities converge to an equilibrium \cite[Chapter~1 \& 3]{norris1998markov}, it makes sense to assume that $X\rightarrow T_1\rightarrow\dots\rightarrow T_{S-1}\rightarrow T_S$ and $X'\rightarrow T'_1\rightarrow\dots\rightarrow T'_{S-1}\rightarrow T'_S$ are ``dual" \cite[Chapter~II]{liggett2012interacting} \cite{jansen2014notion} of each other. Moreover, suppose a linear activation function is used in the bottleneck layer, the input (in virtue of the input feedforward chain) and the target signal (in virtue of the error backpropagate chain) make entropy the leading force for adaptation $\mathbf{I}(T_S;T'_S)=\mathbf{H}(Z)$. Both encoder and encoder enforce the entropy in the bottleneck layer, hence the role of the autoencoder is to maximize the entropy in the hidden layer. Because of this symmetry it makes sense to discuss only the partial feedforward Markov Chain between the input and the bottleneck layer, and the partial dual Markov chain between the output and the bottleneck layer.



The first type of DPI is a natural outcome of our assumption that both $X\rightarrow T_1\rightarrow\dots\rightarrow T_{S-1}\rightarrow T_S$ and $X'\rightarrow T'_1\rightarrow\dots\rightarrow T'_{S-1}\rightarrow T'_S$ form a Markov chain. The second type of DPI is actually built upon these two chains jointly. In fact, given a strictly convex function $f:(0,\infty)\mapsto\mathbb{R}$, the $f$-divergence can be defined \cite{csiszar1972class} as a generalized notion of the divergence between two probability distributions:
\begin{equation}
D_f(P_1\|P_2)=\int dxP_1(x)f\big(\frac{P_2(x)}{P_1(x)}\big).
\end{equation}

When the $f$-divergence was applied to the joint distribution (in the role of $P_1$) and the product of marginals (in the role of $P_2$) of two random variables (such as $X$ and $X'$), it yields a generalized notion of mutual information\footnote{The classical Kullback-Leibler (KL) divergence is a special case of $f$-divergence when $f(t)=t\log{t}$, $t\in (0,\infty)$. In this sense, the standard mutual information $\mathbf{I}(X;X')$, defined as $\mathbf{I}(X;X')=\int dxdx'P(x,x')\log\big(\frac{P(x,x')}{P(x)P(x')}\big)$, is just a special case of its generalized version $\mathbf{I}^Q(X;X')$.} \cite{cover2012elements}:
\begin{equation}
\begin{split}
\mathbf{I}^Q(X;X')&=\int dxdx'P(x,x')Q\big(\frac{P(x)P(x')}{P(x,x')}\big) \\
&=\int dxdx'P(x,x')Q\big(\frac{P(x')}{P(x'|x)}\big),
\end{split}
\end{equation}
which was shown in \cite{csiszar1972class} to obey a second type of DPI, thus extending the famed (first type of) DPI in a broader sense, i.e.,
\begin{equation}
\mathbf{I}^Q(X;X')\geq \mathbf{I}(\bar{X};\bar{X}'),
\end{equation}
where $\bar{X}$ and $\bar{X}'$ denote indirect observations to $X$ and $X'$, respectively. The equality holds if and only if $\bar{X}$ and $\bar{X}'$ are the sufficient statistic with respect to $\{X,X'\}$ \cite{csiszar1972class,merhav2011data}. By referring to the above descriptions, we expect a monotonically non-increasing trend (as the number of layers increases) of the mutual information between the layer output and their ``symmetric" counterparts, i.e., we cannot gain more mutual information when we process the original observations in a deeper layer. Therefore this imposes an upper limit to the number of layers in practical situations, which has not been yet recognized as a limitation in deep learning empirical validation.

\subsection{Two types of Information Planes (IPs)} \label{section3.2}
The IP, initiated in \cite{tishby2015deep} and matured in \cite{shwartz2017opening}, creates an observable space for how SGD optimizes the DNN: compression by diffusion creates efficient internal representations in each layer. However, we would like to note that this work only applies to the information bottleneck training method and has not been exploited to help us to appropriately design DNNs, nor learning, i.e., the mechanism of compression has not been fully elucidated yet. Very likely, it is produced by the progressive saturation of the network units during training, which changes the PDF of the outputs of the units, creating very peaky distributions near the saturation values. Additionally, although the IP presents an explicit way to inspect pairs of layer-wise mutual information simultaneously, we demonstrate that inspecting only the mutual information that each layer preserves about the input with respect to the target is insufficient to provide a comprehensive understanding of neural network training.

To this end, we extend the definition of IP into a broader and more general perspective and suggest two novel IPs: 1) the plane of information quantities that each hidden layer preserves about the input with respect to the output, i.e., $\mathbf{I}(X;T)$ with respect to $\mathbf{I}(T;Y')$ ($\mathbf{I}(T;X')$ for SAEs); 2) the plane of information quantity that each hidden layer (in the encoder) preserves about the input with respect to the information quantity that counterpart (or symmetric) hidden layer (in the decoder) preserves on the output, i.e., $\mathbf{I}(X;T)$ with respect to $\mathbf{I}(Y';T')$ ($\mathbf{I}(X';T')$ for SAEs). We term them \emph{Information Plane} I (IP-I) and \emph{Information Plane} II (IP-II), respectively. One should note that, due to the special architecture of SAE, there are other ways to define IP. For example, the plane of information quantity that each hidden layer (in the encoder) preserves about the input with respect to the information quantity that the counterpart (or symmetric) hidden layer (in the decoder) preserves also on the input, i.e., $\mathbf{I}(X;T)$ with respect to $\mathbf{I}(X;T')$. This plane can demonstrate the tendency of $\mathbf{I}(X;T)$ to approach the value of $\mathbf{I}(X;T')$ in a more obvious manner. We term it IP-III and demonstrate this property in Section~\ref{conclusions}.

The IP-I makes a simple modification to the original IP in \cite{shwartz2017opening} by substituting $\mathbf{I}(T,Y)$ with $\mathbf{I}(T,Y')$. The motivation for this modification is straightforward: the output layer contains significant signals to analyze in any neural network architecture \cite{haykin1994neural}. Moreover, if we insist on using $\mathbf{I}(T,Y)$ for analyzing SAEs, the IP curve reduces to a line because the target is just a mirrored input, thus resulting in a poor visualization and the loss of useful information. The IP-II, on the other hand, compares the amount of information that $T_i$ gained from $X$ with the amount of information that $T'_i$ gained from $X'$, which also provide an implicit measure on how marginal distributions $P(T_i)$ and $P(T'_i)$ match each other. Such visualization is promising, as it tells us when the symmetric layer-wise SAE pairs matches well under the objective of minimizing reconstruction error. We believe it has the potential to guide the development of new training methods in a feedforward manner\footnote{A similar vision is shown in \cite{bengio2014auto}, but no solid examples are presented.}, as an alternative to the standard back-propagation method, and may also help answer questions about generalization.


~\\
\noindent
\textbf{Fundamental Property III:} We expect the existence of a different behavior in the IP (a bifurcation point) associated with the SAE bottleneck layer size (e.g., the number of units in this layer) that is controlled by the intrinsic dimensionality of the given data, i.e., the curves in the IP-I or IP-II might demonstrate two distinct patterns depending upon the size of bottleneck layer is larger or smaller than the intrinsic dimensionality of input data.


~\\
\noindent
\textbf{Reasoning:} The exploration of bifurcation or critical points is not new in machine learning and time series analysis. An interesting example comes from time series analysis, in which the Takens' Theorem \cite{takens1981detecting} states that if the degree of freedom of a dynamical system is confined to an attractor $\mathcal{M}$ of dimension $D$ in the state space, then the topology of the attractor that characterizes the dynamical system can be discovered from the analysis of the time series data when concatenating $M>2D$ previous outputs of this dynamical system into a vector (called a delay coordinate map)~\cite{yap2011stable}. In other words, when $M\leq2D$, it is impossible to recover the attractor without any distortion, so results will suffer. Taken's Theorem links the intrinsic dimensionality $D$ of a nonlinear time series with the required embedding length $M$. From the perspective of manifold learning, the bottleneck layer of SAE should play the similar role as the delay coordinate map~\cite{potapov2002neural}.
Therefore, it is reasonable to conjecture that the SAE's bottleneck layer is controlled by the data's characteristics (e.g., the intrinsic dimensionality) for good performance. If the IP is a good indicator to the dynamics of training, then we should see a difference in the dynamics of learning for bottleneck layers that are above and below the intrinsic dimensionality of the data.

We test this property by altering the topologies of SAEs, specifically the number of units in the bottleneck layer. Note that, the \textbf{Fundamental Property III} can also guide the search of the intrinsic dimensionality (range) of given data. Please refer to Algorithm~\ref{ID_estimation_alg} for more details. However, one should also note that in order to make Algorithm~\ref{ID_estimation_alg} more applicable in practice, more efforts need to be made. For example, how to determine a reliable effective dimensionality searching range and how to quantitatively determine the existence of ``compression" phase without human observation. We leave it as future work.

\begin{algorithm}[!ht]
\caption{Effective Dimensionality Estimation with Information Plane}
\label{ID_estimation_alg}
\begin{algorithmic}[1]
\Require
Input data $\mathbf{X}\in \mathbb{R}^{N\times m}$, effective dimensionality searching range $[S_{lo},S_{up}]$.
\Ensure
Effective dimensionality lower bound $E_{lo}$; Effective dimensionality upper bound $E_{up}$.
\For {$i = 1$ to $\left[\frac{S_{up}-S_{lo}}{2}\right]$}
\State Training SAE with bottleneck layer size $S_{lo}+i$; Plot and observe flow of information in the IP-I.
\If {(Most curves in the IP-I have the ``compression" phase.)}
\State $E_{lo}\leftarrow S_{lo}+i$; Break.
\EndIf
\EndFor
\For {$i = 1$ to $\left[\frac{S_{up}-S_{lo}}{2}\right]$}
\State Training SAE with bottleneck layer size $S_{up}-i$; Plot and observe flow of information in the IP-I.
\If {(One or two curves in the IP-I do not have the ``compression" phase.)}
\State $E_{up}\leftarrow S_{up}-i$; Break.
\EndIf
\EndFor
\end{algorithmic}
\end{algorithm}

\section{Experiments} \label{experiments}

This section presents two sets of experiments to corroborate our Section \ref{section3} fundamental properties directly from data and the nonparametric statistical estimators put forth in this work. Specifically, Section \ref{section4.2} validates the first type of DPI and also demonstrates the two IPs defined in Section~\ref{section3.2} to illustrate the existence of bifurcation point that is controlled by the given data, whereas Section \ref{section4.1} validates the second type of DPI raised in Section~\ref{section3.1}. Note that, we also give a preliminary interpretation to the observations shown in Section \ref{section4.1}, by inspecting the hidden codes distribution in the training phase. All the experiments reported in this work were conducted in MATLAB $2016$b under a Windows $10$ $64$bit operating system. Companion source code is available from \\ \url{http://bit.ly/ITL_autoencoder}.

\begin{figure}[!htbp]
\centering
\begin{tabular}{ccc}
\subfigure[] {\includegraphics[width=.15\textwidth]{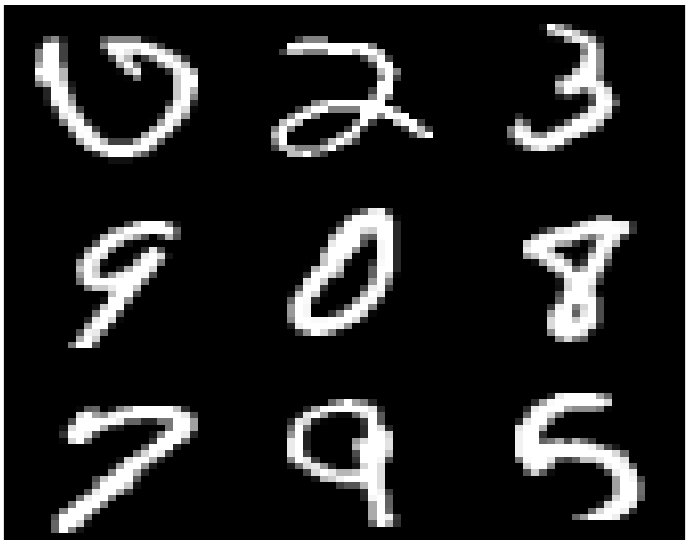}}
\subfigure[] {\includegraphics[width=.15\textwidth]{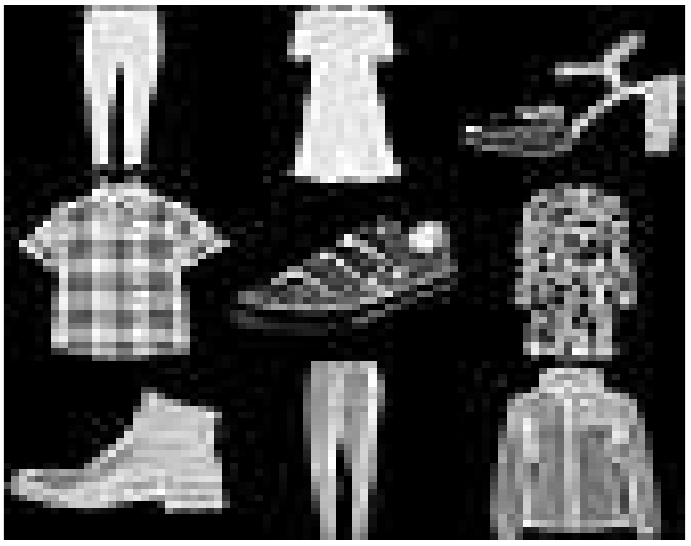}}
\subfigure[] {\includegraphics[width=.15\textwidth]{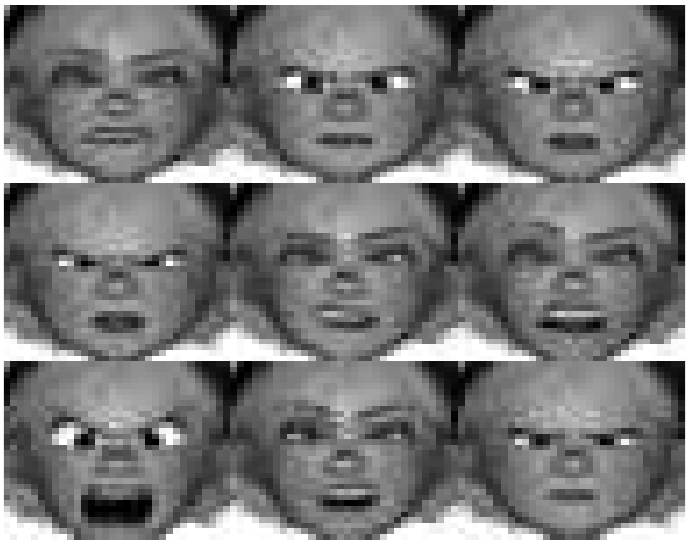}} & \\
\end{tabular}
\caption{Visualization of training samples selected from (a) MNIST; (b) Fashion-MNIST and (c) FERG-DB (after pre-processing).\vspace{-0.0cm}}
\label{fig:sample_image}
\end{figure}

The real-world datasets selected for evaluation are explained next and Fig. \ref{fig:sample_image} depicts the representative images from each dataset.

(a) MNIST \cite{lecun1998gradient}, contains a training set of $60000$ images and a testing set of $10000$ images of $10$ handwritten digits. Each digit has been normalized and centered in a $28\times28$ image. The thickness, height, angular alignment, and relative position in a frame are some of the intrinsic hidden properties that govern the generation of the examples for each digit manifold. The entire data set of images can be considered as an embedded manifold plus additive noise \cite{brahma2016deep}.

(b) Fashion-MNIST \cite{xiao2017fashion}, is a recently released benchmark to test machine learning algorithms. As an alternative to MNIST, it features the same image size, data format and the structure of training and testing splits. The only difference is that the handwritten digits are replaced with different fashion products, like T-Shirts or Trousers. This will provide diversity for the size of the embedded manifolds.

(c) FERG-DB \cite{aneja2016modeling}, contains $55767$ face images from $6$ stylized characters with annotated facial expressions. The images for each character are grouped into $7$ types of expressions, i.e., anger, disgust, fear, joy, neutral, sadness and surprise. Each image has a resolution of either $768\times768$ (full resolution) or $256\times256$ (reduced size). We take the inner $180\times180$ pixels of each reduced size image and resize it to the size of $32\times32$ pixels from which we form a vector with $1024$ dimensions as the input. According to our initial investigation on FERG-DB using t-SNE \cite{maaten2008visualizing}, the variance among different subjects is much higher than the variance among different facial expressions. This means that the embedded manifold of the data set perhaps is too high to be well estiamted with the available data. For this reason, we only conduct one subject-dependent facial expression classification experiment using all facial expression images of ``Bonnie". The selected dataset is separated into $10000$ for training and $1271$ for testing.

In this paper, we use the basic SAE with no other architecture constraints. The activation functions of all the neurons are sigmoid functions which have been theoretically proven effective in encouraging sparse representation \cite{arpit2016regularized}. The only exception comes from the bottleneck layer, in which a simple linear activation function is employed to obey the Folded Markov Chain (FMC) architecture\footnote{Note that, the \textbf{Fundamental Property I} and the \textbf{Fundamental Property II} still hold even though we relax the selection of activation functions. Without loss of generality, this work only considers sigmoid activation function in hidden layers and linear activation function in the bottleneck layer.}~\cite{luttrell1994bayesian}. The networks were trained using SGD under the objective of minimizing reconstruction error power. The topology of SAEs on MNIST and Fashion-MNIST is fixed to be ``$784$-$1000$-$500$-$250$-$K$-$250$-$500$-$1000$-$784$" as suggested in \cite{hinton2006reducing}, where $K$ denotes the number of neurons in the bottleneck layer. Unlike MNIST, the topology of SAEs on FERG-DB is selected as ``$1024$-$512$-$256$-$100$-$K$-$100$-$256$-$512$-$1024$". Due to page limitations, we only demonstrate the results on MNIST in Sections~\ref{section4.2} and~\ref{section4.1}. The corresponding results on Fashion-MNIST and FERG-DB, and the robust analysis on kernel size turning on information quantities estimation, are shown in the appendix.

The training of SAE is iterated for $100$ epochs, with the mini-batch size set to $100$. The information quantities mentioned in this paper are estimated using the matrix-based functional of Renyi's $\alpha$-entropy \cite{giraldo2015measures} with $\alpha=1.01$ to approximate Shannon's entropy as suggested in \cite{giraldo2015measures,giraldo2013rate}. Since the kernel size $\sigma$ in the estimation of Renyi's $\alpha$-entropy is a compromise between bias and variance of the estimator, we must select the kernel size properly because the estimated entropy values depend upon the kernel size. We tune the kernel size $\sigma$ by the Silverman's rule of thumb \cite{silverman1986density}\footnote{More details on selection of $\sigma$ is demonstrated in Section~\ref{conclusions}.}, which takes into consideration the change in kernel size with the dimension of the data.

\subsection{Experimental validation of Fundamental Properties I $\&$ III} \label{section4.2}

\begin{figure*}[!htbp]
\centering
\begin{tabular}{ccc}
\subfigure[IP-I (encoder part) when $K=14$] {\includegraphics[width=.45\textwidth]{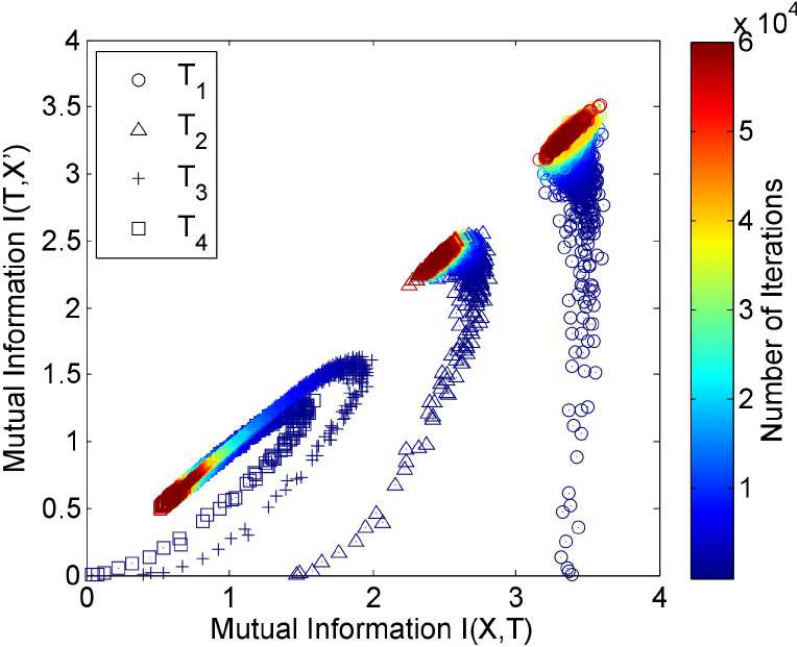}}
\subfigure[IP-I (decoder part) when $K=14$] {\includegraphics[width=.45\textwidth]{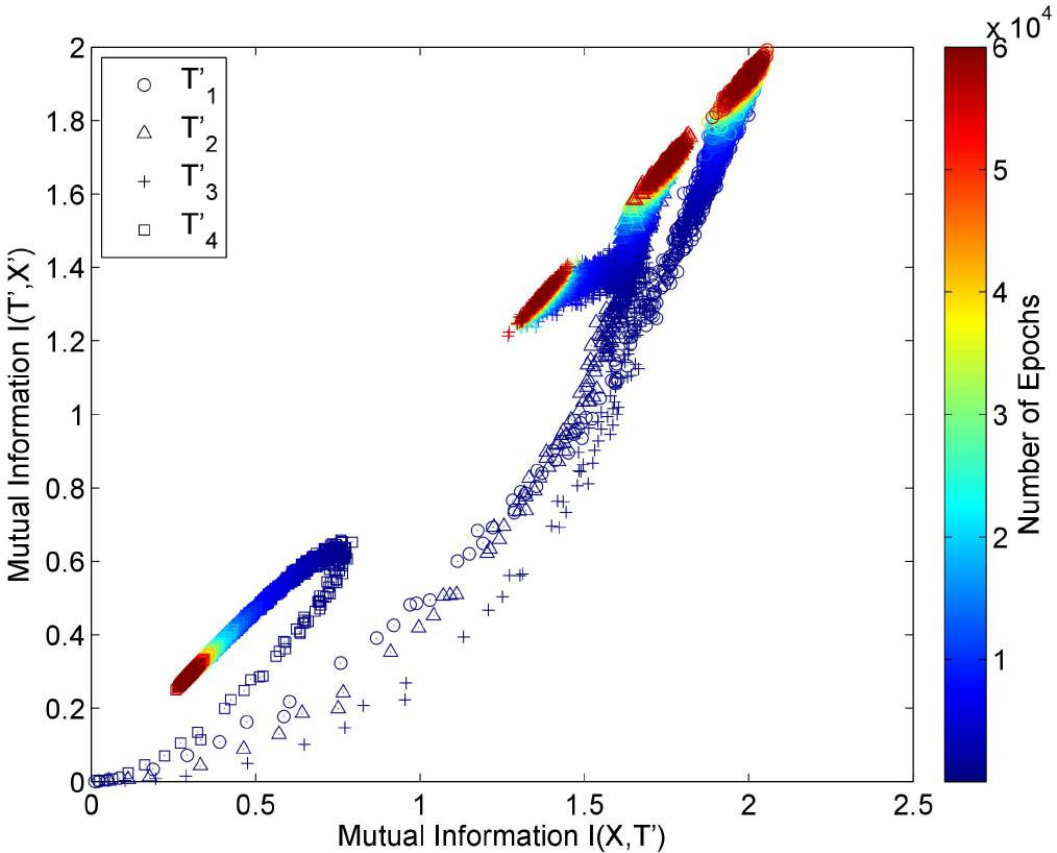}} & \\
\subfigure[IP-I (encoder part) when $K=36$] {\includegraphics[width=.45\textwidth]{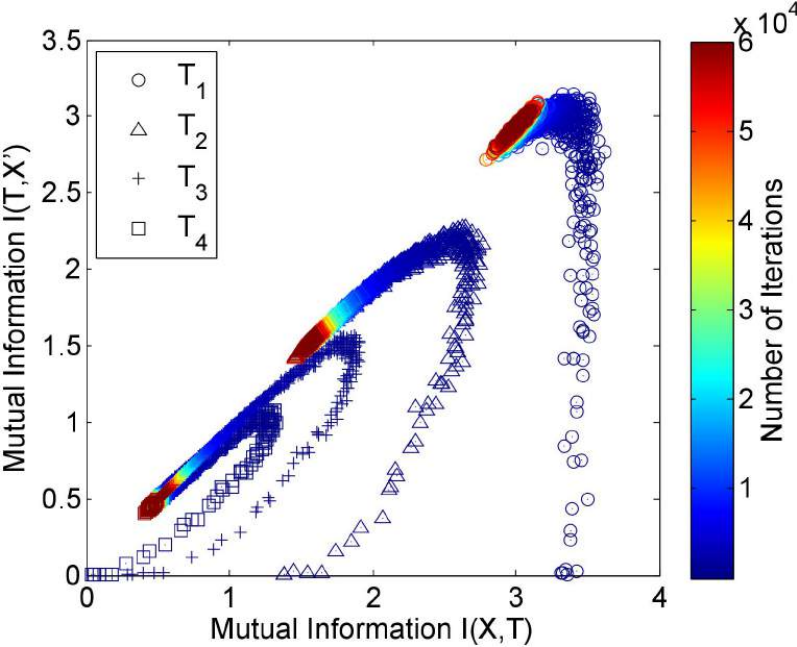}}
\subfigure[IP-I (decoder part) when $K=36$] {\includegraphics[width=.45\textwidth]{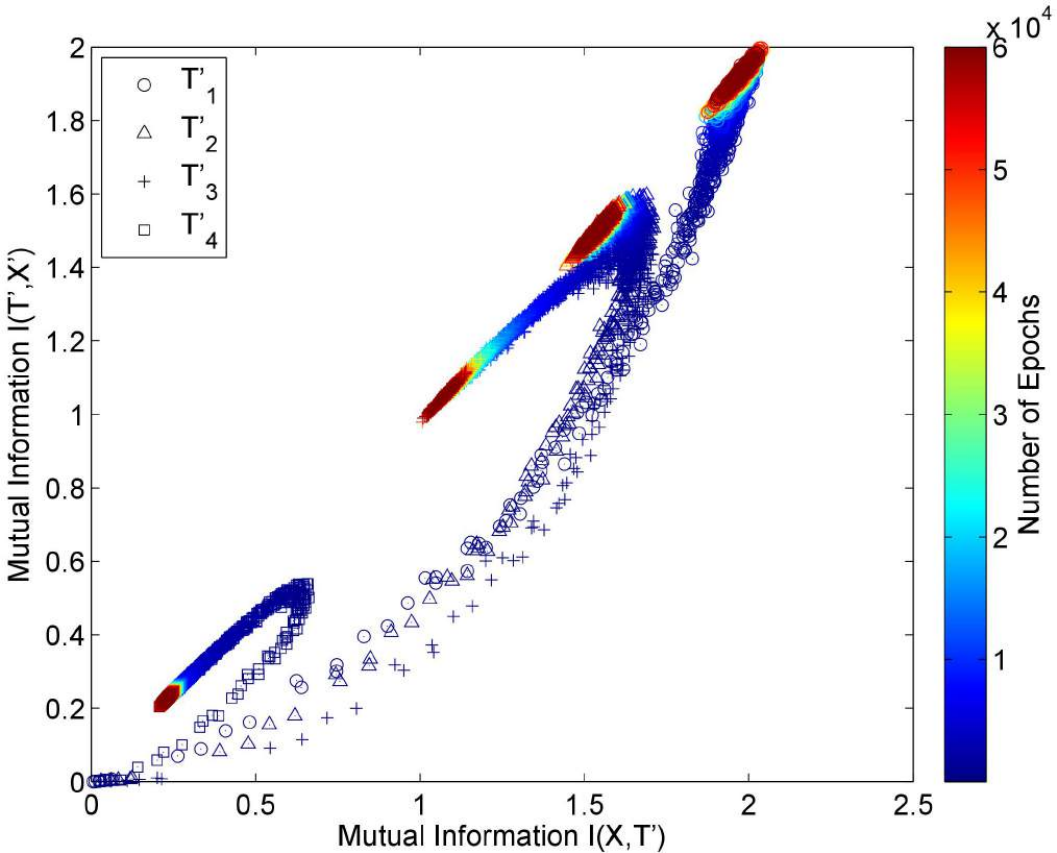}} & \\
\end{tabular}
\caption{The validation of bifurcation point associated with $K$. (a), (c), (e), (g) and (i) demonstrate the IP-I ($T$ is in the encoder module) with $K$ equals to $2$, $7$ $11$, $14$ and $32$ respectively, whereas (b), (d), (f), (h) and (j) demonstrate the corresponding IP-I when $T$ is in the decoder module. As can be seen, the general patterns of curves in IPs begin to have a transition between $K=11$ and $K=14$. This suggests an effective dimensionality of MNIST dataset is approximately $12$ or $13$.\vspace{-0.0cm}}
\label{fig:bifurcation}
\end{figure*}

We first validate the \textbf{Fundamental Properties I} $\&$ \textbf{III}, since these two properties can be easily verified with IPs. Specifically, we expect the existence of DPI such that $\mathbf{I}(X;T_1)\geq\mathbf{I}(X;T_2)\geq\dots\geq\mathbf{I}(X;T_S)$ and $\mathbf{I}(X';T'_1)\geq\mathbf{I}(X';T'_2)\geq\dots\geq\mathbf{I}(X';T'_S)$. We also expect a bifurcation point associated with the value of $K$ that is controlled by the intrinsic dimensionality $D$ \cite{camastra2016intrinsic} of given data\footnote{Note that, the intrinsic dimensionality mentioned in this paper only refers to an effective dimensionality that can give a reasonable fit \cite{wang2008scale}. We leave a rigorous investigation to the physical meaning of this dimensionality as future work.}: the curves in the IPs may demonstrate distinct behavior depending on $K>D$ or $K\leq D$. To corroborate this argument, we test different SAE topologies with $K$ ranging from $2$ to $36$. The corresponding IP-I is shown in Fig. \ref{fig:bifurcation}.

Fig. \ref{fig:bifurcation} shows the behavior of the IP-I in the encoder and the decoder for several values of the bottleneck layer size $K$. As can be seen, $\mathbf{I}(X;T_1)$ is consistently larger than $\mathbf{I}(X;T_2)$, $\mathbf{I}(X;T_2)$ is consistently larger than $\mathbf{I}(X;T_3)$ and $\mathbf{I}(X;T_3)$ is consistently larger than $\mathbf{I}(X;T_4)$, no matter the value of $K$. Moreover, after a very short period of training (the SAE is trained with a certain fidelity), $\mathbf{I}(X';T'_1)$ is consistently larger than $\mathbf{I}(X';T'_2)$, $\mathbf{I}(X';T'_2)$ is consistently larger than $\mathbf{I}(X';T'_3)$ and $\mathbf{I}(X';T'_3)$ is consistently larger than $\mathbf{I}(X';T'_4)$, no matter the value of $K$.
Therefore, the first type of DPI always holds, i.e., $\mathbf{I}(X;T_1)\geq\mathbf{I}(X;T_2)\geq\dots\geq\mathbf{I}(X;T_S)$ and $\mathbf{I}(X';T'_1)\geq\mathbf{I}(X';T'_2)\geq\dots\geq\mathbf{I}(X';T'_S)$.

\begin{figure*}[!htbp]
\centering
\begin{tabular}{ccc}
\subfigure[IP-I (encoder part) when $K=2$] {\includegraphics[width=.45\textwidth]{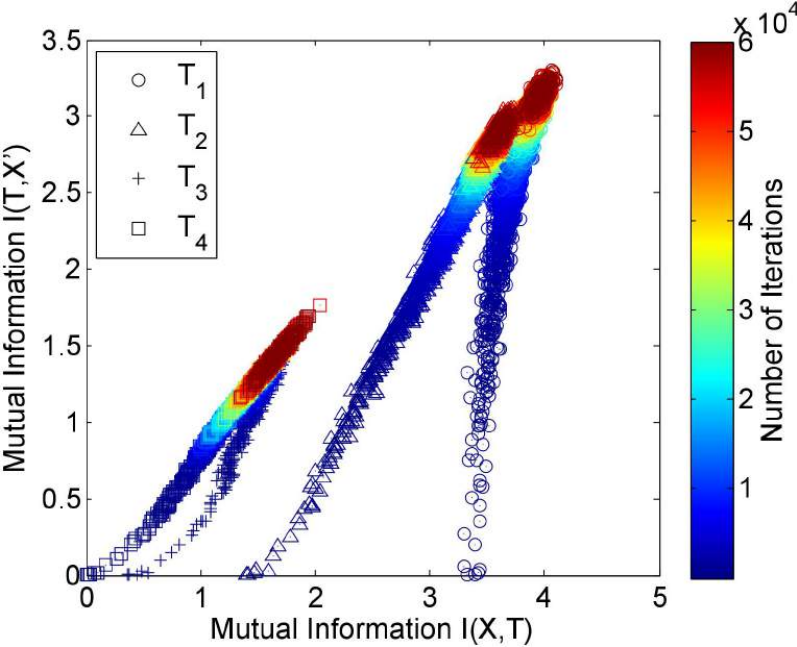}}
\subfigure[IP-I (decoder part) when $K=2$] {\includegraphics[width=.45\textwidth]{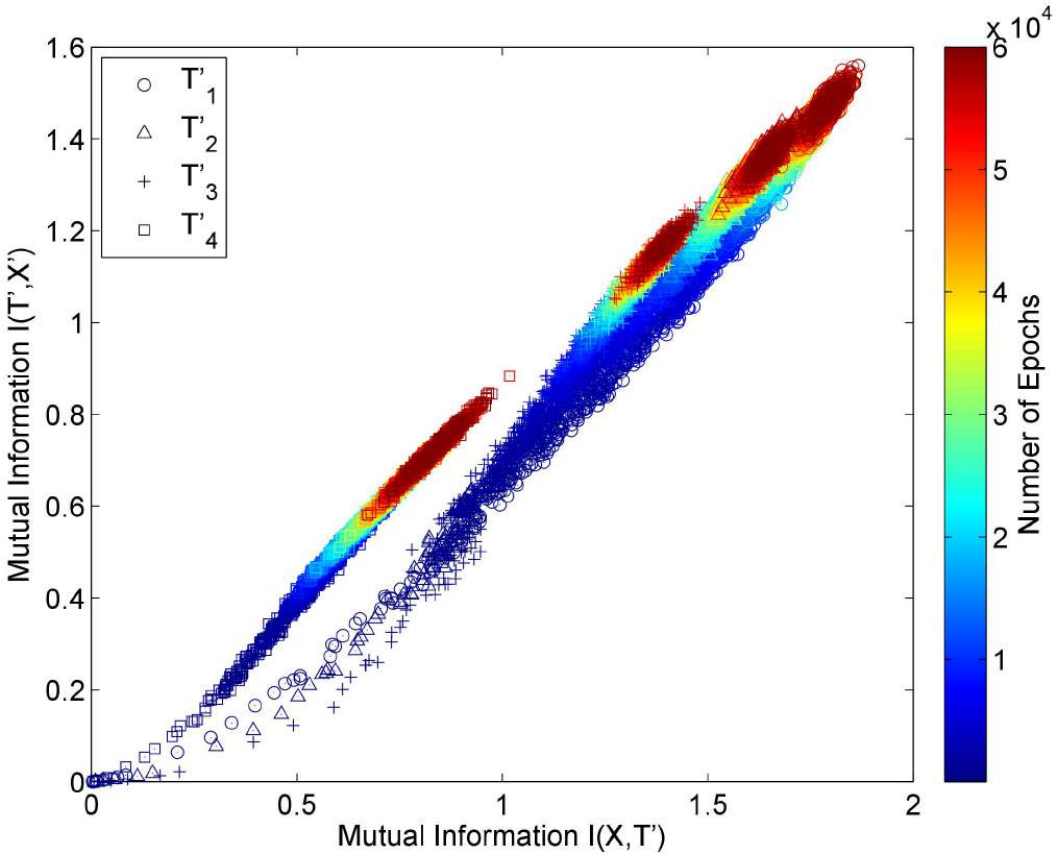}} & \\
\subfigure[IP-I (encoder part) when $K=7$] {\includegraphics[width=.45\textwidth]{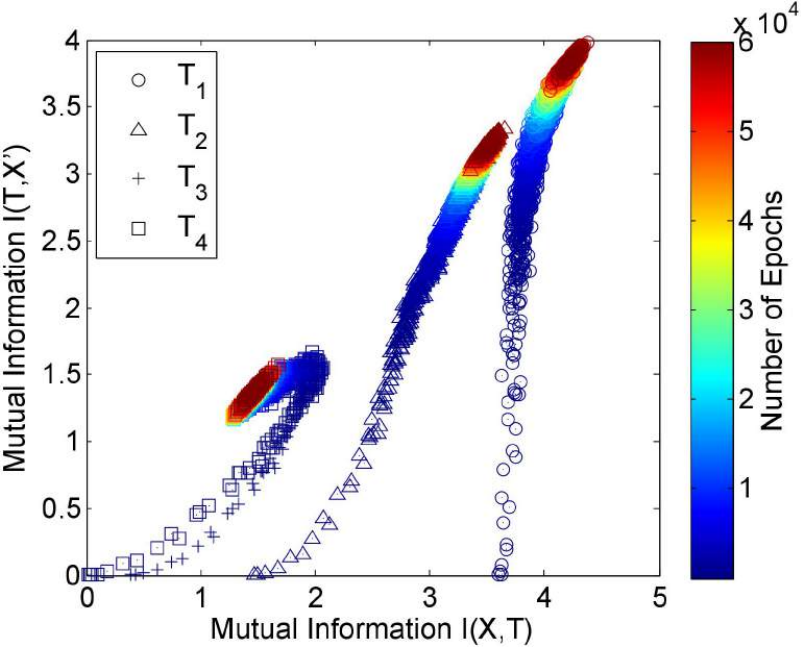}}
\subfigure[IP-I (decoder part) when $K=7$] {\includegraphics[width=.45\textwidth]{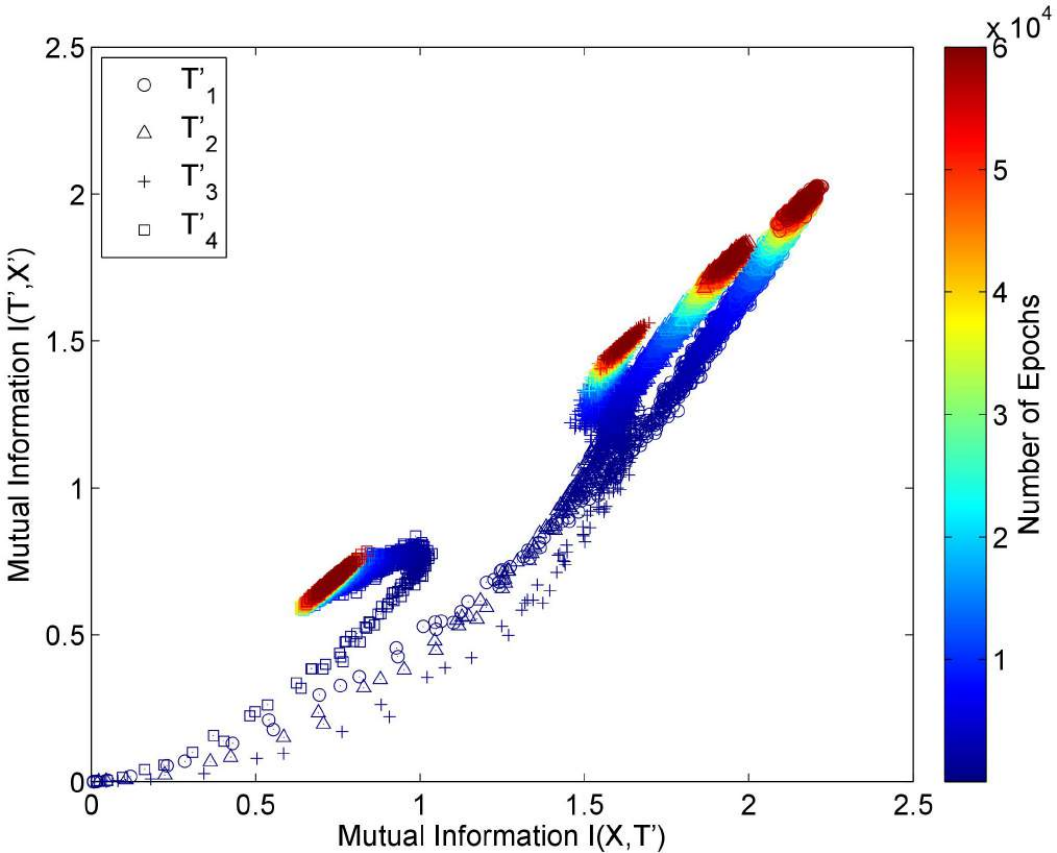}} & \\
\subfigure[IP-I (encoder part) when $K=11$] {\includegraphics[width=.45\textwidth]{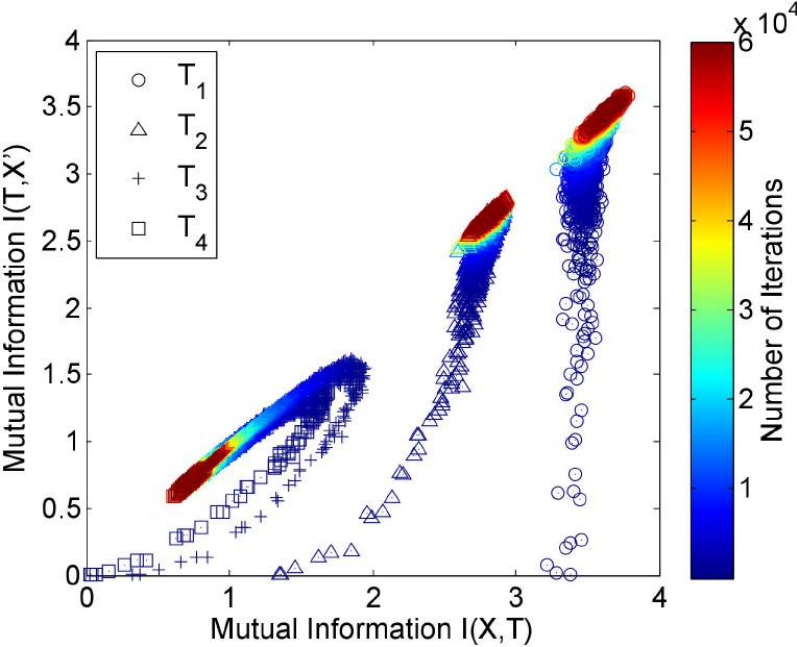}}
\subfigure[IP-I (decoder part) when $K=11$] {\includegraphics[width=.45\textwidth]{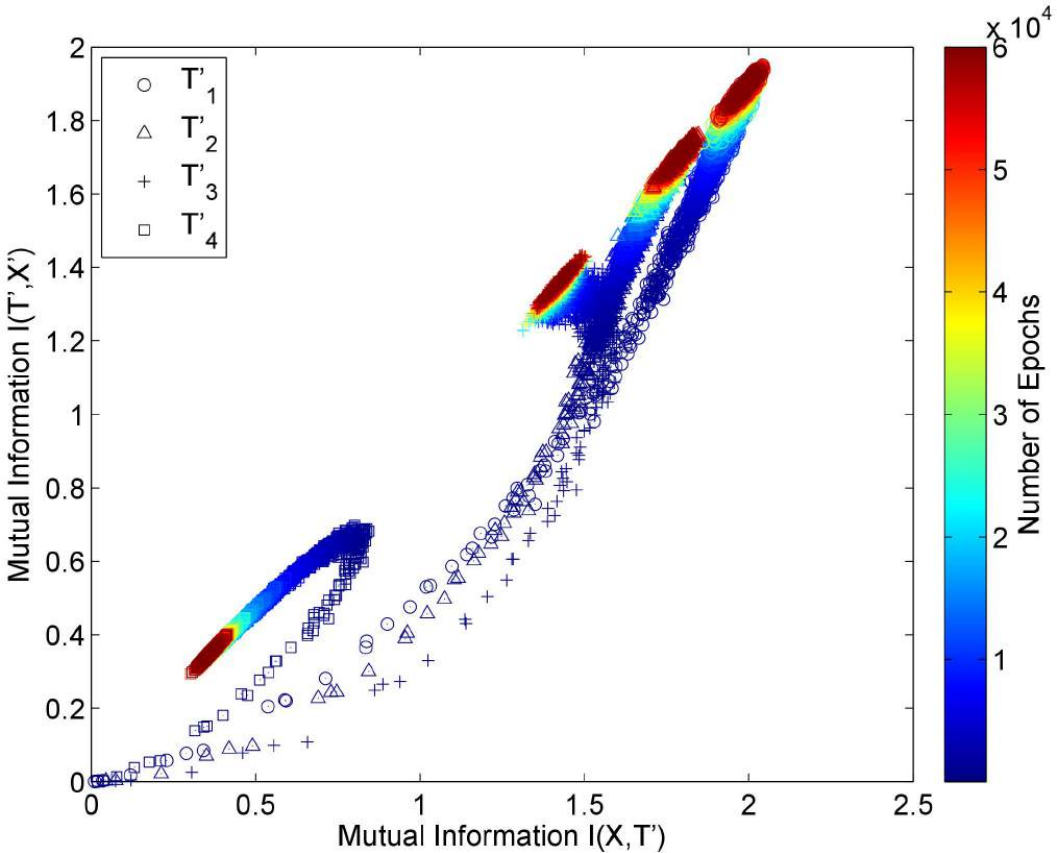}} & \\
\end{tabular}
\end{figure*}

A finer analysis of the IP-I curves shows that $\mathbf{I}(X;T)$ starts higher for the layers closer to the input (shallow), but the rate of increase of $\mathbf{I}(X;T')$ is the fastest for the first layer of the encoder, showing that early in learning the shallow layers learn faster about the desired than the deeper layers. The vertical increase of $\mathbf{I}(X;T'_1)$ is very likely due to the overcomplete first layer projection space. For a properly set bottleneck layer, the final value of $\mathbf{I}(X;T')$ in each layer tends to be close to the initial value of $\mathbf{I}(X;T)$ (see IP-I encoder curves), which is very interesting since it means that the mutual information between the input and the layer is transferred to the mutual information between the layer and the desired. This can be potentially used to  to evaluate if the overall system is trained well enough, as well as to properly set the learning rates for each layer. Notice also that the shallow layers are more sensitive to the size of the bottleneck layer $K$. The behavior of the curves in the layers close to the bottleneck layer approaches the entropy of the codes, which means that the SAE learning is controlled by the evolution of the entropy in the bottleneck layer codes, which does not conform with the IB principle. Therefore, the change of curve patterns in IP-I seems a good indicator of the DPI property. The picture for the IP-I layers in the decoder is not as clear, because the mutual information only stabilizes once the bottleneck layer settles.

We now start the analysis of the IP-I encoder because it is the one that refers to the coding of information. The effective dimensionality for this dataset is between $D=12$ or $13$. When $K>D$, the (majority of) pair-wise mutual information curves in the IP-I start to increase up to a point and then go back approaching the bisector of the plane, i.e., converging to the line $x=y$. This is not surprising as the optimal\footnote{Here the ``optimal" means the SAE is trained with the objective of minimizing the distortion measure, i.e., mean square error.} solution resides in this bisector because $X=X'$. However, for the case of $K\leq D$, we observed a different behavior. In fact, the curves associated with $T_1$ and $T_2$ keep increasing with two different slopes up to a point, while the $\mathbf{I}(T;X')$ is smaller and increases slower such that the curve is further away from the bisector.

The above results corroborate partially the conclusion in \cite{shwartz2017opening}: there indeed exists two separate phases when using the standard SGD to train DNNs. In the first and short phase, the network is progressively fitting the data manifold, whereas in the second and much longer phase, the purpose of training is to fine tune the representation locally. If the cost function is mutual information  as in \cite{shwartz2017opening} one could in fact talk about compression of representations. However, here with mean square error (MSE) training, this metaphor does not hold since MSE is not a sparsifying criterion. Nevertheless, we also see that the combined nonlinearity of the units enhances the quality of the local representations, perhaps by moving the units to saturation. However, what we discovered is that, for the SAE, this conclusion only holds in an ideal scenario (i.e., $K>D$). By contrast, if $K\leq D$, the network is incapable of fitting the data with high fidelity. As a result, the representations are unable to match the local neighborhoods of the data manifold (see Fig.~\ref{fig:distribution}).

Finally, it is worth noting that our estimated $D$ matches well the values of intrinsic dimensionality given by benchmarking estimators. In fact, the intrinsic dimensionality estimated by the Maximum Likelihood Estimation (MLE) \cite{levina2005maximum}, the Minimum Neighbor Distance (MiND) Estimator \cite{lombardi2011minimum} and the Dimensionality from Angle and Norm Concentration (DANCo) \cite{ceruti2014danco} are $12$, $13$ and $15$, respectively.

\subsection{Experimental validation of Fundamental Property II} \label{section4.1}
\begin{figure*}[!htbp]
\centering
\begin{tabular}{ccc}
\subfigure[] {\includegraphics[width=.3\textwidth]{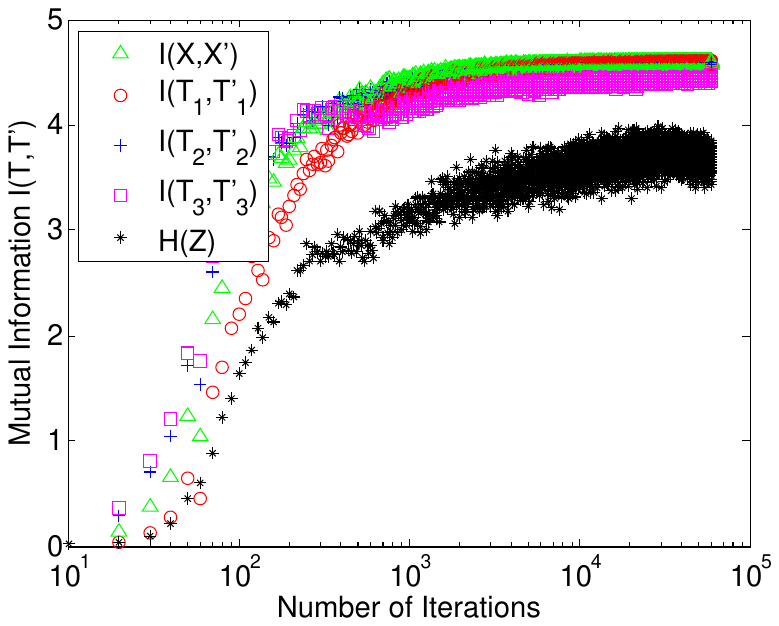}}
\subfigure[] {\includegraphics[width=.3\textwidth]{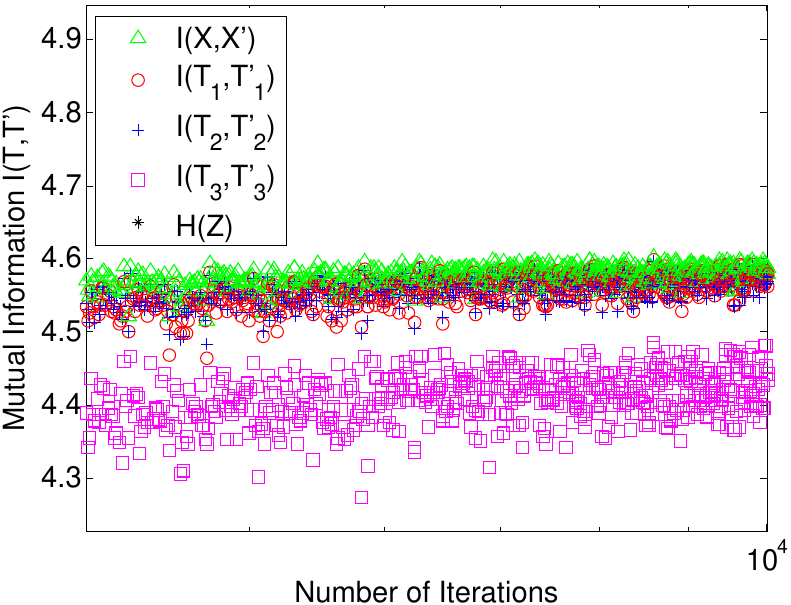}}
\subfigure[] {\includegraphics[width=.3\textwidth]{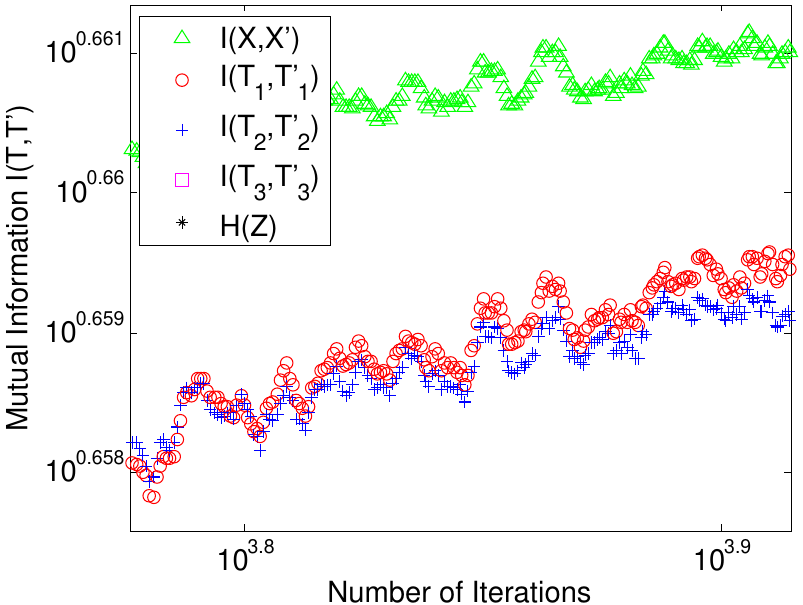}} & \\
\subfigure[] {\includegraphics[width=.3\textwidth]{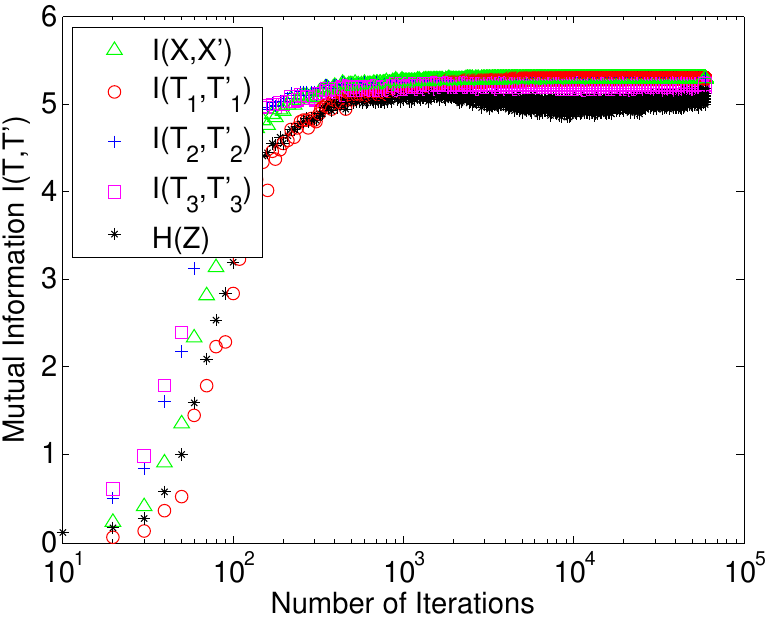}}
\subfigure[] {\includegraphics[width=.3\textwidth]{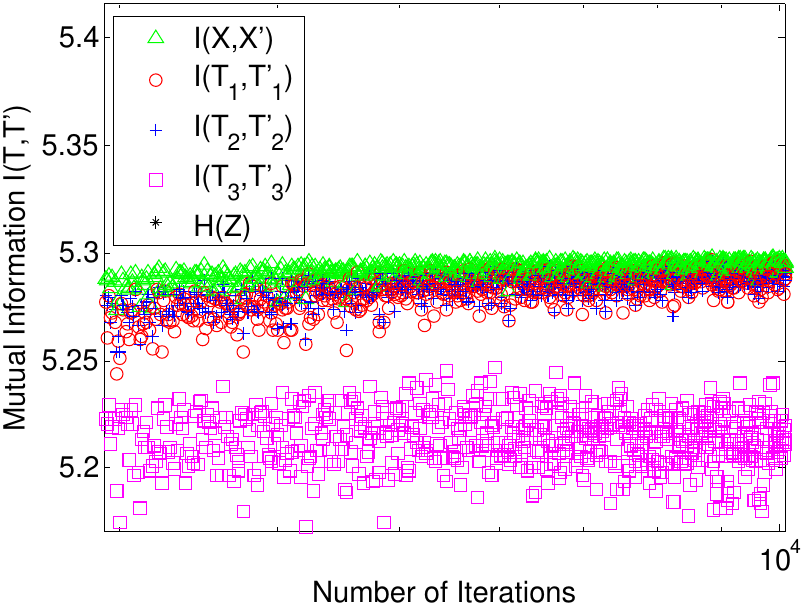}}
\subfigure[] {\includegraphics[width=.3\textwidth]{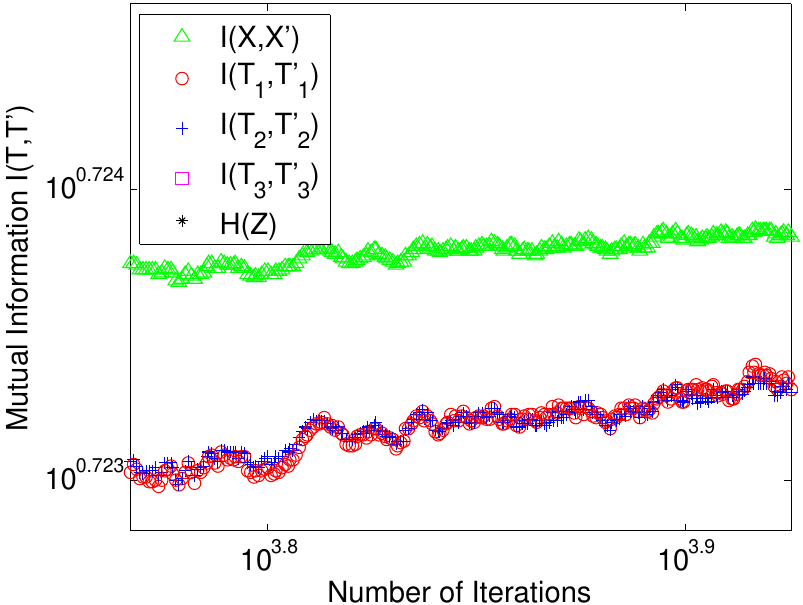}} & \\
\subfigure[] {\includegraphics[width=.3\textwidth]{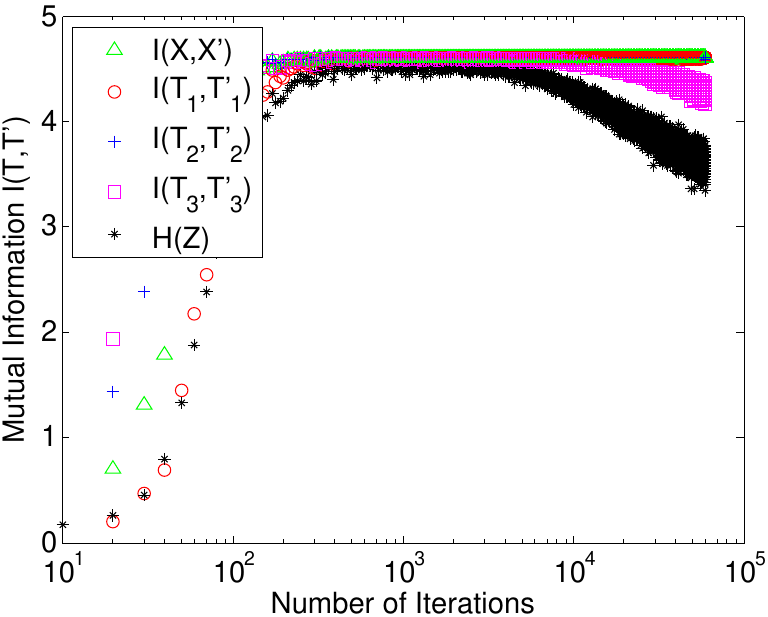}}
\subfigure[] {\includegraphics[width=.3\textwidth]{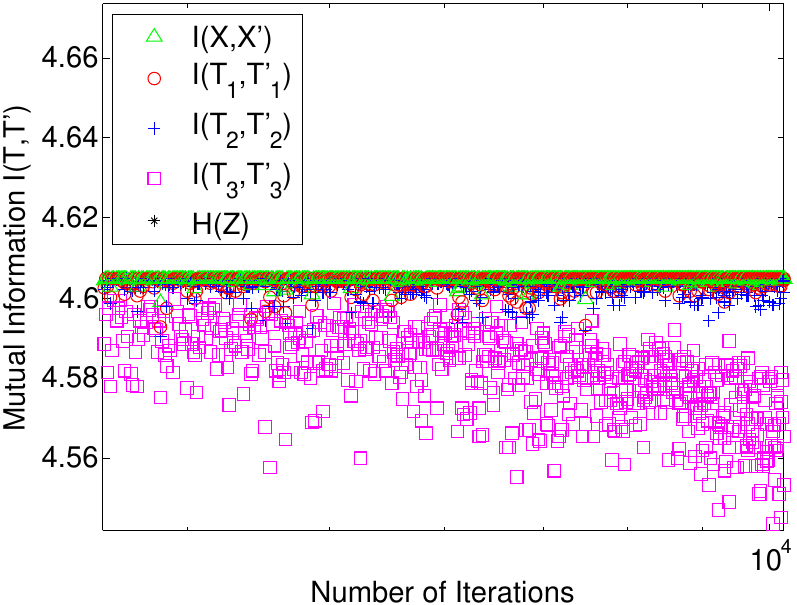}}
\subfigure[] {\includegraphics[width=.3\textwidth]{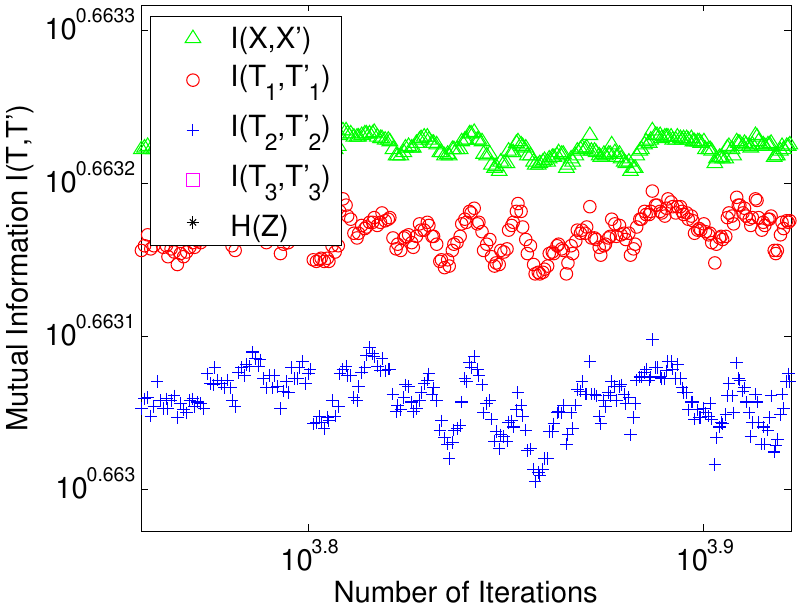}} & \\
\subfigure[] {\includegraphics[width=.3\textwidth]{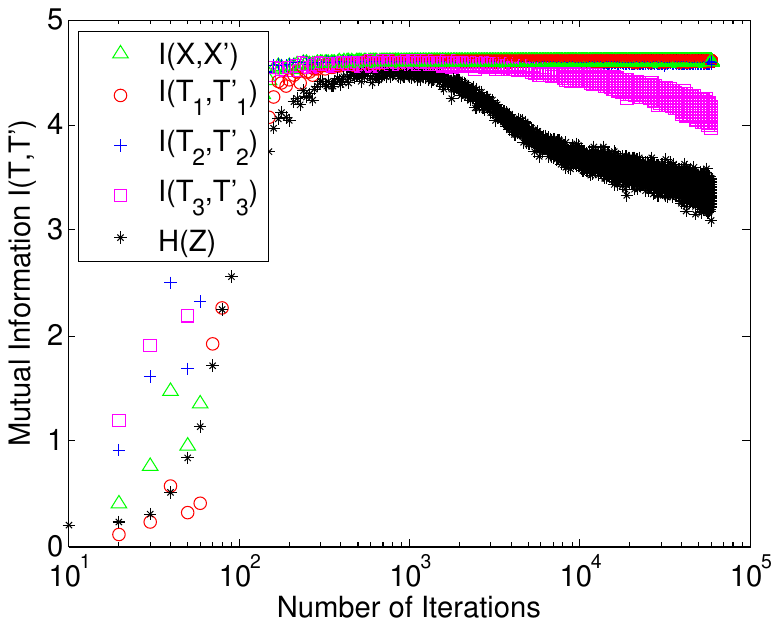}}
\subfigure[] {\includegraphics[width=.3\textwidth]{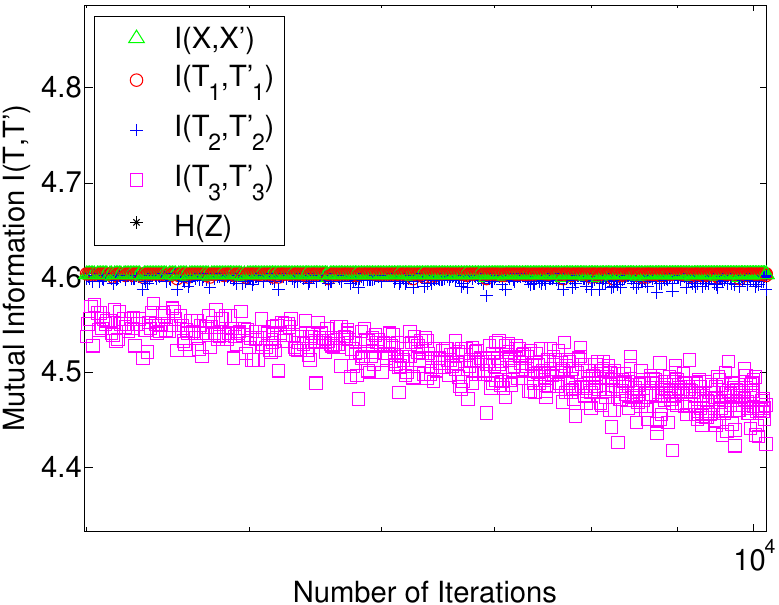}}
\subfigure[] {\includegraphics[width=.3\textwidth]{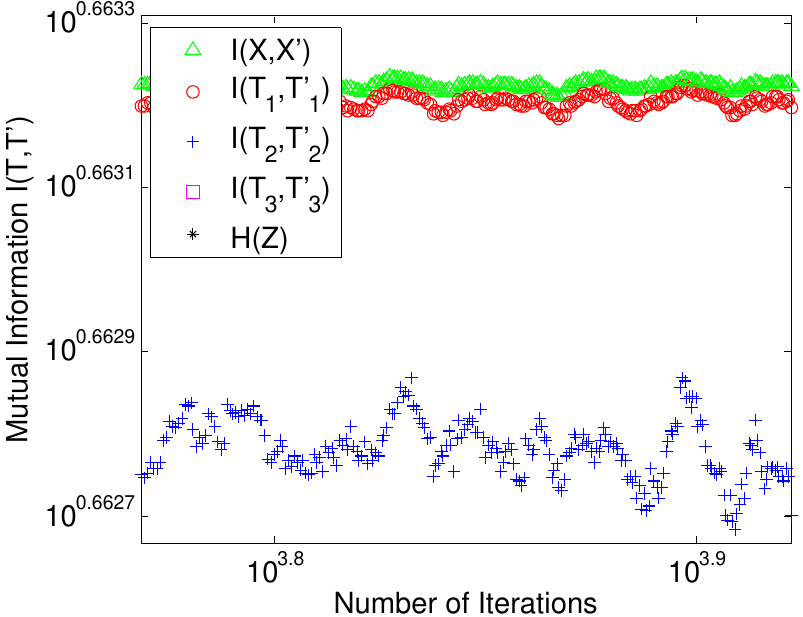}} & \\
\end{tabular}
\caption{The validation of data processing inequality (DPI) associated with the layer-wise mutual information. (a) demonstrates the layer-wise mutual information when $K=2$; (b) shows the zoom-in result of (a); (c) shows the moving average (over $100$ samples) result of (a) on a log scale. Similarly, (d) demonstrates the layer-wise mutual information when $K=11$, (e) and (f) show, respectively, the zoom-in result and the moving average result of (d); (g) demonstrates the layer-wise mutual information when $K=14$, (h) and (i) show, respectively, the zoom-in result and the moving average result of (g); (j) demonstrates the layer-wise mutual information when $K=36$, (k) and (l) show, respectively, the zoom-in result and the moving average result of (j). In all sub-figures, the green triangles denote the mutual information between $X$ and $X'$, the red circles denote the mutual information between $T_1$ and $T'_1$, the blue plus signs denote the mutual information between $T_2$ and $T'_2$, the magenta squares denote the mutual information between $T_3$ and $T'_3$, and the black asterisks denote the mutual information between $T_4$ and $T'_4$ (reduces to the entropy of $T_4$ in our case).\vspace{-0.0cm}}
\label{fig:DPI_validation}
\end{figure*}

We then validate the second type of DPI in the \textbf{Fundamental Property II}, that is $\mathbf{I}(X;X')\geq \mathbf{I}(T_1;T'_1)\geq\dots\geq \mathbf{I}(T_S;T'_S)=\mathbf{H}(Z)$. To this end, we demonstrate, in Fig. \ref{fig:DPI_validation}, the layer-wise mutual information corresponding to four network topologies with different number of neurons in the bottleneck layer, i.e., $K=2$, $K=11$, $K=14$ and $K=36$. The experimental results corroborate the DPI property - the deeper the neural network, the more information about the input is lost, regardless of the network topologies. 
In fact, by referring to the first two columns of Fig.~\ref{fig:DPI_validation}, both $\mathbf{I}(T_3;T'_3)$ and $\mathbf{H}(Z)$ gradually deviate from $\mathbf{I}(X;X')$, $\mathbf{I}(T_1;T'_1)$ and $\mathbf{I}(T_2;T'_2)$, with $\mathbf{I}(T_3;T'_3)$ consistently larger than $\mathbf{H}(Z)$. But there are distinct differences between $K=2$ and $K=36$. For $K=2$, $\mathbf{I}(X;X')$ is strictly larger than $\mathbf{I}(T_1;T'_1)$, $\mathbf{I}(T_1;T'_1)$ is strictly larger than $\mathbf{I}(T_2;T'_2)$ (see Fig.~\ref{fig:DPI_validation}(c)), and $\mathbf{H}(Z)$ is rapidly increasing but still much smaller than $\mathbf{I}(T_3;T'_3)$ (see Fig.~\ref{fig:DPI_validation}(a)). However, for $K=36$, although $\mathbf{I}(X;X')$ and $\mathbf{I}(T_1;T'_1)$ are larger than $\mathbf{I}(T_2;T'_2)$, these two values are almost the same (see Fig.~\ref{fig:DPI_validation}(l)). Moreover, $\mathbf{H}(Z)$ is decreasing drastically after a short period of training (see as shown in Fig.~\ref{fig:DPI_validation}(j)). This phenomenon does not occur for $K=2$, because a $2$-dimensional projection space is insufficient to guarantee lossless reconstruction of input data, thus the encoder layers, in the goal of transferring maximal entropy to the bottleneck are also distorting the representations.


\begin{figure*}[!htbp]
\centering
\begin{tabular}{ccc}
\subfigure[Iteration $1$ ($K=2$)] {\includegraphics[width=.23\textwidth]{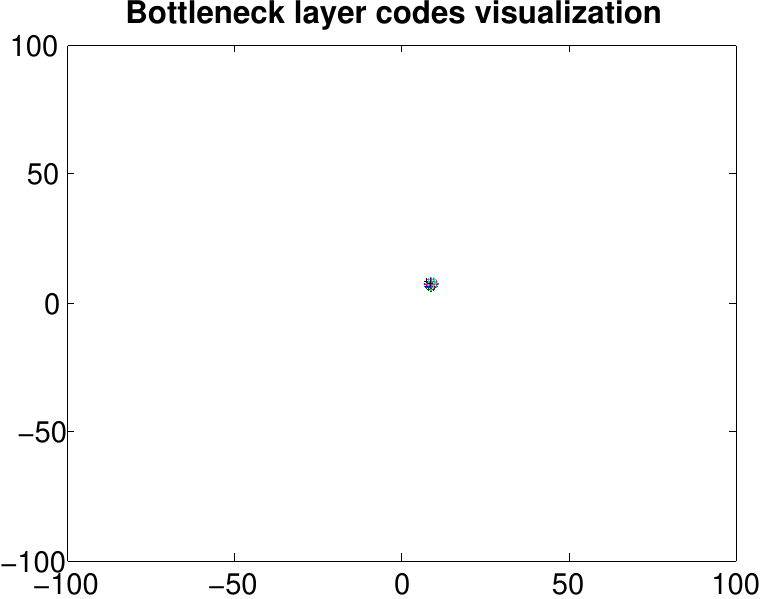}}
\subfigure[Iteration $1\times10^3$ ($K=2$)] {\includegraphics[width=.23\textwidth]{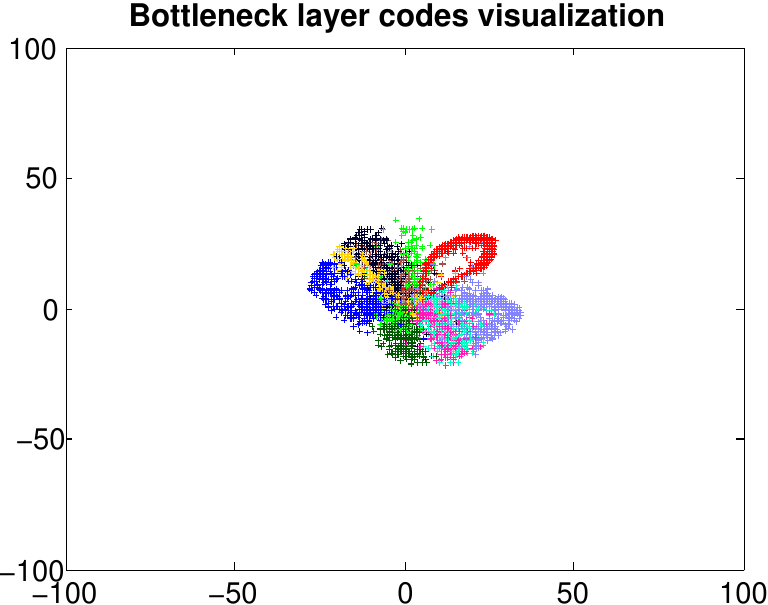}}
\subfigure[Iteration $1\times10^4$ ($K=2$)] {\includegraphics[width=.23\textwidth]{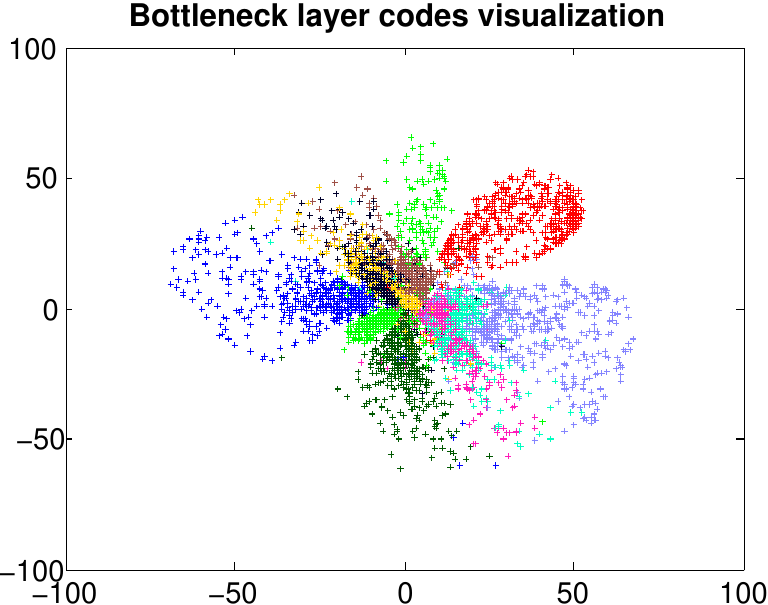}}
\subfigure[Iteration $6\times10^4$ ($K=2$)] {\includegraphics[width=.23\textwidth]{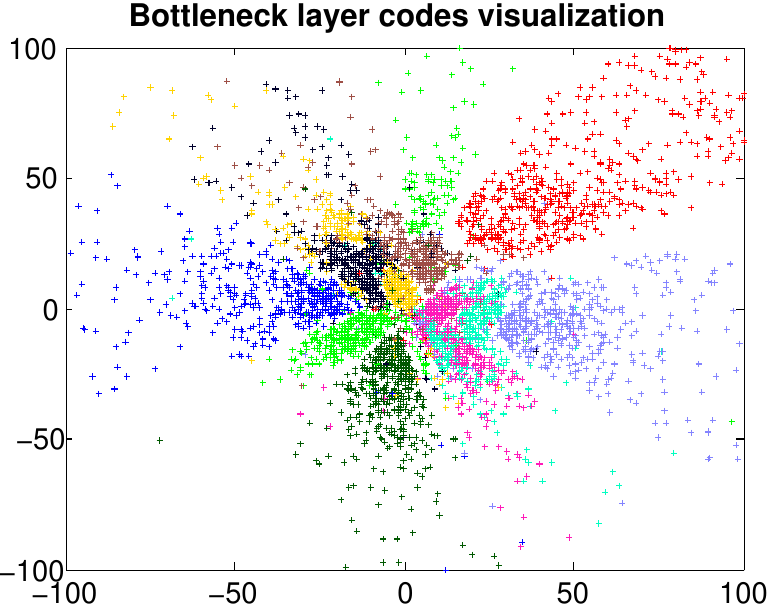}} & \\
\subfigure[Iteration $1$ ($K=36$)] {\includegraphics[width=.23\textwidth]{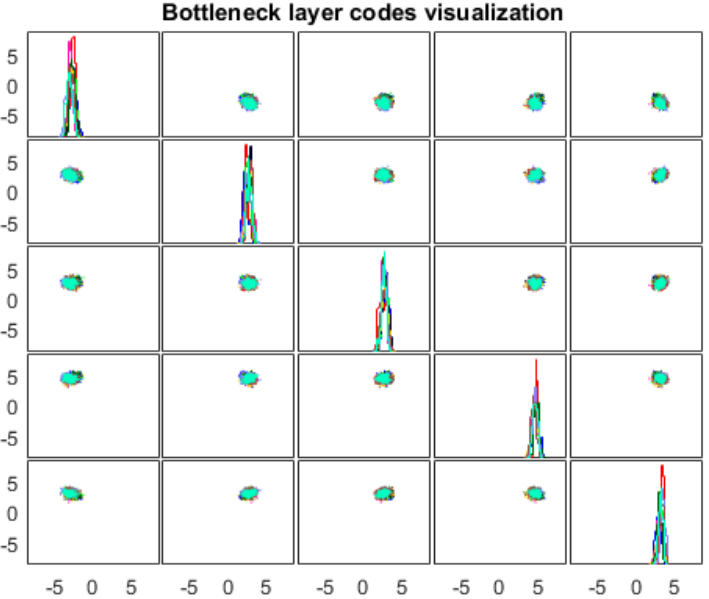}}
\subfigure[Iteration $1\times10^3$ ($K=36$)] {\includegraphics[width=.23\textwidth]{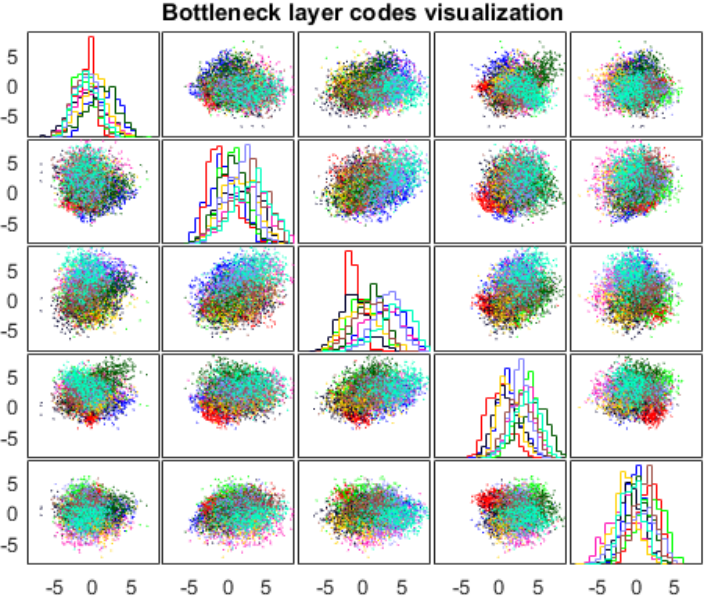}}
\subfigure[Iteration $1\times10^4$ ($K=36$)] {\includegraphics[width=.23\textwidth]{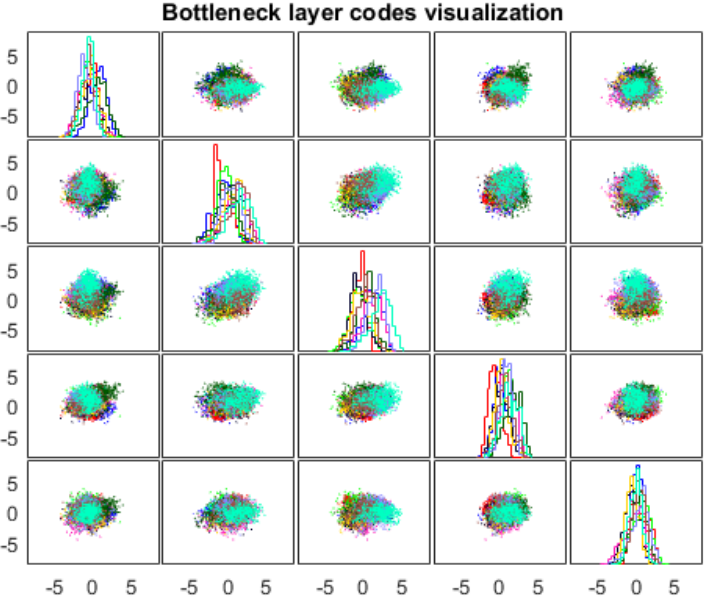}}
\subfigure[Iteration $6\times10^4$ ($K=36$)] {\includegraphics[width=.23\textwidth]{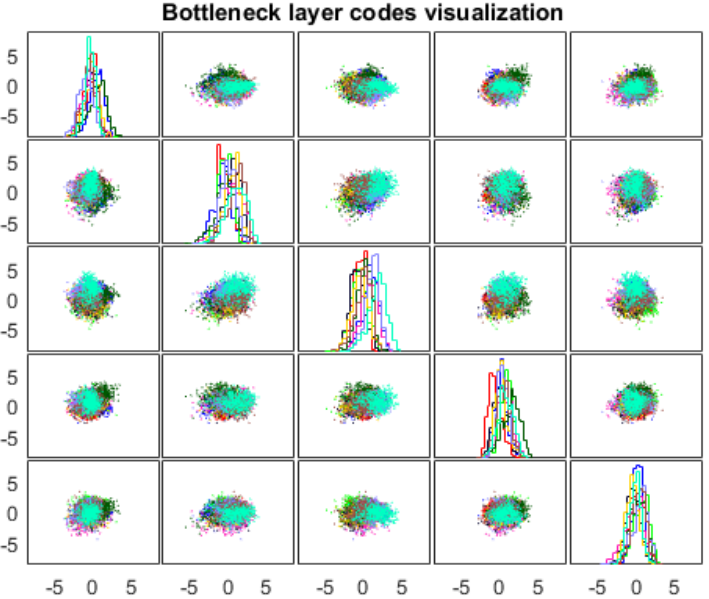}} & \\
\end{tabular}
\caption{Bottleneck layer codes visualization. (a)-(d) demonstrate the codes distribution at iteration $1$, $1\times10^3$, $1\times10^4$ and $6\times10^4$ respectively when $K=2$, whereas (e)-(h) demonstrate the codes distribution at iteration $1$, $1\times10^3$, $1\times10^4$ and $6\times10^4$ respectively when $K=36$. In each sub-figure, different color denotes different class.\vspace{-0.0cm}}
\label{fig:distribution}
\end{figure*}

But the most interesting observation is that the entropy of bottleneck codes $\mathbf{H}(T_4)$ begins to decrease after a certain number of iterations when $K=36$ while for $K=2$ no similar phenomena occurs. This suggests that the bottleneck codes undergo different forms of specialization when the reconstruction reaches to a certain fidelity, but the existence of compression phase depends on whether the topology can guarantee an information-lossless reconstruction. Otherwise, we think that this specialization results in distortion of the original manifold. To verify this, we demonstrate the geometric distribution of $T_4$ for both $K=36$ and $K=2$ in Fig. \ref{fig:distribution}. Note that, it is impossible to explicitly observe $36$-dimensional point clouds in a $2$-dimensional (or $3$-dimensional) space, thus we randomly select $5$ (out of $36$) neurons and use scatterplot matrix to visualize the geometric distribution changes.    	

As can be seen, in both cases the codes attempt to fill up the projection space, thus increasing the overall entropy (see Figs \ref{fig:distribution}(b) and \ref{fig:distribution}(f)). However, in the case of $K=2$, the clusters break up (green, yellow) and the codes persistently enlarge to cover the projection space, with no trend to decrease the redundancy (see Figs \ref{fig:distribution}(c) and \ref{fig:distribution}(d)). This is because the compressed $2$-dimensional space is insufficient to accommodate the natural structure of the data, so continuous training distorts the local structure of the data manifold. The parameter adaptation tries to minimize the error by spreading the codes in a larger region of the space, which reduces classification error until unit saturation takes over. However, the volume of the projected data is maximum so local structure is lost. By contrast, in the case of $K=36$, when the reconstruction reaches a certain fidelity and the network has sufficient discriminative power with $36$ degrees of freedom (see Fig. \ref{fig:distribution}(f)), the manifold of each class begins to shrink (see Figs \ref{fig:distribution}(g) and \ref{fig:distribution}(h)), thus decreasing the overall entropy but preserving cluster separability, which achieves also a very good classification accuracy.

\begin{figure*}[!htbp]
\centering
\begin{tabular}{ccccc}
\subfigure[Iteration $1$ ($K=2$)]{\includegraphics[width=.23\textwidth]{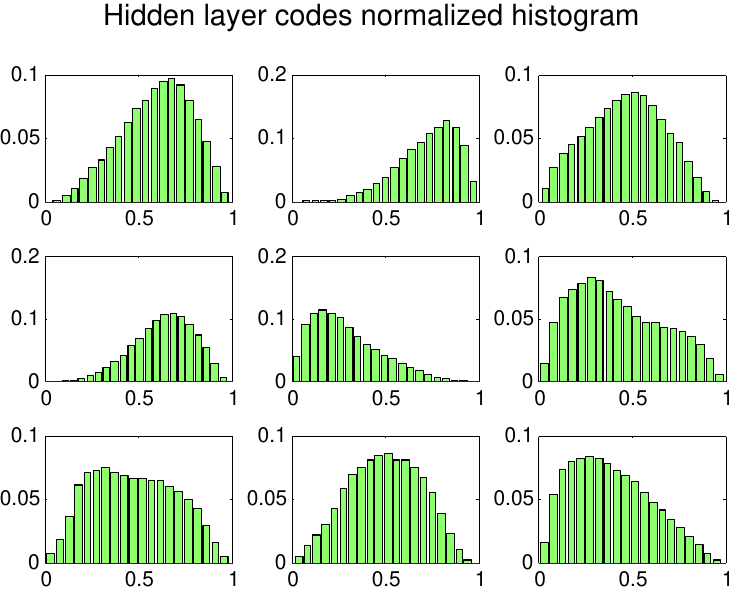}}
\subfigure[Iteration $1\times10^3$ ($K=2$)]{\includegraphics[width=.23\textwidth]{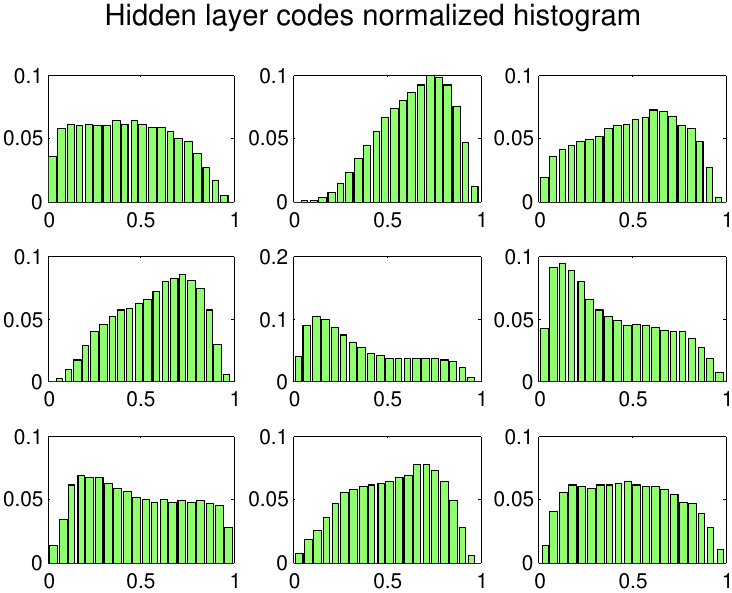}}
\subfigure[Iteration $1\times10^4$ ($K=2$)]{\includegraphics[width=.23\textwidth]{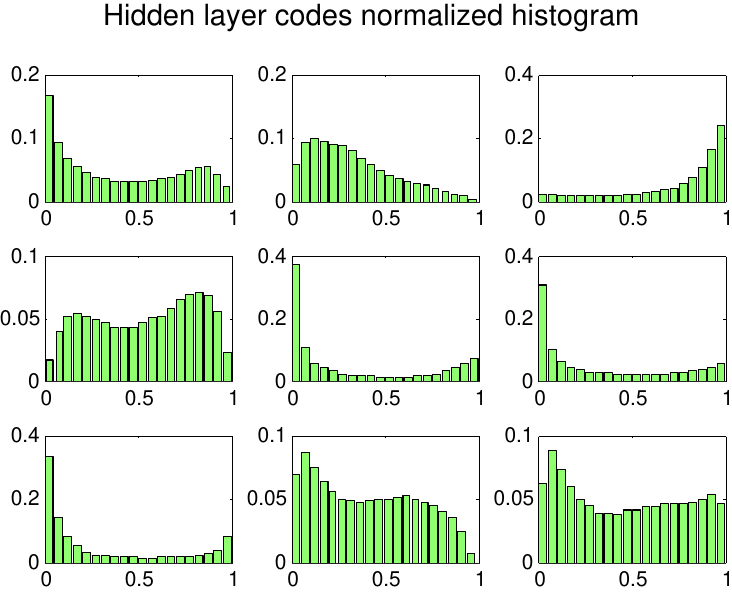}}
\subfigure[Iteration $6\times10^4$ ($K=2$)]{\includegraphics[width=.23\textwidth]{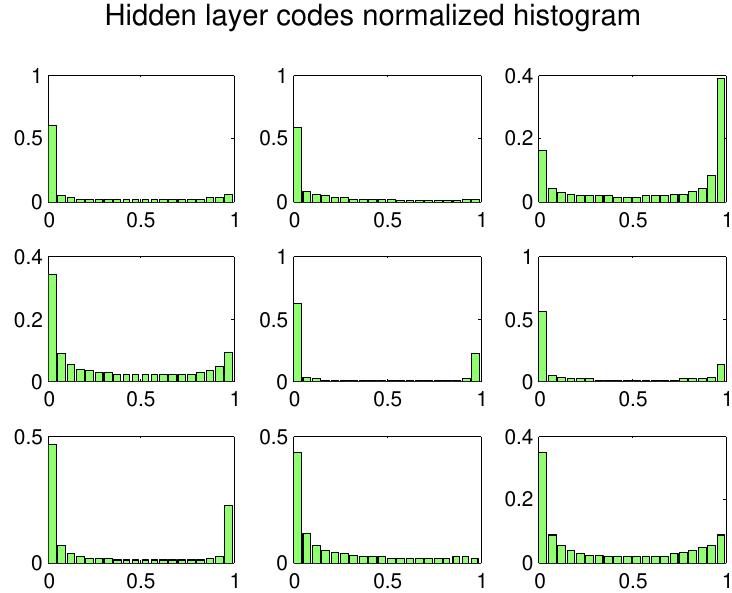}} & \\
\subfigure[Iteration $1$ ($K=36$)]{\includegraphics[width=.23\textwidth]{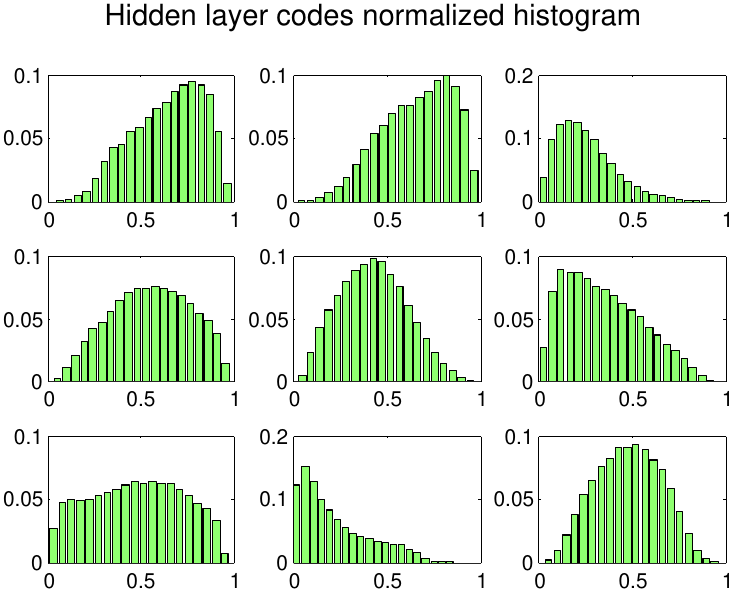}}
\subfigure[Iteration $1\times10^3$ ($K=36$)]{\includegraphics[width=.23\textwidth]{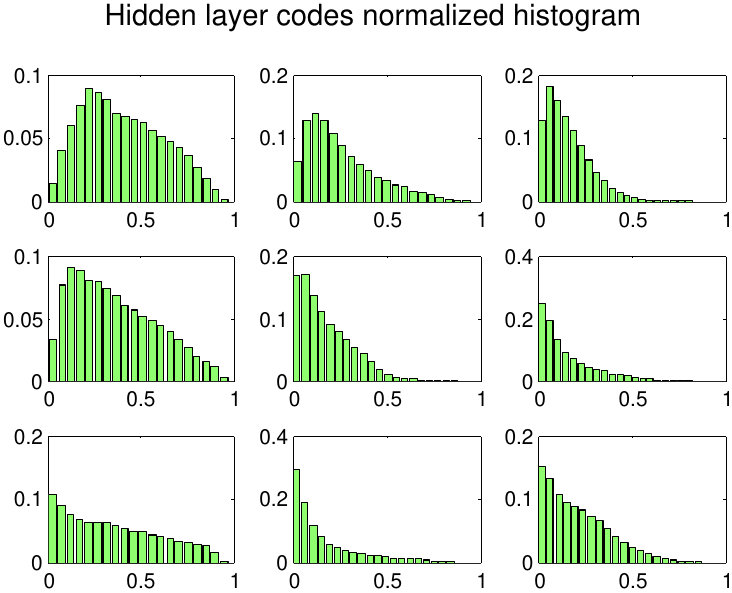}}
\subfigure[Iteration $1\times10^4$ ($K=36$)]{\includegraphics[width=.23\textwidth]{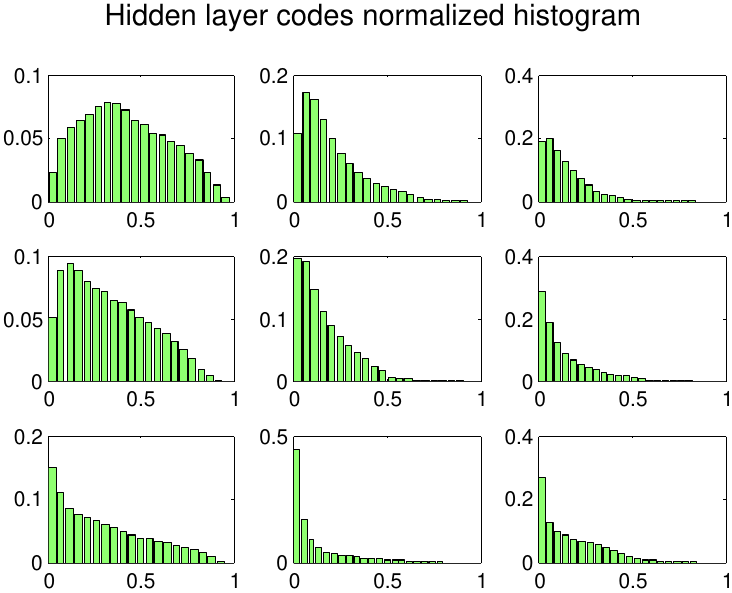}}
\subfigure[Iteration $6\times10^4$ ($K=36$)]{\includegraphics[width=.23\textwidth]{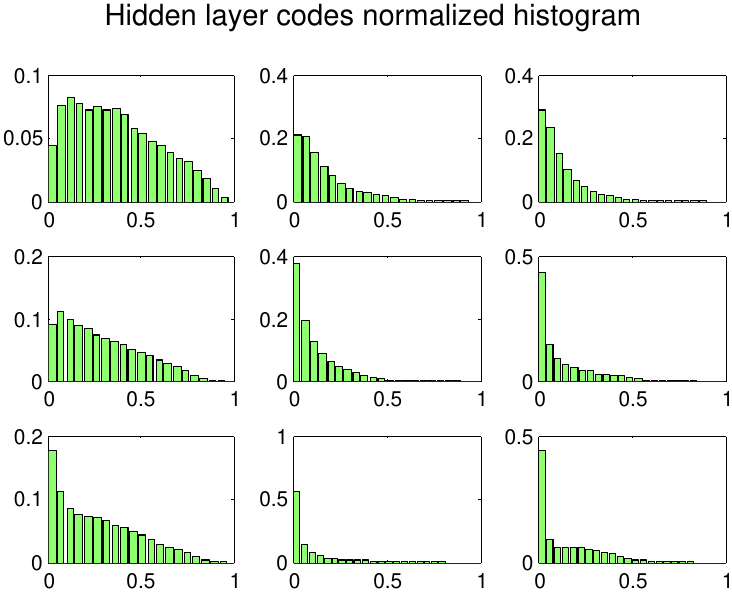}} & \\
\end{tabular}
\caption{The frequency histograms of activation values of $9$ randomly selected neurons in the first hidden layer of the encoder (i.e., $T_1$).  (a)-(d) demonstrate the histograms of these neurons at training iteration $1$, $1\times10^3$, $1\times10^4$ and $6\times10^4$ respectively when $K=2$, whereas (e)-(h) demonstrate the histograms at training iteration $1$, $1\times10^3$, $1\times10^4$ and $6\times10^4$ respectively when $K=36$. In each sub-figure, the $x$-axis denotes the bounded activation range (divided into $20$ bins of same length), whereas the $y$-axis denotes the frequency in each bin. For each neuron, a high frequency value in the leftmost or the rightmost bin indicates a high degree of saturation. More saturated neurons suggest a high possibility that the hidden layer codes occupy a broader space.\vspace{-0.0cm}}
\label{fig:histogram}
\end{figure*}

We expect a similar phenomenon to happen for other hidden layer representations. To verify this, we demonstrate the geometric distribution of $T_1$ for both $K=2$ and $K=36$ in Fig. \ref{fig:histogram}. Specifically, we randomly selected $9$ (out of $1000$) neurons and plot the normalized histograms (by frequency) of their activation values to infer the geometric distribution changes of $T_1$ in $\mathbb{R}^{1000}$. Intuitively, the broader the space the codes occupy (before activation), the higher the possibility of neuron saturation. In fact, from Fig.~\ref{fig:histogram}, almost all the neurons in $T_1$ tend to be saturated at the end of iteration when $K=2$. This suggests that the original hidden representations of $T_1$ persistently enlarge the projection space, just like what $T_4$ does. By contrast, except for a few neurons, there is no obvious saturation for other neurons of $T_1$ when $K=36$. Moreover, the normalized histograms remain almost the same from iteration $1\times10^4$ to $6\times10^4$. This suggests that the hidden representations of $T_1$ are self-constrained in $\mathbb{R}^{1000}$ - same as $T_4$.

Finally, it is worth noting that this behavior is expected to generalize to other DNNs, because the DPI is an intrinsic characteristic of any feedforward DNNs. Further work should validate this property in DNNs and make the proper modifications to analyze recurrent neural networks (RNNs).

\section{Conclusions} \label{conclusions}

In this paper, we analyzed DNNs learning from a joint geometric and information theoretic perspective, thus emphasizing the role that pair-wise mutual information plays in understanding DNNs. As an application of this idea, three fundamental properties are presented concentrating on stacked autoencoders (SAEs). The experiments on three real-world datasets validated the data processing inequality associated with layer-wise mutual information and the existence of bifurcation point associated with the topology of SAEs that is controlled by training data. Moreover, this indirectly corroborates the appropriateness of the non-parametric estimators that were used to apply the information theoretic understanding.

Our observations have some critical insights and implications for future research:

1) The potential of Information Theoretic Learning (ITL) \cite{principe2010information} in understanding DNNs.

Using information theory to explain DNNs remains a promising avenue, but there are still several important issues in the implementation. Among them is the accurate and tractable estimation of information quantities from large data. This is because Shannon's definition is hard to estimate, which severely limits its powers to analyze machine learning algorithms \cite{gao2015efficient}. For example, employing Shannon's discrete entropy, \cite{khadivi2016flow} limits the analysis to simple Boolean networks (discrete codes), whereas \cite{shwartz2017opening} still concentrates on a small toy datasets.

ITL~\cite{principe2010information}, on the other hand, utilizes Renyi's quadratic information measures \cite{renyi1961measures} and Parzen windowing \cite{parzen1962estimation} to estimate information quantities directly from continuous random variables with few assumptions. The recently proposed matrix formulation of Renyi's information \cite{giraldo2015measures} is a departure from the original quadratic information measures and allows estimation of high dimensional data. This useful property makes it well suited to analyze the dynamics or information flow of any deep neural networks, thus achieving the goal of explaining DNN mappings. Moreover, it is worth noting that, the proposed methodologies can be simply applied to other DNN architectures, e.g., CNN~\cite{yu2018understanding}. The only difference is that we need to use the multivariate extension of matrix-based Renyi's entropy functional~\cite{yu2018multivariate} to quantify the information flow of CNN.

However, as emphasized in previous sections, care must be taken to select an appropriate value for the kernel size $\sigma$. In this paper, $\sigma$ is defined by the Silverman's rule of thumb \cite{silverman1986density}:
\begin{equation}
\sigma=h\times n^{-1/(4+d)},
\end{equation}
where $n$ is the number of samples (mini-batch size of SGD training in our application), $d$ is the sample dimensionality (number of neurons for each layer in our application), $h$ is an empirical value selected experimentally by taking into account the data's average marginal variance. We understand that this density estimation perspective may not be the best to select the RKHS inner product, but its advantage of showing a dependence on dimension made it still effective. Theoretically, for small $\sigma$, the Gram matrix approaches identity and thus its eigenvalues become more similar, with $1/n$ as the limit case. Therefore, both entropy and mutual information monotonically increase as $\sigma\rightarrow0$ \cite{giraldo2015measures}. We select $h=6$, as the entropy estimated using $h=6$ matches well with the geometric distributional changes mentioned in section \ref{section4.1} (see Fig. \ref{fig:parameter}). However, we also show, in~\ref{appendix_C}, that even though $h$ (hence $\sigma$) is not optimized, we can still observe the same trends of general patterns of the curves in the IP, although the values of entropy change with the kernel size as expected. \ref{appendix_C} also shows a similar behavior with the estimation of mutual information that now depends upon two kernel sizes.

\begin{figure}[!htbp]
\centering
\begin{tabular}{ccc}
\includegraphics[width=.45\textwidth]{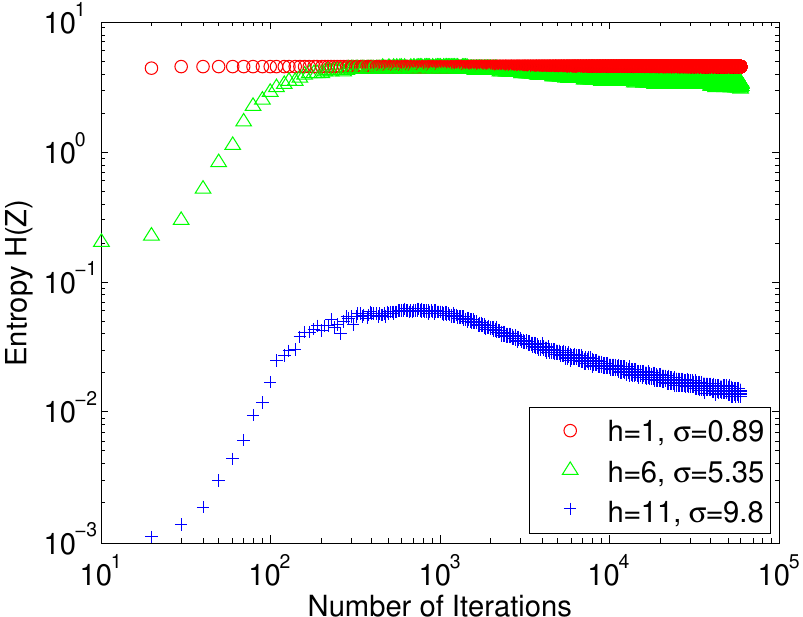}
\end{tabular}
\caption{The entropy $\mathbf{H}(Z)$ of bottleneck layer codes (in logarithmic scale) with respect to different values of $h$ and kernel size $\sigma$. As can be seen, $\mathbf{H}(Z)$ is monotonically increase as $h$ or $\sigma$ decreases. $h=1$ makes $\mathbf{H}(Z)$ quickly approaches to its upper bound, thus failing to inspect any distributional changes in the bottleneck layer codes. Although the entropy given by $h=11$ matches well with the general distribution's trend (the green blue curves show the same trends), a large $h$ makes $\mathbf{H}(Z)$ tend to $0$ ($0.01$ in our application) which is not convincing. By contrast, $h=6$ provides an entropy estimator using the full range and is discriminative of codes distributional changes.\vspace{-0.0cm}}
\label{fig:parameter}
\end{figure}

2) Implications on the design of DNNs topology.

The optimal design of DNNs topology is essential in many practical applications. Unfortunately, there is still a lack of fixed rules or widely acknowledged methods currently available. Previous works either employ a trial-and-error process starting from a set of rules of thumb or dynamically adjust the network configuration (e.g., the cascade-correlation algorithm \cite{fahlman1990cascade}).

With the advent of deep learning the tendency is to design much deeper neural networks to guarantee favorable performance on different tasks, especially for image classification \cite{he2016deep}. However, from the DPI perspective validated in this work (we also refer to another type of DPI specifically for MLP as shown in \cite{shwartz2017opening}), the deeper the neural networks, the more information about the input is lost, thus the less information the network can manipulate. In this sense, one can expect an upper bound on the number of layers in DNNs that achieves optimal performance. The advantage of the proposed methodology is that the experimentalist can find out how much residual information exists in the intermediate layers to guarantee a generalization performance. Alternatively, this may help ``principled tweaking" by replacing the nonlinear units by linear units (as proposed by \cite{glorot2011deep}) in specific layers that substantially reduce the mutual information. In fact, more layers will not only result in more information loss, but also will introduce much more parameters that are hard to be tuned and will compromise generalization.

As an application of our observed fundamental properties in SAEs, we conducted a simple out-of-sample experiment to validate the classification accuracy of bottleneck layer codes with respect to different number of hidden layers $K$ in the encoder or decoder and different bottleneck layer sizes $S$. To this end, we train a Softmax Regression classifier using the bottleneck layer codes obtained from different SAE architectures and evaluate the performance of learned classifier on unseen testing data. We repeat the procedure $10$ times with the same train-test split ratio and report the average classification accuracy in Table~\ref{Tab:summary_accuracy}. As can be seen, $S=5$ or $S=6$ is sufficient to guarantee a satisfactory classification accuracy, adding more layers cannot introduce any performance gain. Moreover, in the scenario of $S\geq5$, if the bottleneck layer size is equal to or larger than the intrinsic dimensionality of given data, there is no significant difference on the classification performance of bottleneck layer codes.
Therefore, the classification accuracy corroborates our fundamental properties described in previous sections.

\begin{table}[!hbpt]
\begin{center}
\caption{Summary of classification accuracy with respect to different SAE architectures.}\label{Tab:summary_accuracy}
\begin{tabular}{ccccccc}\hline
 & $K=2$ & $K=7$ & $K=13$ & $K=20$ & K=36 \\ \hline
$S=2$ & 22.23 & 69.57 & 83.30 & 85.78 & 88.39 \\
$S=3$ & 27.46 & 77.82 & 86.73 & 88.34 & 88.72   \\
$S=4$ & 29.46 & 82.01 & 87.85 & 88.52 & 88.50  \\
$S=5$ & 33.18 & 81.96 & 91.14 & 91.35 & 91.38  \\
$S=6$ & 35.94 & 85.37 & 91.04 & 91.64 & 91.61  \\
$S=7$ & 29.64 & 80.17 & 88.15 & 89.82 & 90.20  \\
$S=8$ & 21.35 & 76.07 & 86.65 & 86.93 & 86.12  \\ \hline
\end{tabular}
\end{center}
\end{table}

3) Implications on the feedforward training of DNNs.

The idea of training DNNs using information theoretic concepts has a long history dating back to the celebrated ``InfoMax" principle proposed by Linsker \cite{linsker1988self,linsker1989generate}, which states that the most informative learner is the one that maximizes the mutual information between input (e.g., sample attributes) and target (e.g., class label).  Motivated by the ``InfoMax" principle, several training methods have been developed concentrating on different types of network architectures, including MLP \cite{xu1999training}, Autoencoders \cite{miranda2013breaker}, Restricted Boltzmann Machine (RBM) \cite{peng2016mutual}, etc.

\begin{figure}[!htbp]
\centering
\begin{tabular}{ccc}
\subfigure[IP-I when $K=36$] {\includegraphics[width=.23\textwidth]{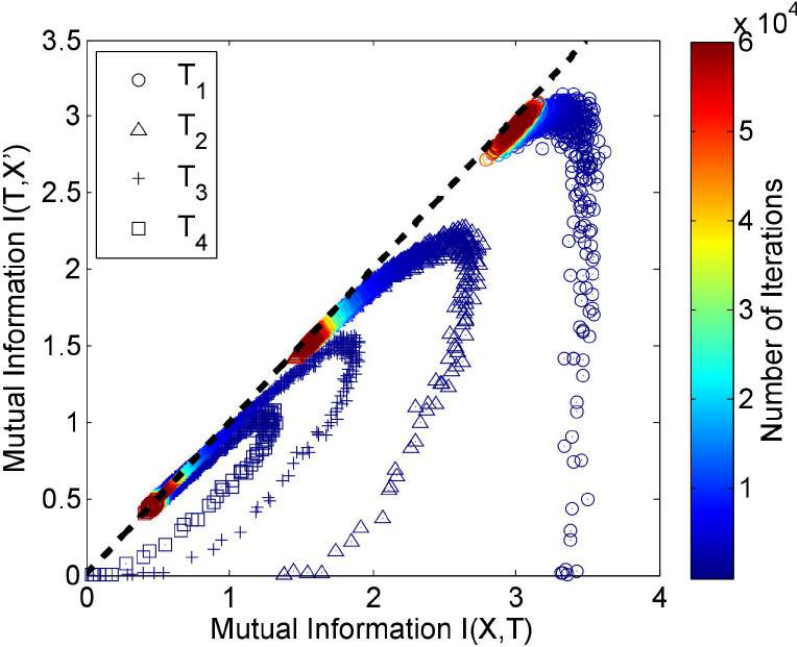}}
\subfigure[IP-II when $K=36$] {\includegraphics[width=.23\textwidth]{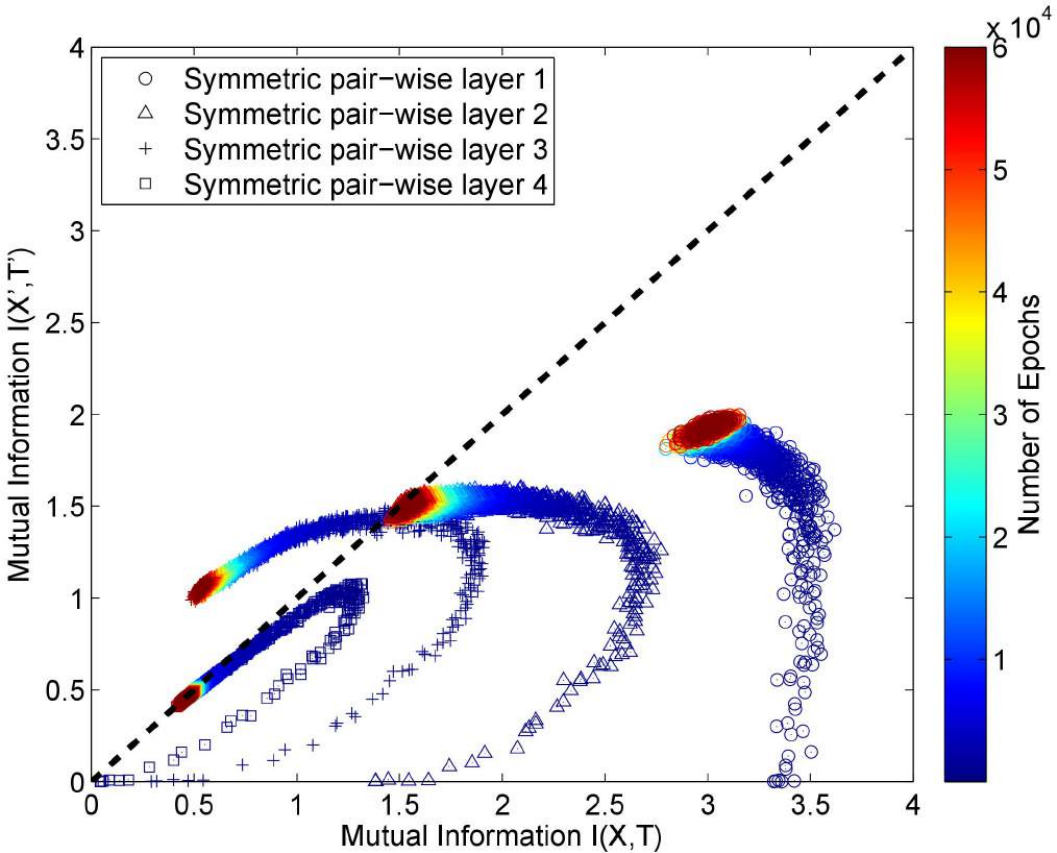}} \\
\subfigure[IP-III when $K=36$] {\includegraphics[width=.23\textwidth]{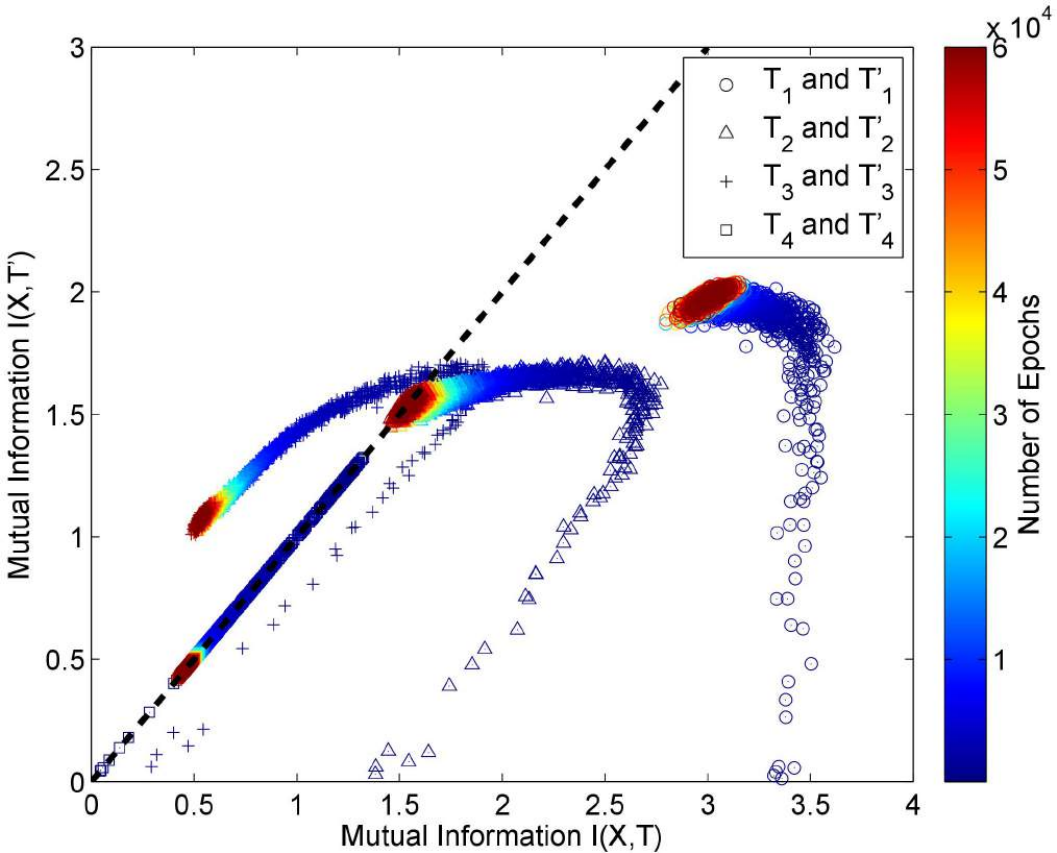}}
\subfigure[generalization curve] {\includegraphics[width=.23\textwidth]{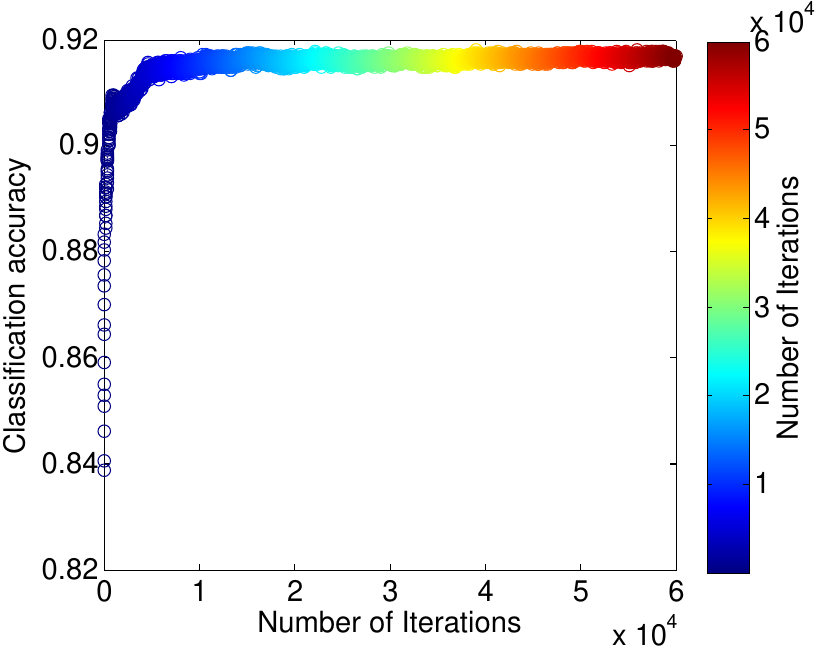}} &
\end{tabular}
\caption{The relationships between IP-I, IP-II, IP-III and the generalization performance in classification case. We consistently looking at the bottleneck layer codes $Z$ when the system is trained as a SAE and test the codes' accuracy in classification using a Softmax Regression classifier (without fine-tuning). (a), (b) and (c) demonstrate the IP-I (encoder module), IP-II and IP-III when $K=36$, where the black dashed line indicates the bisector. (d) demonstrates the corresponding classification accuracy using $Z$ with respect to the number of iterations. A fundamental difference between IP-I and IP-II is that $\mathbf{I}(X',T')$ can exceed $\mathbf{I}(X,T)$. Therefore, it is very probable that when IP-II goes beyond the bisector, that particular layer is overtrained. Moreover, it is interesting that for the classification case, the bottleneck layer code $Z$ that corresponds to the training phase close to the knee (in both IP-I and IP-II) is the point where classification accuracy on testing set is maximum.\vspace{-0.0cm}}
\label{fig:generalization}
\end{figure}

We believe the utilization of ITL and the IPs (e.g., Figs \ref{fig:generalization}(a) and \ref{fig:generalization}(b)) holds potential for feedforward training of DNNs, as an alternative to the basic backpropagation method. Our argument stems from three main reasons. First, ITL gives a tractable estimation of information quantities, which is critical to implement the ``InfoMax" principle. Note that the gradient of the mutual information becomes much less dependent on kernel size, because it is insensitive to the bias caused by the selection of the kernel size. Second, the IPs provide an explicit and flexible way to visualize information flow between any layer of interest. For example, by monitoring the information quantities $\mathbf{Q}$ between symmetric pair-wise layers (e.g., $\mathbf{Q}(X,T)$ and $\mathbf{Q}(X',T')$) in a greedy manner, \cite{bengio2014auto} suggests the possibility of using ``target propagation" to train SAEs. In this sense, the IP-II might be a good complement to implement this idea. In fact, by referring to Fig.~\ref{fig:generalization}(b), it is interesting to find that $\mathbf{I}(X',T')$ can exceed $\mathbf{I}(X,T)$. This is a fundamental difference to IP-I, in which all curves are strictly below the bisector of the IP. Moreover, by comparing Figs.~\ref{fig:generalization}(a) and \ref{fig:generalization}(b) with Fig.~\ref{fig:generalization}(d), it is interesting that for the classification case, the bottleneck layer code $Z$ that corresponds to the training phase close to the knee (in both IP-I and IP-II) is the point where classification accuracy on testing set is maximum. These results may provide an explicit cut-off point to ``early stopping" for optimal generalization \cite{haykin2009neural}.

4) Implications on optimal generalization.

We extend the above observations to the problem of generalization, i.e., the ability of the model (learned from training data) to fit unseen instances (or testing data) \cite{haykin2009neural}. This is perhaps the most challenging topic in DNNs, as it has been experimentally proven that popular techniques including explicit regularization (e.g., weight decay or Dropout \cite{srivastava2014dropout}) or implicit methods (e.g., early stopping or batch normalization \cite{ioffe2015batch}) cannot explain the generalization of DNNs very well \cite{zhang2016understanding}.

To the best of our knowledge, the analysis of generalization ability using information theoretic concepts has seldom been investigated before, except for some recent published works (e.g., \cite{raginsky2017information,alabdulmohsin2017information}). Different from these works, we present an alternative perspective herein. In fact, we experimentally found that the bottleneck layer code $Z$ that corresponds to the training phase (shown in the IPs) for the SAE is a stable indicator of the knee of the generalization performance when the codes $Z$ are used for classification using a Softmax Regression classifier (see Fig. \ref{fig:generalization}). If this preliminary observation extends to other cases, it may be possible to address the problem of generalization of a classifier using an ITL framework. We leave a rigorous implementation of this idea as future work.

5) A deeper insight on the role of information theoretic estimators.

As a last point we would like to comment that some researchers point out different behaviors of the curves in the IP using different types of mutual information estimators~\cite{shwartz2017opening,saxe2018on,noshad2018scalable}. We have to remember that not all the properties of the statistical definition of mutual information are transferred to the estimators of mutual information~\cite{paninski2003estimation}. Therefore different types of estimators may lead to different behavior of the curves. Again, care must be taken to properly select the type and hyperparameters of the estimators, as well as to verify if the expected behavior of the statistical models (DPIs) are verified in practice with the selected hyperparamters. Otherwise, spurious conclusions may happen.

Recall the experimental discrepancy reported by Shwartz-Ziv~$et$ $al.$~\cite{shwartz2017opening} and Saxe~$et$ $al.$~\cite{saxe2018on}. According to~\cite{saxe2018on}, the existence of compression phase observed in~\cite{shwartz2017opening} depends on the adopted nonlinearity functions: double-sided saturating nonlinearities like tanh or sigmoid yield a compression phase, but linear activation functions and single-sided saturating nonlinearities like the ReLU do not. We argue that it is may not simply be the matter of nonlinearity functions, but rather the used mutual information estimators. Shwartz-Ziv~$et$ $al.$ use the basic mutual information definition and estimate the mutual information values just by dividing neuron activation values into $30$ equal-interval bins, whereas Saxe~$et$ $al.$ use Kernel Density Estimator (KDE)~\cite{kolchinsky2017estimating} and KNN based estimator~\cite{kraskov2004estimating}. Both papers estimate $\mathbf{I}(X;\mathrm{Bin}(T))$ rather than $\mathbf{I}(X;T)$, where $\mathrm{Bin}$ denotes a discretization of activation values in $T$ into a user-selected number of bins. Moreover, one should note that although KNN based estimator or KDE is more data efficient, the adaptive bin size will jeopardize the shape of PDF, thus makes the estimator deviates from true mutual information values. In fact, the base estimator used in ~\cite{saxe2018on} just provides KDE-based lower and upper bounds on the true mutual information~\cite{kolchinsky2017estimating}. Interestingly, a recent paper by Noshad $et$ $al.$~\cite{noshad2018scalable} suggested using dependence graphs to estimate true mutual information values and observed the compression phase even using ReLU activation functions.

\begin{figure}[!htbp]
\centering
\begin{tabular}{ccc}
\subfigure[IP-I when $K=2$, two layers with ReLU] {\includegraphics[width=.23\textwidth]{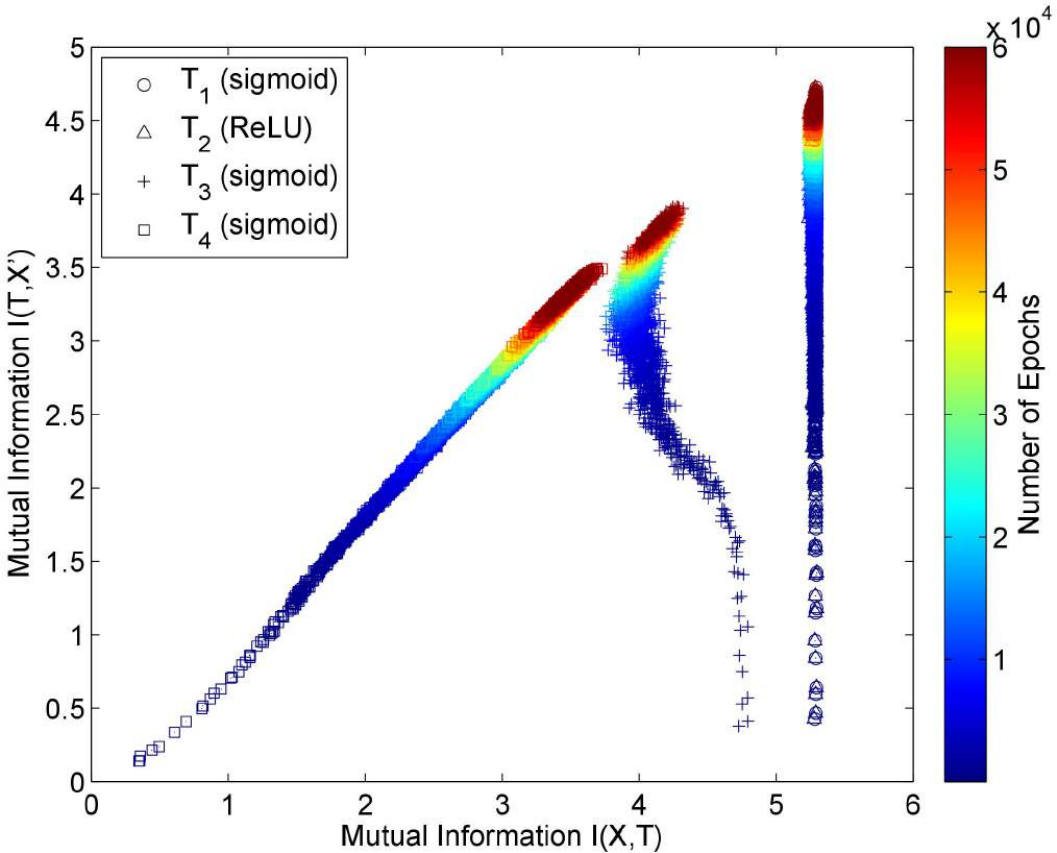}}
\subfigure[IP-I when $K=2$, four layers with ReLU] {\includegraphics[width=.23\textwidth]{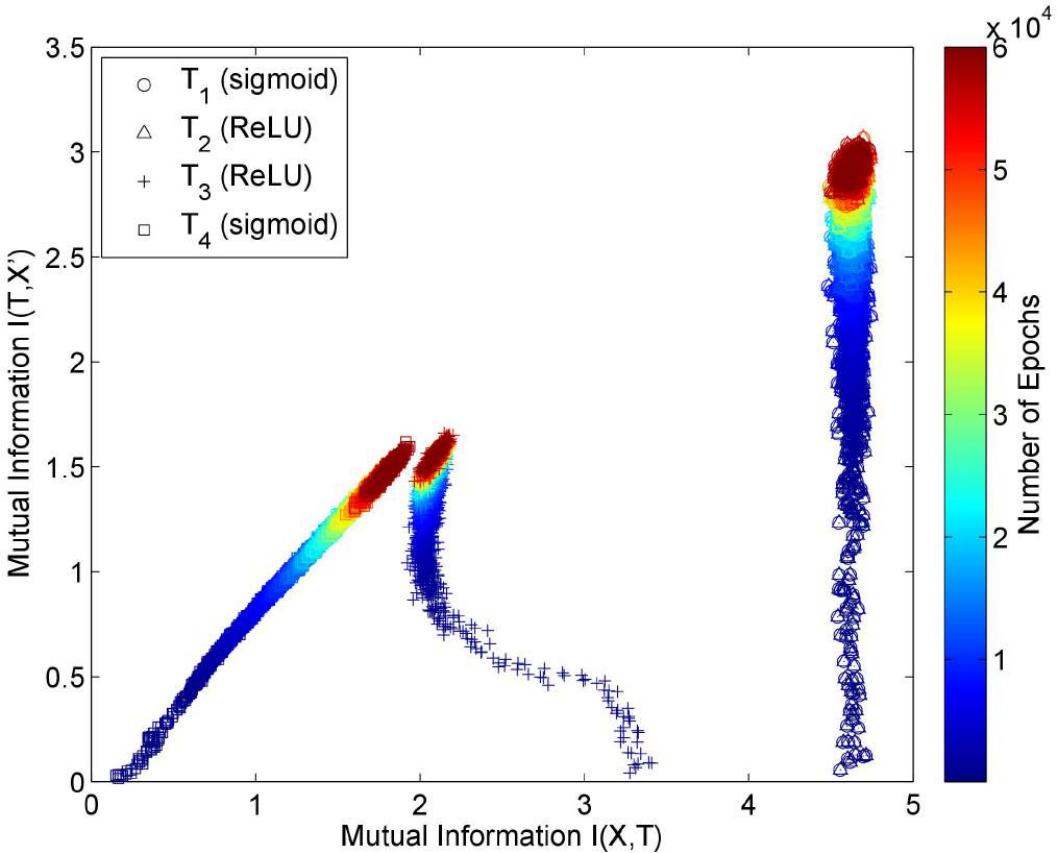}} \\
\subfigure[IP-I when $K=36$, two layers with ReLU] {\includegraphics[width=.23\textwidth]{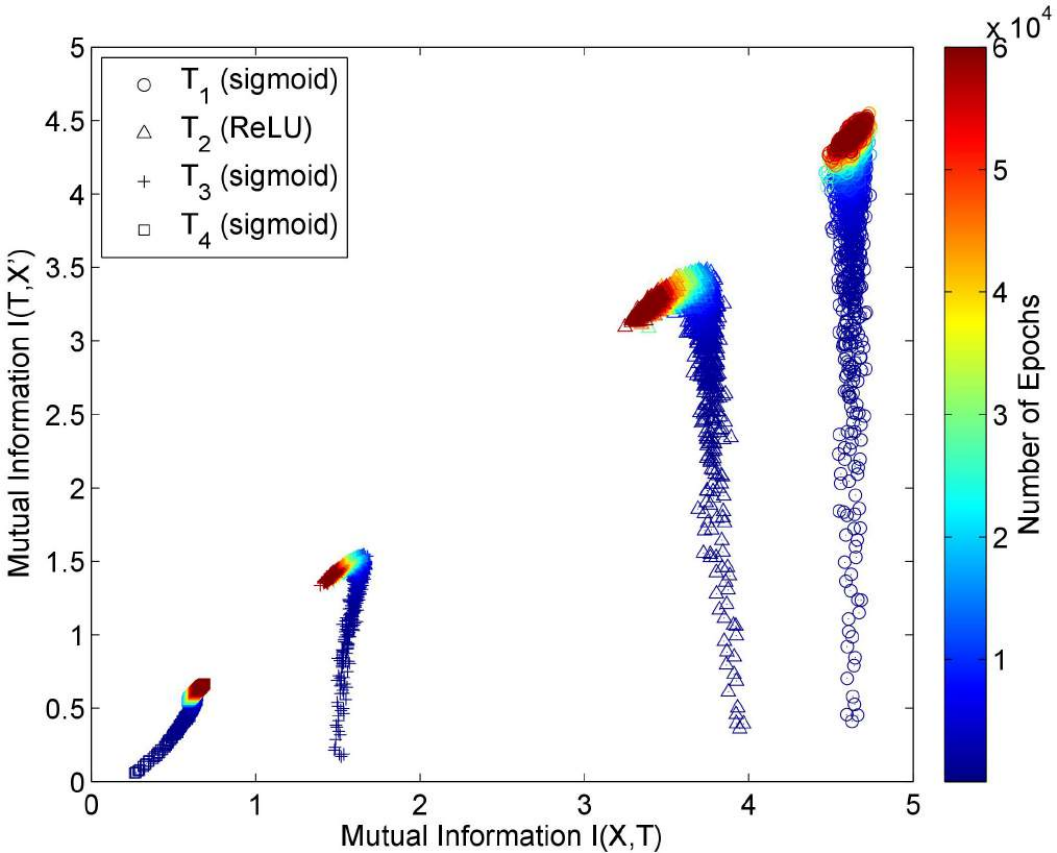}}
\subfigure[IP-I when $K=36$, four layers with ReLU] {\includegraphics[width=.23\textwidth]{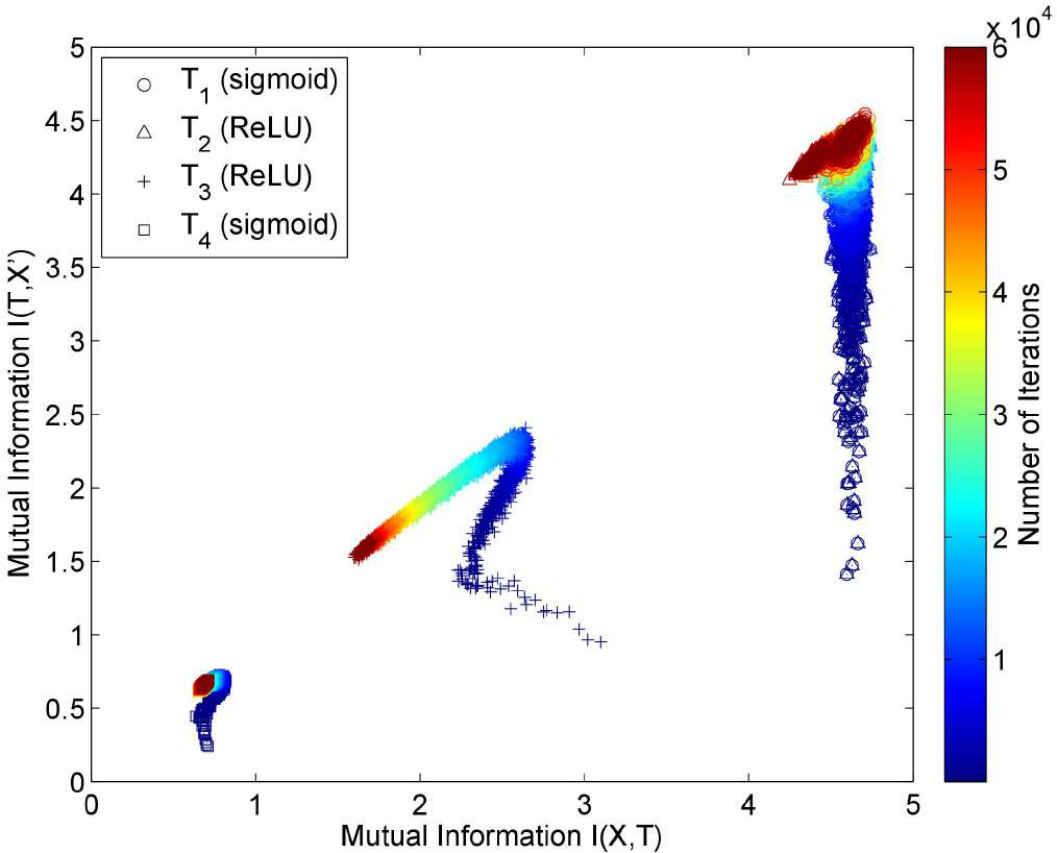}} &
\end{tabular}
\caption{IP-I (encoder part) of SAEs with different setups: (a) $K=2$, $T_2$ uses ReLU activation functions; (b) $K=2$, $T_2$ and $T_3$ use ReLU activation functions; (c) $K=36$, $T_2$ uses ReLU activation functions; (d) $K=36$, $T_2$ and $T_3$ use ReLU activation functions. Curves in (c) and (d) demonstrated obvious ``compression" phase compared with those in (a) and (b).\vspace{-0.0cm}}
\label{fig:relu}
\end{figure}

Regarding our suggested \textbf{Fundamental Property III} that links the different behaviors of curves in the IP with respect to the bottleneck layer size of SAE. Although our results reported in Section~\ref{experiments} is based on Sigmoid activation functions. We can also observe the same phenomena with ReLU as shown in Fig.~\ref{fig:relu}. Of course, the amount of compressions is different for ReLU and Sigmoid.

\section*{Acknowledgement}
The authors would like to express their sincere gratitude to Dr. Luis~Gonzalo~S\'{a}nchez~Giraldo from the University of Miami and Dr. Robert Jenssen from the UiT - The Arctic University of Norway for their careful reading of our manuscript and many insightful comments and suggestions. The authors also thank the anonymous reviewers for their very helpful suggestions, which led to substantial improvements of the paper. This work is supported in part by the U.S. Office of Naval Research under Grant N$00014$-$15$-$1$-$2103$ and N$00014$-$18$-$1$-$2306$.


\section*{References}

\bibliographystyle{elsarticle-num}
\bibliography{NN_manuscript}

\begin{thebibliography}{10}
\expandafter\ifx\csname url\endcsname\relax
  \def\url#1{\texttt{#1}}\fi
\expandafter\ifx\csname urlprefix\endcsname\relax\def\urlprefix{URL }\fi
\expandafter\ifx\csname href\endcsname\relax
  \def\href#1#2{#2} \def\path#1{#1}\fi

\bibitem{krizhevsky2012imagenet}
A.~Krizhevsky, I.~Sutskever, G.~E. Hinton, Imagenet classification with deep
  convolutional neural networks, in: Advances in neural information processing
  systems, 2012, pp. 1097--1105.

\bibitem{graves2013speech}
A.~Graves, A.-r. Mohamed, G.~Hinton, Speech recognition with deep recurrent
  neural networks, in: Acoustics, speech and signal processing (icassp), 2013
  ieee international conference on, IEEE, 2013, pp. 6645--6649.

\bibitem{mesnil2013investigation}
G.~Mesnil, X.~He, L.~Deng, Y.~Bengio, Investigation of recurrent-neural-network
  architectures and learning methods for spoken language understanding., in:
  Interspeech, 2013, pp. 3771--3775.

\bibitem{alain2016understanding}
G.~Alain, Y.~Bengio, Understanding intermediate layers using linear classifier
  probes, arXiv preprint arXiv:1610.01644.

\bibitem{principe2000neural}
J.~C. Principe, N.~R. Euliano, W.~C. Lefebvre, Neural and adaptive systems:
  fundamentals through simulations, Vol. 672, Wiley New York, 2000.

\bibitem{light1992ridge}
W.~Light, Ridge functions, sigmoidal functions and neural networks,
  Approximation theory VII (1992) 163--206.

\bibitem{minsky2017perceptrons}
M.~Minsky, S.~A. Papert, Perceptrons: An introduction to computational
  geometry, 2017.

\bibitem{principe2015universal}
J.~C. Principe, B.~Chen, Universal approximation with convex optimization:
  Gimmick or reality?[discussion forum], IEEE Computational Intelligence
  Magazine 10~(2) (2015) 68--77.

\bibitem{kolmogorov1939interpolation}
A.~N. Kolmogorov, Sur l'interpolation et extrapolation des suites
  stationnaires, CR Acad. Sci 208 (1939) 2043--2045.

\bibitem{principe2010information}
J.~C. Principe, Information theoretic learning: Renyi's entropy and kernel
  perspectives, Springer Science \& Business Media, 2010.

\bibitem{shwartz2017opening}
R.~Shwartz-Ziv, N.~Tishby, Opening the black box of deep neural networks via
  information, arXiv preprint arXiv:1703.00810.

\bibitem{mackay2003information}
D.~J. MacKay, Information theory, inference and learning algorithms, Cambridge
  university press, 2003.

\bibitem{stratonovich1965value}
R.~L. Stratonovich, On value of information, Izvestiya of USSR Academy of
  Sciences, Technical Cybernetics 5~(1) (1965) 3--12.

\bibitem{yu2017autoencoders}
S.~Yu, M.~Emigh, E.~Santana, J.~C. Pr{\'\i}ncipe, Autoencoders trained with
  relevant information: Blending shannon and wiener's perspectives, in:
  Acoustics, Speech and Signal Processing (ICASSP), 2017 IEEE International
  Conference on, IEEE, 2017, pp. 6115--6119.

\bibitem{giraldo2015measures}
L.~G.~S. Giraldo, M.~Rao, J.~C. Principe, Measures of entropy from data using
  infinitely divisible kernels, IEEE Transactions on Information Theory 61~(1)
  (2015) 535--548.

\bibitem{yu2018understanding}
S.~Yu, K.~Wickstr{\o}m, R.~Jenssen, J.~C. Principe, Understanding convolutional
  neural networks with information theory: An initial exploration, arXiv
  preprint arXiv:1804.06537.

\bibitem{haykin2014adaptive}
S.~S. Haykin, Adaptive filter theory (5th edition), Pearson Education, 2014.

\bibitem{stigler1981gauss}
S.~M. Stigler, Gauss and the invention of least squares, The Annals of
  Statistics (1981) 465--474.

\bibitem{chen2013system}
B.~Chen, Y.~Zhu, J.~Hu, J.~C. Principe, System parameter identification:
  information criteria and algorithms, Newnes, 2013.

\bibitem{kokiopoulou2011trace}
E.~Kokiopoulou, J.~Chen, Y.~Saad, Trace optimization and eigenproblems in
  dimension reduction methods, Numerical Linear Algebra with Applications
  18~(3) (2011) 565--602.

\bibitem{baldi1989neural}
P.~Baldi, K.~Hornik, Neural networks and principal component analysis: Learning
  from examples without local minima, Neural networks 2~(1) (1989) 53--58.

\bibitem{renyi1961measures}
A.~R{\'e}nyi, et~al., On measures of entropy and information, in: Proceedings
  of the Fourth Berkeley Symposium on Mathematical Statistics and Probability,
  Volume 1: Contributions to the Theory of Statistics, The Regents of the
  University of California, 1961.

\bibitem{silverman1986density}
B.~W. Silverman, Density estimation for statistics and data analysis, Vol.~26,
  CRC press, 1986.

\bibitem{giraldo2013rate}
L.~G.~S. Giraldo, J.~C. Principe, Rate-distortion auto-encoders, arXiv preprint
  arXiv:1312.7381.

\bibitem{huang2016flow}
C.-W. Huang, S.~S.~S. Narayanan, Flow of renyi information in deep neural
  networks, in: Machine Learning for Signal Processing (MLSP), 2016 IEEE 26th
  International Workshop on, IEEE, 2016, pp. 1--6.

\bibitem{alvarez2017kernel}
A.~M. {\'A}lvarez-Meza, J.~A. Lee, M.~Verleysen, G.~Castellanos-Dominguez,
  Kernel-based dimensionality reduction using renyi's $\alpha$-entropy measures
  of similarity, Neurocomputing 222 (2017) 36--46.

\bibitem{bhatia2006infinitely}
R.~Bhatia, Infinitely divisible matrices, The American Mathematical Monthly
  113~(3) (2006) 221--235.

\bibitem{mehta2014exact}
P.~Mehta, D.~J. Schwab, An exact mapping between the variational
  renormalization group and deep learning, arXiv preprint arXiv:1410.3831.

\bibitem{lin2017does}
H.~W. Lin, M.~Tegmark, D.~Rolnick, Why does deep and cheap learning work so
  well?, Journal of Statistical Physics 168~(6) (2017) 1223--1247.

\bibitem{tishby2015deep}
N.~Tishby, N.~Zaslavsky, Deep learning and the information bottleneck
  principle, in: Information Theory Workshop (ITW), 2015 IEEE, IEEE, 2015, pp.
  1--5.

\bibitem{tishby2000information}
N.~Tishby, F.~C. Pereira, W.~Bialek, The information bottleneck method, arXiv
  preprint physics/0004057.

\bibitem{khadivi2016flow}
P.~Khadivi, R.~Tandon, N.~Ramakrishnan, Flow of information in feed-forward
  deep neural networks, arXiv preprint arXiv:1603.06220.

\bibitem{burgess2018understanding}
C.~P. Burgess, I.~Higgins, A.~Pal, L.~Matthey, N.~Watters, G.~Desjardins,
  A.~Lerchner, Understanding disentangling in $\beta$-vae, arXiv preprint
  arXiv:1804.03599.

\bibitem{higgins2016beta}
I.~Higgins, L.~Matthey, A.~Pal, C.~Burgess, X.~Glorot, M.~Botvinick,
  S.~Mohamed, A.~Lerchner, beta-vae: Learning basic visual concepts with a
  constrained variational framework, in: International Conference on Learning
  Representations, 2016.

\bibitem{o2017introduction}
T.~O'Shea, J.~Hoydis, An introduction to deep learning for the physical layer,
  IEEE Transactions on Cognitive Communications and Networking 3~(4) (2017)
  563--575.

\bibitem{saxe2018on}
A.~M. Saxe, Y.~Bansal, J.~Dapello, M.~Advani, A.~Kolchinsky, B.~D. Tracey,
  D.~D. Cox, On the information bottleneck theory of deep learning, in:
  International Conference on Learning Representations, 2018.

\bibitem{bengio2013better}
Y.~Bengio, G.~Mesnil, Y.~Dauphin, S.~Rifai, Better mixing via deep
  representations, in: Proceedings of the 30th International Conference on
  Machine Learning (ICML-13), 2013, pp. 552--560.

\bibitem{brahma2016deep}
P.~P. Brahma, D.~Wu, Y.~She, Why deep learning works: A manifold
  disentanglement perspective, IEEE transactions on neural networks and
  learning systems 27~(10) (2016) 1997--2008.

\bibitem{pascanu2013number}
R.~Pascanu, G.~Montufar, Y.~Bengio, On the number of response regions of deep
  feed forward networks with piece-wise linear activations, arXiv preprint
  arXiv:1312.6098.

\bibitem{montufar2014number}
G.~F. Montufar, R.~Pascanu, K.~Cho, Y.~Bengio, On the number of linear regions
  of deep neural networks, in: Advances in neural information processing
  systems, 2014, pp. 2924--2932.

\bibitem{achille2017emergence}
A.~Achille, S.~Soatto, Emergence of invariance and disentangling in deep
  representations, arXiv preprint arXiv:1706.01350.

\bibitem{zeiler2014visualizing}
M.~D. Zeiler, R.~Fergus, Visualizing and understanding convolutional networks,
  in: European conference on computer vision, Springer, 2014, pp. 818--833.

\bibitem{zeiler2011adaptive}
M.~D. Zeiler, G.~W. Taylor, R.~Fergus, Adaptive deconvolutional networks for
  mid and high level feature learning, in: Computer Vision (ICCV), 2011 IEEE
  International Conference on, IEEE, 2011, pp. 2018--2025.

\bibitem{mahendran2015understanding}
A.~Mahendran, A.~Vedaldi, Understanding deep image representations by inverting
  them, in: Proceedings of the IEEE conference on computer vision and pattern
  recognition, 2015, pp. 5188--5196.

\bibitem{yosinski2015understanding}
J.~Yosinski, J.~Clune, A.~Nguyen, T.~Fuchs, H.~Lipson, Understanding neural
  networks through deep visualization, arXiv preprint arXiv:1506.06579.

\bibitem{nguyen2016multifaceted}
A.~Nguyen, J.~Yosinski, J.~Clune, Multifaceted feature visualization:
  Uncovering the different types of features learned by each neuron in deep
  neural networks, arXiv preprint arXiv:1602.03616.

\bibitem{zhang2017interpreting}
Q.~Zhang, R.~Cao, F.~Shi, Y.~N. Wu, S.-C. Zhu, Interpreting cnn knowledge via
  an explanatory graph, arXiv preprint arXiv:1708.01785.

\bibitem{bach2015pixel}
S.~Bach, A.~Binder, G.~Montavon, F.~Klauschen, K.-R. M{\"u}ller, W.~Samek, On
  pixel-wise explanations for non-linear classifier decisions by layer-wise
  relevance propagation, PloS one 10~(7) (2015) e0130140.

\bibitem{montavon2017explaining}
G.~Montavon, S.~Lapuschkin, A.~Binder, W.~Samek, K.-R. M{\"u}ller, Explaining
  nonlinear classification decisions with deep taylor decomposition, Pattern
  Recognition 65 (2017) 211--222.

\bibitem{samek2017evaluating}
W.~Samek, A.~Binder, G.~Montavon, S.~Lapuschkin, K.-R. M{\"u}ller, Evaluating
  the visualization of what a deep neural network has learned, IEEE
  transactions on neural networks and learning systems 28~(11) (2017)
  2660--2673.

\bibitem{kuroe1993learning}
Y.~Kuroe, Y.~Nakai, T.~Mori, A learning method of nonlinear mappings by neural
  networks with considering their derivatives, in: Neural Networks, 1993.
  IJCNN'93-Nagoya. Proceedings of 1993 International Joint Conference on,
  Vol.~1, IEEE, 1993, pp. 528--531.

\bibitem{krawczak2013multilayer}
M.~Krawczak, Multilayer neural networks: a generalized net perspective, Vol.
  478, Springer, 2013.

\bibitem{luttrell1994bayesian}
S.~P. Luttrell, A bayesian analysis of self-organizing maps, Neural Computation
  6~(5) (1994) 767--794.

\bibitem{norris1998markov}
J.~R. Norris, Markov chains, no.~2, Cambridge university press, 1998.

\bibitem{liggett2012interacting}
T.~M. Liggett, Interacting particle systems, Vol. 276, Springer Science \&
  Business Media, 2012.

\bibitem{jansen2014notion}
S.~Jansen, N.~Kurt, et~al., On the notion (s) of duality for markov processes,
  Probability surveys 11 (2014) 59--120.

\bibitem{csiszar1972class}
I.~Csisz{\'a}r, A class of measures of informativity of observation channels,
  Periodica Mathematica Hungarica 2~(1-4) (1972) 191--213.

\bibitem{cover2012elements}
T.~M. Cover, J.~A. Thomas, Elements of information theory, John Wiley \& Sons,
  2012.

\bibitem{merhav2011data}
N.~Merhav, Data processing theorems and the second law of thermodynamics, IEEE
  Transactions on Information Theory 57~(8) (2011) 4926--4939.

\bibitem{haykin1994neural}
S.~Haykin, Neural networks: a comprehensive foundation, Prentice Hall PTR,
  1994.

\bibitem{bengio2014auto}
Y.~Bengio, How auto-encoders could provide credit assignment in deep networks
  via target propagation, arXiv preprint arXiv:1407.7906.

\bibitem{takens1981detecting}
F.~Takens, et~al., Detecting strange attractors in turbulence, Lecture notes in
  mathematics 898~(1) (1981) 366--381.

\bibitem{yap2011stable}
H.~L. Yap, C.~J. Rozell, Stable takens' embeddings for linear dynamical
  systems, IEEE Transactions on Signal Processing 59~(10) (2011) 4781--4794.

\bibitem{potapov2002neural}
A.~Potapov, M.~Ali, Neural networks for estimating intrinsic dimension,
  Physical Review E 65~(4) (2002) 046212.

\bibitem{lecun1998gradient}
Y.~LeCun, L.~Bottou, Y.~Bengio, P.~Haffner, Gradient-based learning applied to
  document recognition, Proceedings of the IEEE 86~(11) (1998) 2278--2324.

\bibitem{xiao2017fashion}
H.~Xiao, K.~Rasul, R.~Vollgraf, Fashion-mnist: a novel image dataset for
  benchmarking machine learning algorithms, arXiv preprint arXiv:1708.07747.

\bibitem{aneja2016modeling}
D.~Aneja, A.~Colburn, G.~Faigin, L.~Shapiro, B.~Mones, Modeling stylized
  character expressions via deep learning, in: Asian Conference on Computer
  Vision, Springer, 2016, pp. 136--153.

\bibitem{maaten2008visualizing}
L.~v.~d. Maaten, G.~Hinton, Visualizing data using t-sne, Journal of Machine
  Learning Research 9~(Nov) (2008) 2579--2605.

\bibitem{arpit2016regularized}
D.~Arpit, Y.~Zhou, H.~Ngo, V.~Govindaraju, Why regularized auto-encoders learn
  sparse representation?, in: International Conference on Machine Learning,
  2016, pp. 136--144.

\bibitem{hinton2006reducing}
G.~E. Hinton, R.~R. Salakhutdinov, Reducing the dimensionality of data with
  neural networks, science 313~(5786) (2006) 504--507.

\bibitem{camastra2016intrinsic}
F.~Camastra, A.~Staiano, Intrinsic dimension estimation: Advances and open
  problems, Information Sciences 328 (2016) 26--41.

\bibitem{wang2008scale}
X.~Wang, J.~S. Marron, et~al., A scale-based approach to finding effective
  dimensionality in manifold learning, Electronic Journal of Statistics 2
  (2008) 127--148.

\bibitem{levina2005maximum}
E.~Levina, P.~J. Bickel, Maximum likelihood estimation of intrinsic dimension,
  in: Advances in neural information processing systems, 2005, pp. 777--784.

\bibitem{lombardi2011minimum}
G.~Lombardi, A.~Rozza, C.~Ceruti, E.~Casiraghi, P.~Campadelli, Minimum neighbor
  distance estimators of intrinsic dimension, Machine Learning and Knowledge
  Discovery in Databases (2011) 374--389.

\bibitem{ceruti2014danco}
C.~Ceruti, S.~Bassis, A.~Rozza, G.~Lombardi, E.~Casiraghi, P.~Campadelli,
  Danco: An intrinsic dimensionality estimator exploiting angle and norm
  concentration, Pattern recognition 47~(8) (2014) 2569--2581.

\bibitem{gao2015efficient}
S.~Gao, G.~Ver~Steeg, A.~Galstyan, Efficient estimation of mutual information
  for strongly dependent variables, in: Artificial Intelligence and Statistics,
  2015, pp. 277--286.

\bibitem{parzen1962estimation}
E.~Parzen, On estimation of a probability density function and mode, The annals
  of mathematical statistics 33~(3) (1962) 1065--1076.

\bibitem{yu2018multivariate}
S.~Yu, L.~G.~S. Giraldo, R.~Jenssen, J.~C. Principe, Multivariate extension of
  matrix-based renyi's $\alpha$-order entropy functional, arXiv preprint
  arXiv:1808.07912.

\bibitem{fahlman1990cascade}
S.~E. Fahlman, C.~Lebiere, The cascade-correlation learning architecture, in:
  Advances in neural information processing systems, 1990, pp. 524--532.

\bibitem{he2016deep}
K.~He, X.~Zhang, S.~Ren, J.~Sun, Deep residual learning for image recognition,
  in: Proceedings of the IEEE conference on computer vision and pattern
  recognition, 2016, pp. 770--778.

\bibitem{glorot2011deep}
X.~Glorot, A.~Bordes, Y.~Bengio, Deep sparse rectifier neural networks, in:
  Proceedings of the Fourteenth International Conference on Artificial
  Intelligence and Statistics, 2011, pp. 315--323.

\bibitem{linsker1988self}
R.~Linsker, Self-organization in a perceptual network, Computer 21~(3) (1988)
  105--117.

\bibitem{linsker1989generate}
R.~Linsker, How to generate ordered maps by maximizing the mutual information
  between input and output signals, Neural computation 1~(3) (1989) 402--411.

\bibitem{xu1999training}
D.~Xu, J.~C. Principe, Training mlps layer-by-layer with the information
  potential, in: Neural Networks, 1999. IJCNN'99. International Joint
  Conference on, Vol.~3, IEEE, 1999, pp. 1716--1720.

\bibitem{miranda2013breaker}
V.~Miranda, J.~Krstulovic, J.~Hora, V.~Palma, J.~C. Principe, Breaker status
  uncovered by autoencoders under unsupervised maximum mutual information
  training, in: 17th International Conference on Intelligent System
  Applications to Power Systems, 2013.

\bibitem{peng2016mutual}
K.-H. Peng, H.~Zhang, Mutual information-based rbm neural networks, in: Pattern
  Recognition (ICPR), 2016 23rd International Conference on, IEEE, 2016, pp.
  2458--2463.

\bibitem{haykin2009neural}
S.~S. Haykin, Neural networks and learning machines, Vol.~3, Pearson Upper
  Saddle River, NJ, USA:, 2009.

\bibitem{srivastava2014dropout}
N.~Srivastava, G.~E. Hinton, A.~Krizhevsky, I.~Sutskever, R.~Salakhutdinov,
  Dropout: a simple way to prevent neural networks from overfitting., Journal
  of machine learning research 15~(1) (2014) 1929--1958.

\bibitem{ioffe2015batch}
S.~Ioffe, C.~Szegedy, Batch normalization: Accelerating deep network training
  by reducing internal covariate shift, in: International Conference on Machine
  Learning, 2015, pp. 448--456.

\bibitem{zhang2016understanding}
C.~Zhang, S.~Bengio, M.~Hardt, B.~Recht, O.~Vinyals, Understanding deep
  learning requires rethinking generalization, arXiv preprint arXiv:1611.03530.

\bibitem{raginsky2017information}
M.~Raginsky, A.~Xu, Information-theoretic analysis of generalization capability
  of learning algorithms, in: Advances in Neural Information Processing
  Systems, 2017, pp. 2521--2530.

\bibitem{alabdulmohsin2017information}
I.~Alabdulmohsin, An information-theoretic route from generalization in
  expectation to generalization in probability, in: Artificial Intelligence and
  Statistics, 2017, pp. 92--100.

\bibitem{noshad2018scalable}
M.~Noshad, Y.~Zeng, A.~O. Hero~III, Scalable mutual information estimation
  using dependence graphs, arXiv preprint arXiv:1801.09125.

\bibitem{paninski2003estimation}
L.~Paninski, Estimation of entropy and mutual information, Neural computation
  15~(6) (2003) 1191--1253.

\bibitem{kolchinsky2017estimating}
A.~Kolchinsky, B.~D. Tracey, Estimating mixture entropy with pairwise
  distances, Entropy 19~(7) (2017) 361.

\bibitem{kraskov2004estimating}
A.~Kraskov, H.~St{\"o}gbauer, P.~Grassberger, Estimating mutual information,
  Physical review E 69~(6) (2004) 066138.

\end{thebibliography}

\newpage
\appendix
\section{Property validation on Fashion-MNIST} \label{appendix_A}

\subsection{Validation of Fundamental Properties I $\&$ III}
\begin{figure*}[!htbp]
\centering
\begin{tabular}{ccc}
\subfigure[IP-I (encoder part) when $K=2$] {\includegraphics[width=.23\textwidth]{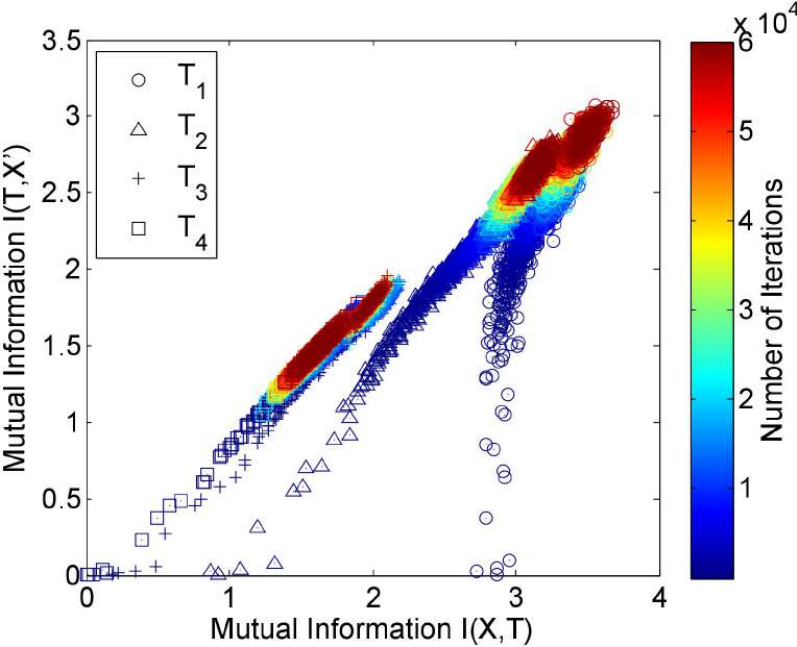}} \hspace{0.00\textwidth}
\subfigure[IP-I (decoder part) when $K=2$] {\includegraphics[width=.23\textwidth]{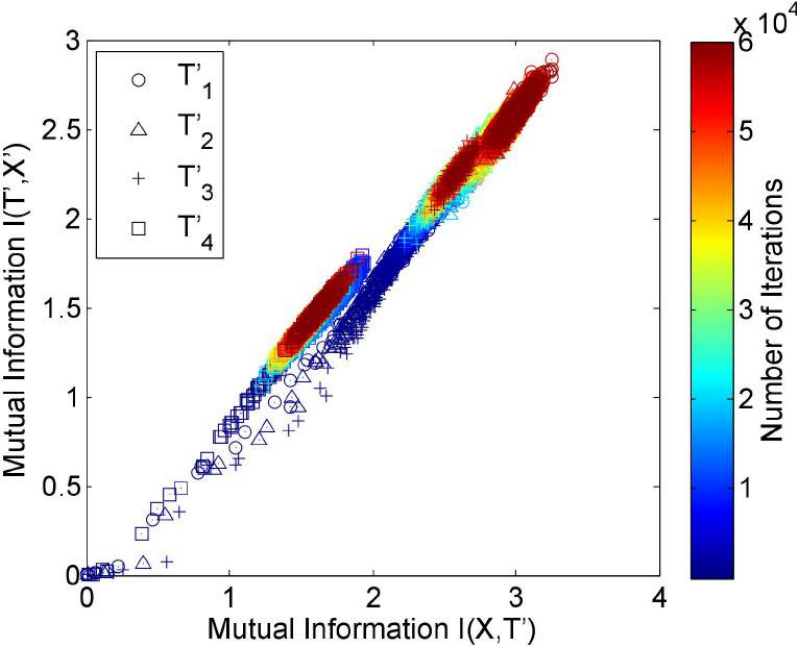}}
\subfigure[IP-I (encoder part) when $K=14$] {\includegraphics[width=.23\textwidth]{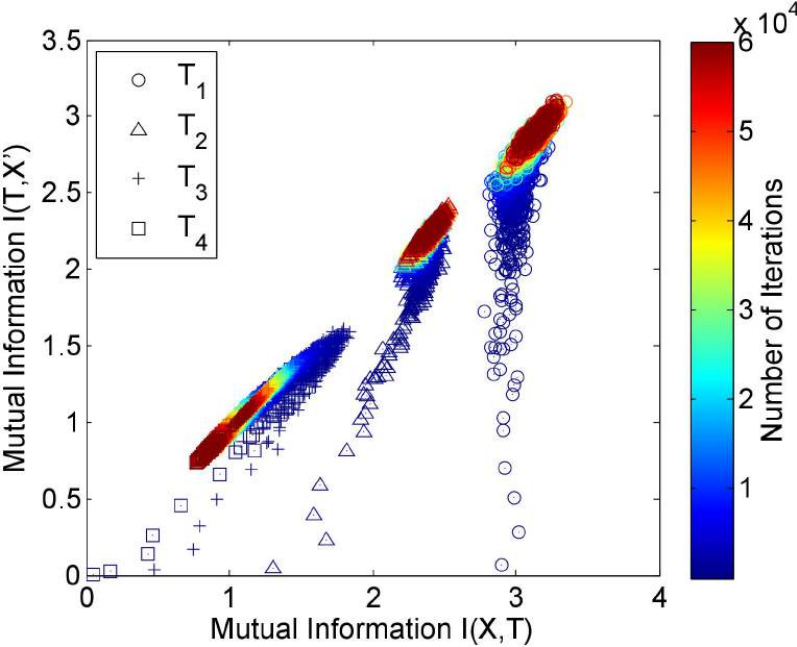}} \hspace{0.00\textwidth}
\subfigure[IP-I (decoder part) when $K=14$] {\includegraphics[width=.23\textwidth]{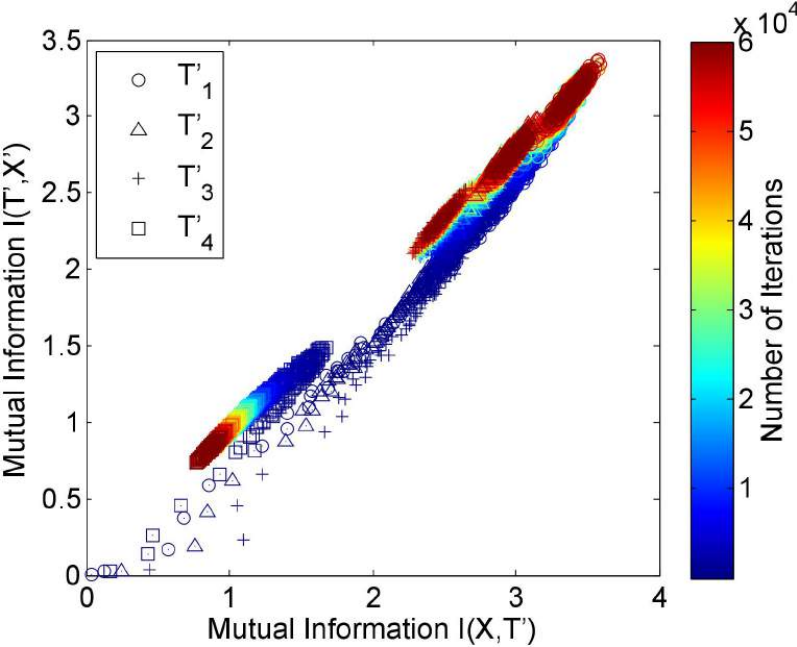}} & \\
\subfigure[IP-I (encoder part) when $K=17$] {\includegraphics[width=.23\textwidth]{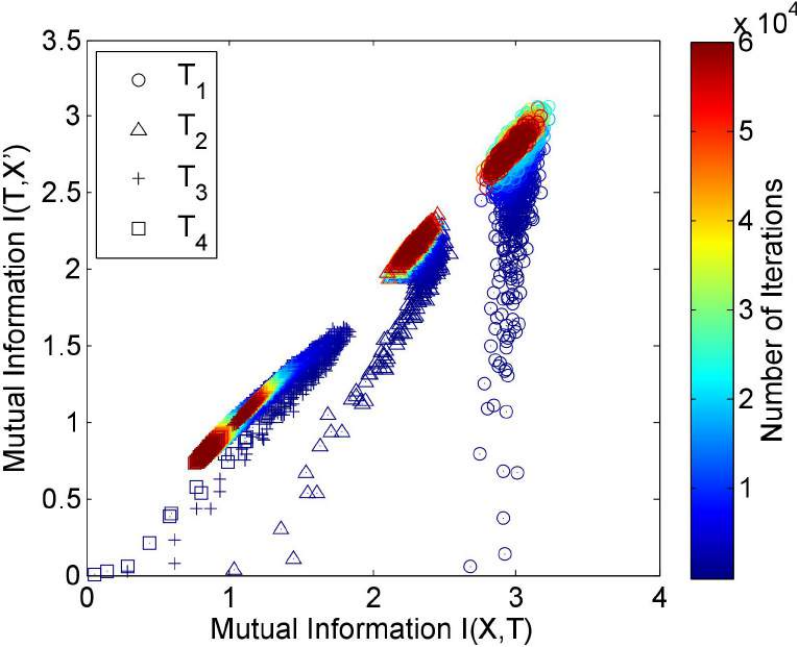}} \hspace{0.00\textwidth}
\subfigure[IP-I (decoder part) when $K=17$] {\includegraphics[width=.23\textwidth]{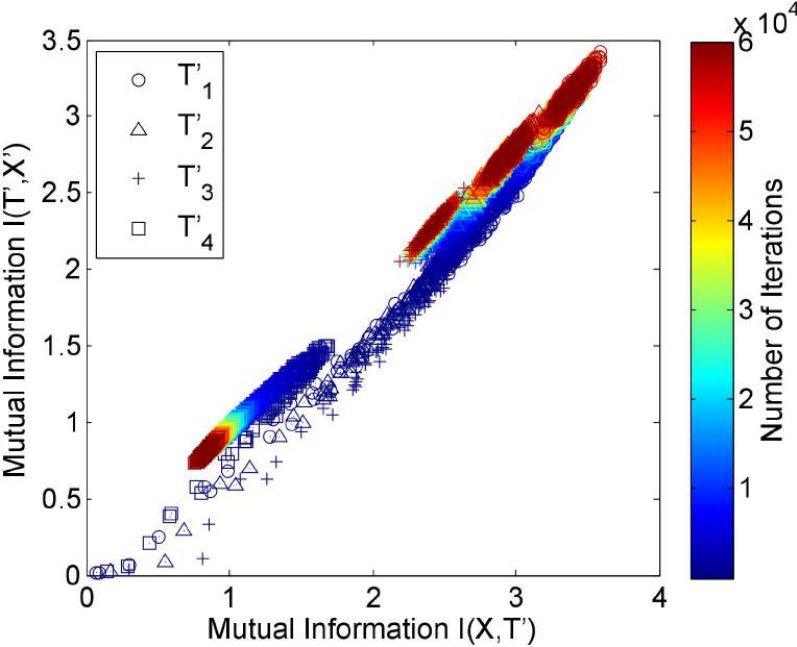}}
\subfigure[IP-I (encoder part) when $K=36$] {\includegraphics[width=.23\textwidth]{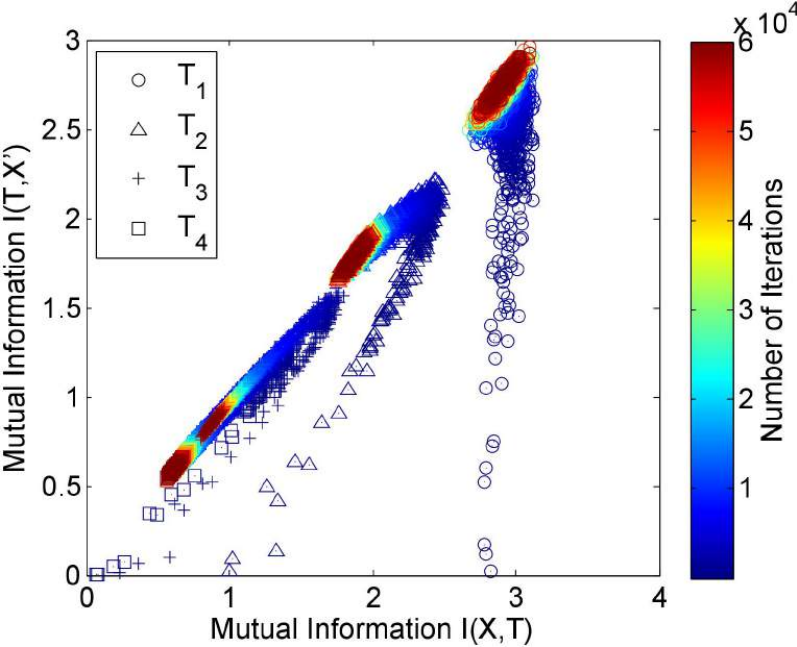}} \hspace{0.00\textwidth}
\subfigure[IP-I (decoder part) when $K=36$] {\includegraphics[width=.23\textwidth]{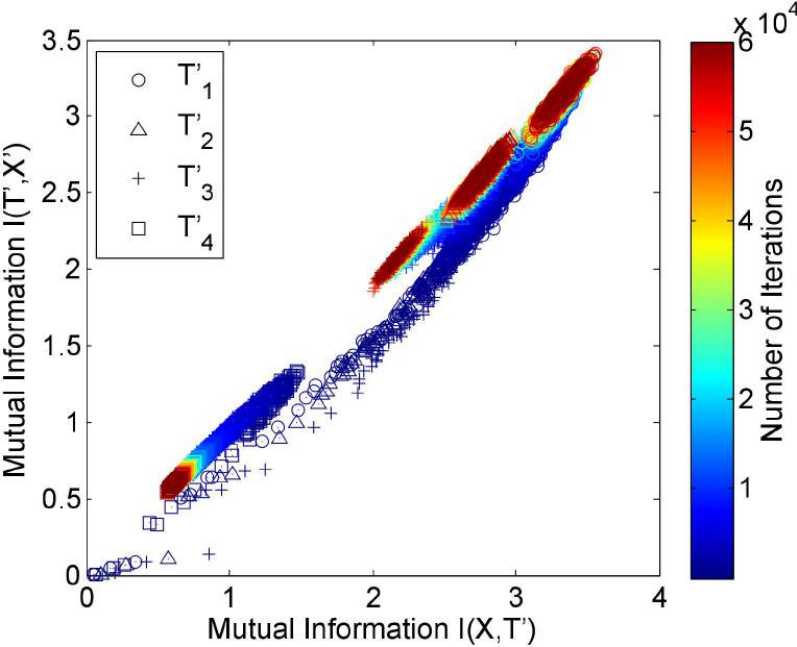}} & \\
\end{tabular}
\caption{The validation of bifurcation point associated with $K$. (a), (c), (e) and (g) demonstrate the IP-I ($T$ is in the encoder module) with $K$ equals to $2$, $14$, $17$ and $36$ respectively, whereas (b), (d), (f) and (h) demonstrate the corresponding IP-I when $T$ is in the decoder module. As can be seen, the general patterns of curves in IPs begin to have a transition between $K=14$ and $K=17$. This suggests an effective dimensionality of Fashion-MNIST dataset is approximately $15$ or $16$.\vspace{-0.0cm}}
\label{fig:bifurcation_Fashion}
\end{figure*}

The network topology of stacked autoencoders (SAEs) on Fashion-MNIST is ``784-1000-500-250-$K$-250-500-1000-784", where $K$ denotes the number of neurons in the bottleneck layer. We first test different SAEs topologies with $K$ ranging from $2$ to $36$. We expect different behaviors of the curves shown in the IPs depending on $K>D$ or $K\leq D$, where $D$ is an effective dimensionality that can fit the training data well. The corresponding IP-I is shown in Fig.~\ref{fig:bifurcation_Fashion}. It is very easy to observe the monotonically decreasing characteristics of $\mathbf{I}(X;T_i)$ and $\mathbf{I}(X';T'_i)$. Thus, the \textbf{Fundamental Property I} (i.e., $\mathbf{I}(X;T_1)\geq\mathbf{I}(X;T_2)\geq\dots\geq\mathbf{I}(X;T_S)$ and $\mathbf{I}(X';T'_1)\geq\mathbf{I}(X';T'_2)\geq\dots\geq\mathbf{I}(X';T'_S)$) is always established.

If we look deeper, the IP-I (encoder part) is more sensitive to the change of $K$, thus providing a good indicator to investigate the data dimensionality property. The experimental results validate the \textbf{Fundamental Property III} very well, because the curves associated with $T_1$ and $T_2$ begin to approaching the bisector after a certain point when $K\geq17$, rather than deviating the bisector as demonstrated when $K\leq14$. Moreover, it is worth noting that our estimated $D$ matches well the values of intrinsic dimensionality given by benchmarking estimators. In fact, the intrinsic dimensionality estimated by the Maximum Likelihood Estimation (MLE) \cite{levina2005maximum}, the Minimum Neighbor Distance (MiND) Estimator \cite{lombardi2011minimum} and the Dimensionality from Angle and Norm Concentration (DANCo) \cite{ceruti2014danco} are $13.47$, $15$ and $16$, respectively.

\subsection{Validation of Fundamental Property II}
\begin{figure*}[!htbp]
\centering
\begin{tabular}{ccc}
\subfigure[] {\includegraphics[width=.3\textwidth]{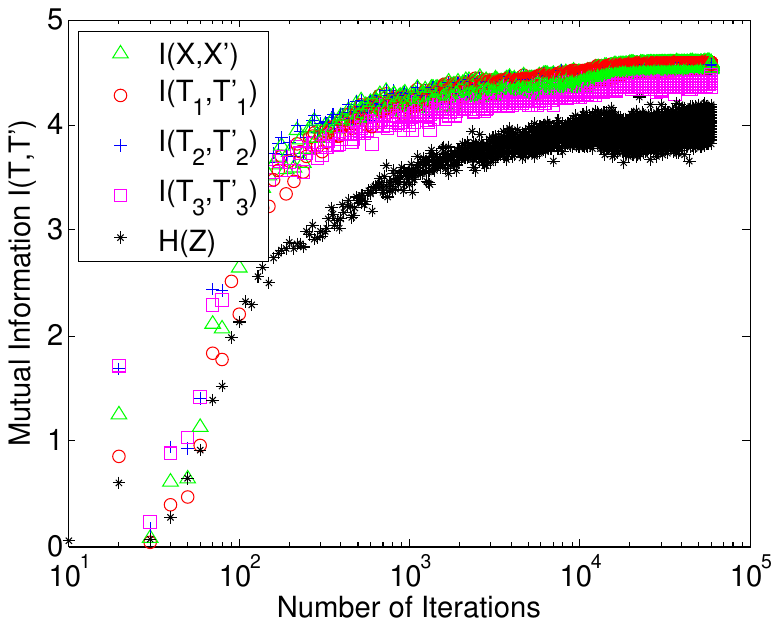}}
\subfigure[] {\includegraphics[width=.3\textwidth]{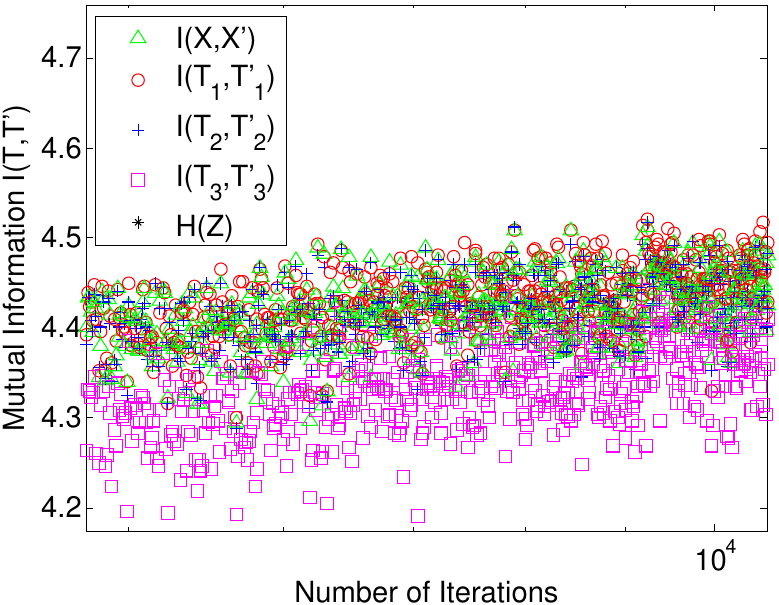}}
\subfigure[] {\includegraphics[width=.3\textwidth]{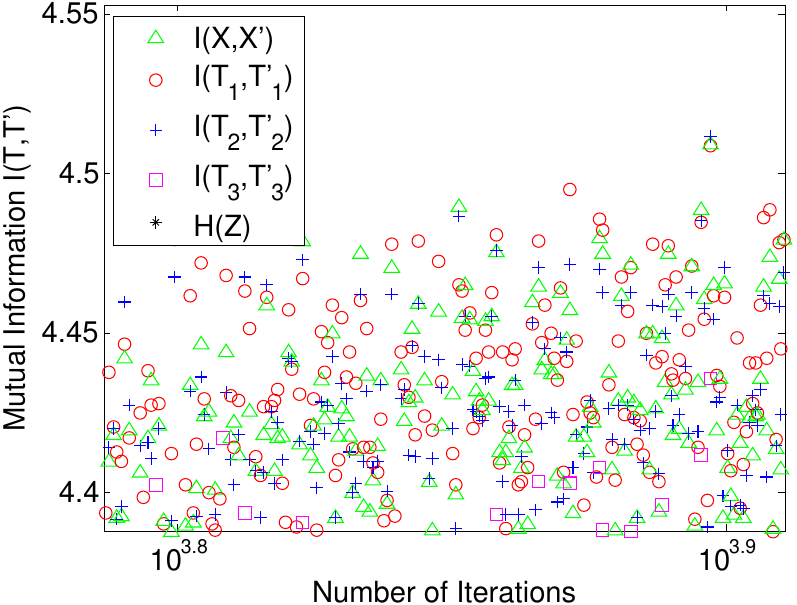}} & \\
\subfigure[] {\includegraphics[width=.3\textwidth]{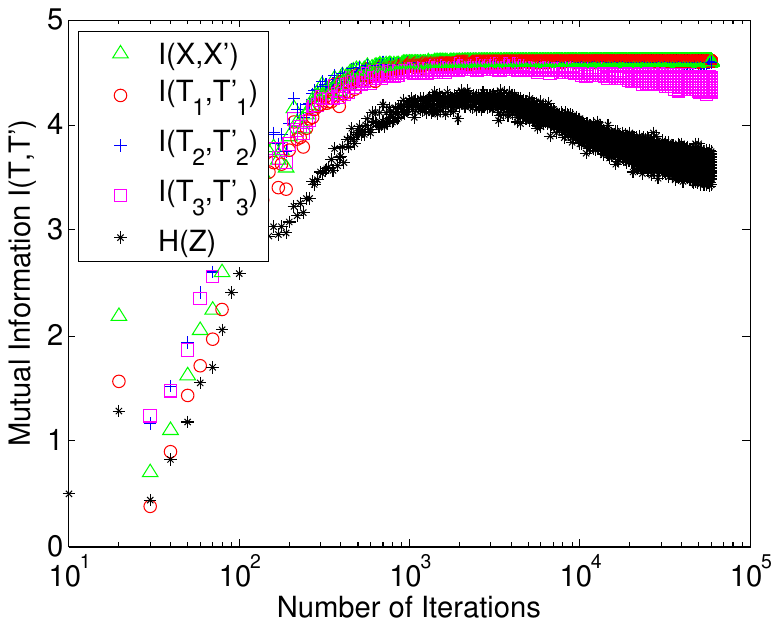}}
\subfigure[] {\includegraphics[width=.3\textwidth]{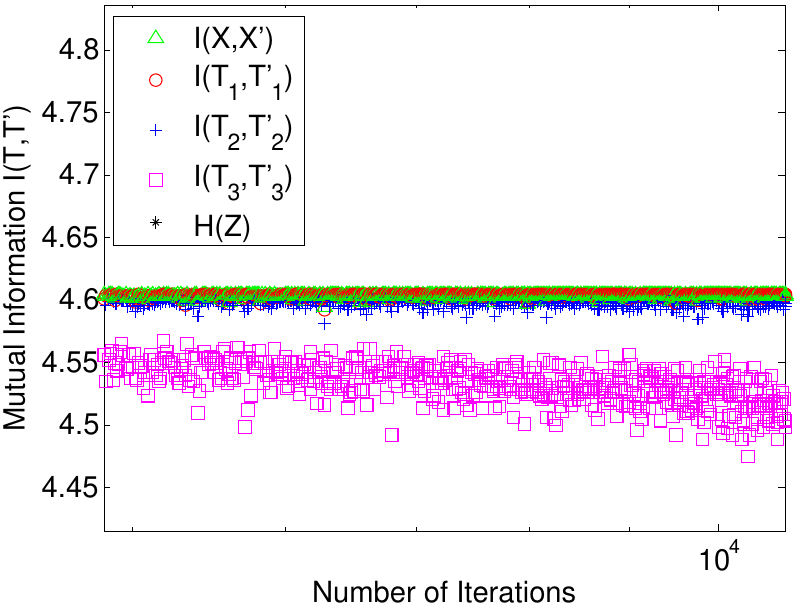}}
\subfigure[] {\includegraphics[width=.3\textwidth]{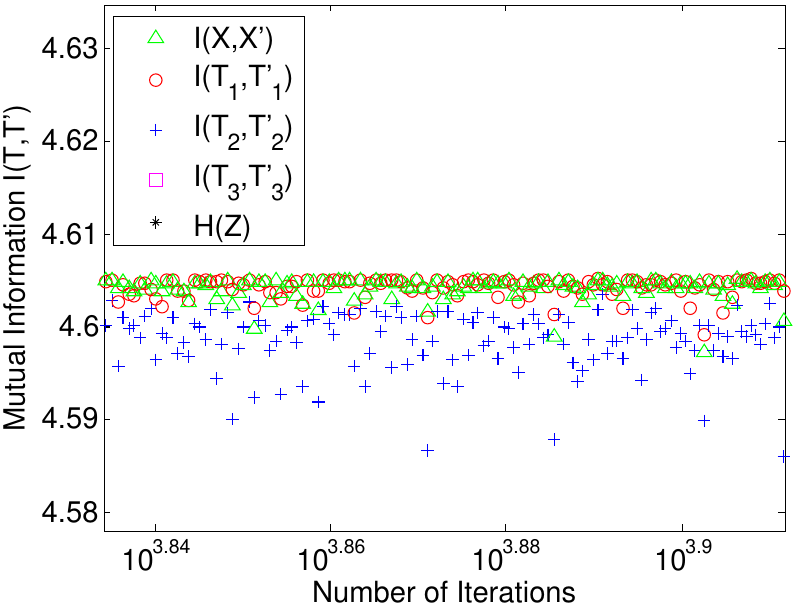}} & \\
\end{tabular}
\caption{The validation of data processing inequality (DPI) associated with the layer-wise mutual information. (a) demonstrates the layer-wise mutual information when $K=2$; (b) shows a zoom-in results of (a) in the iterations between (approximately) $6\times10^3$ to $8\times10^3$; (c) shows a further zoom-in results of (b). Similarly, (d) demonstrates the layer-wise mutual information when $K=36$; (e) shows a zoom-in results of (d) in the iterations between (approximately) $6\times10^3$ to $8\times10^3$; (f) shows a further zoom-in results of (e). In all sub-figures, the green triangles denote the mutual information between $X$ and $X'$, the red circles denote the mutual information between $T_1$ and $T'_1$, the blue plus signs denote the mutual information between $T_2$ and $T'_2$, the magenta squares denote the mutual information between $T_3$ and $T'_3$, and the black asterisks denote the mutual information between $T_4$ and $T'_4$ (reduces to the entropy of $T_4$ in our case).\vspace{-0.0cm}}
\label{fig:DPI_validation_Fashion}
\end{figure*}

Similar to the validation procedure on MNIST, we select two specific values of $K$ to validate the second type of DPI in the \textbf{Fundamental Property II}. Fig. \ref{fig:DPI_validation_Fashion} demonstrates the layer-wise mutual information corresponding to $K=2$ and $K=36$ respectively. It is obviously that, the DPI associated with mutual information (i.e., $\mathbf{I}(X;X')\geq \mathbf{I}(T_1;T'_1)\geq\dots\geq \mathbf{I}(T_S;T'_S)=\mathbf{H}(Z)$) is established no matter the value of $K$.

\section{Property validation on FERG-DB} \label{appendix_B}
\subsection{Validation of Fundamental Properties I, II $\&$ III}
\begin{figure*}[!htbp]
\centering
\begin{tabular}{ccc}
\subfigure[IP-I (encoder part) when $K=2$] {\includegraphics[width=.23\textwidth]{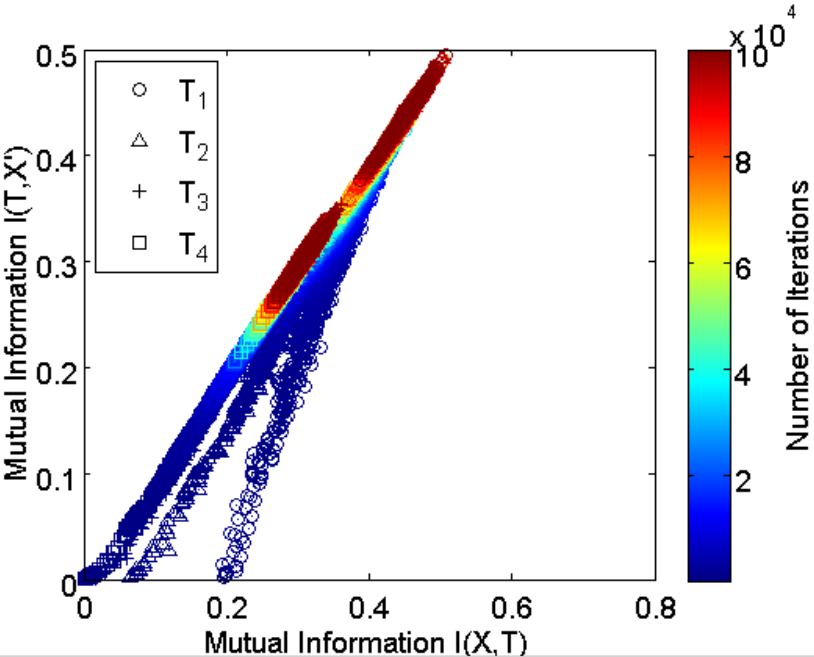}}
\subfigure[DPI when $K=2$] {\includegraphics[width=.23\textwidth]{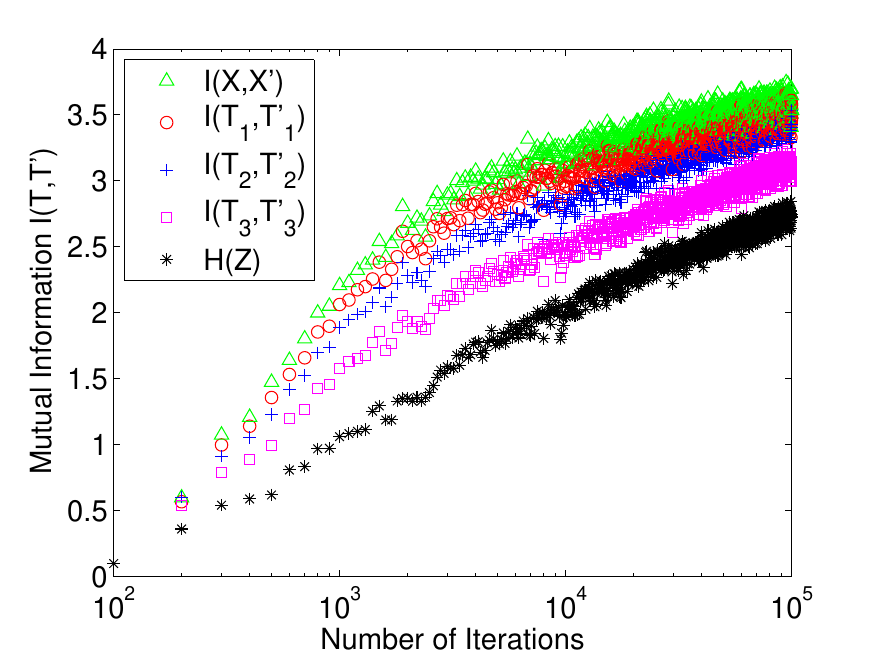}}
\subfigure[IP-I (encoder part) when $K=16$] {\includegraphics[width=.23\textwidth]{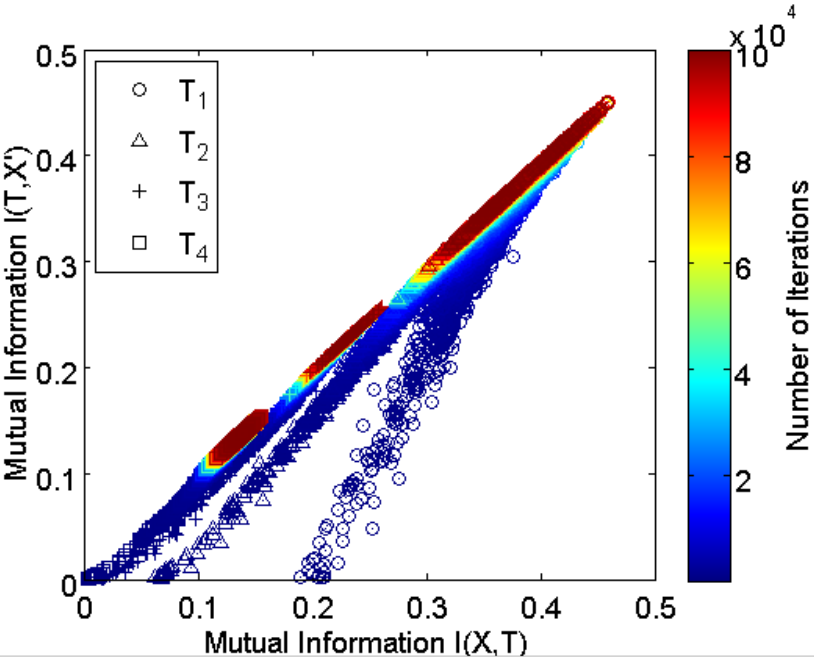}}
\subfigure[DPI when $K=16$] {\includegraphics[width=.23\textwidth]{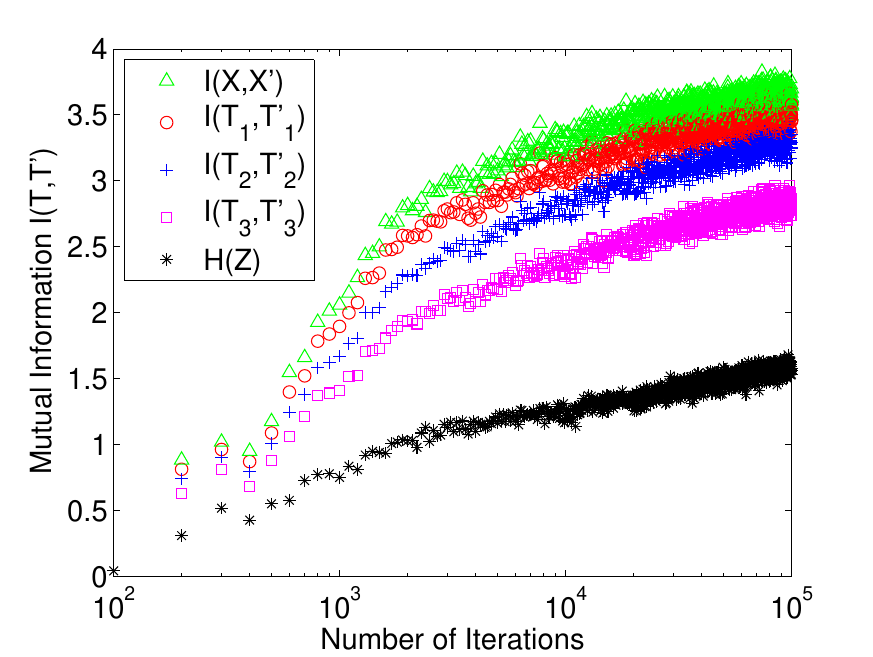}} & \\
\subfigure[IP-I (encoder part) when $K=36$] {\includegraphics[width=.23\textwidth]{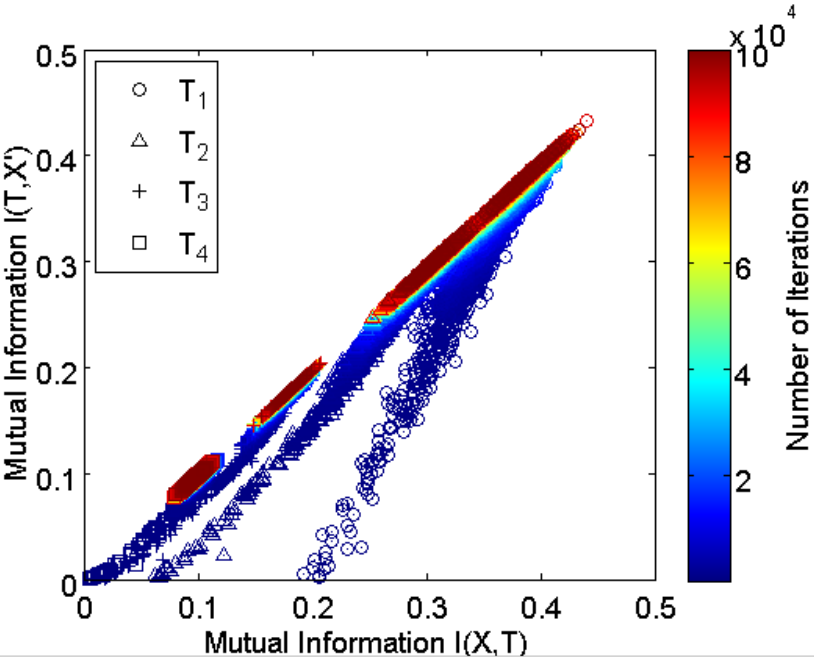}}
\subfigure[DPI when $K=36$] {\includegraphics[width=.23\textwidth]{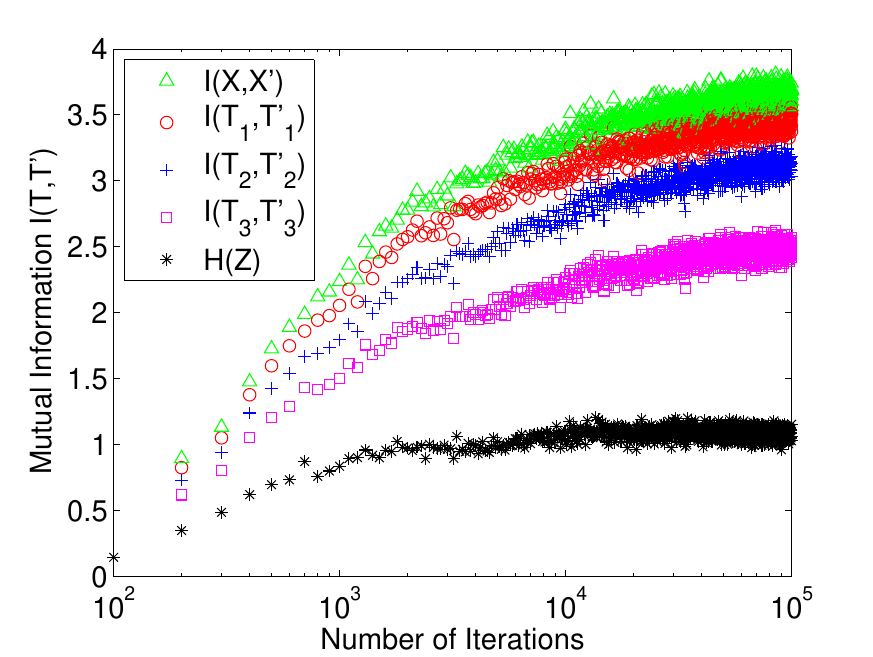}}
\subfigure[IP-I (encoder part) when $K=81$] {\includegraphics[width=.23\textwidth]{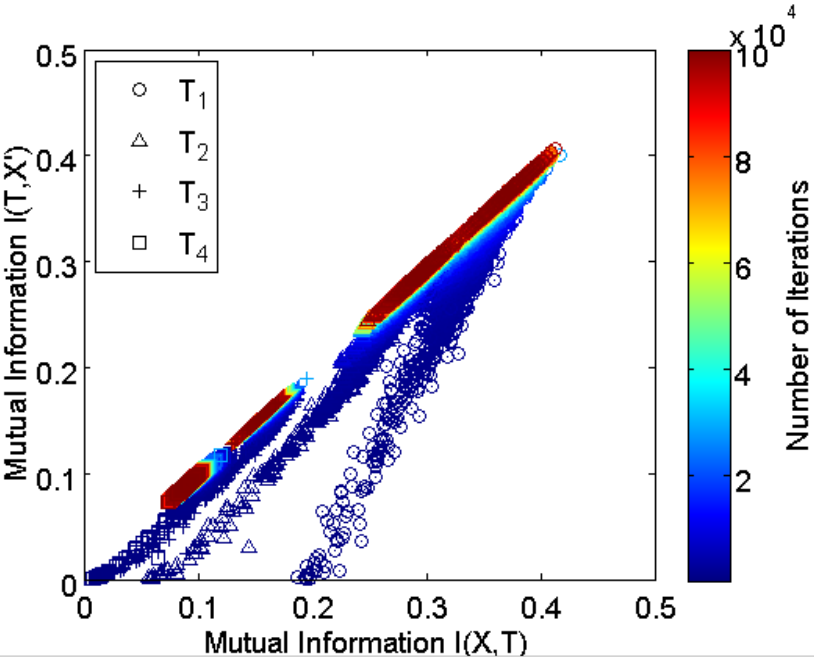}}
\subfigure[DPI when $K=81$] {\includegraphics[width=.23\textwidth]{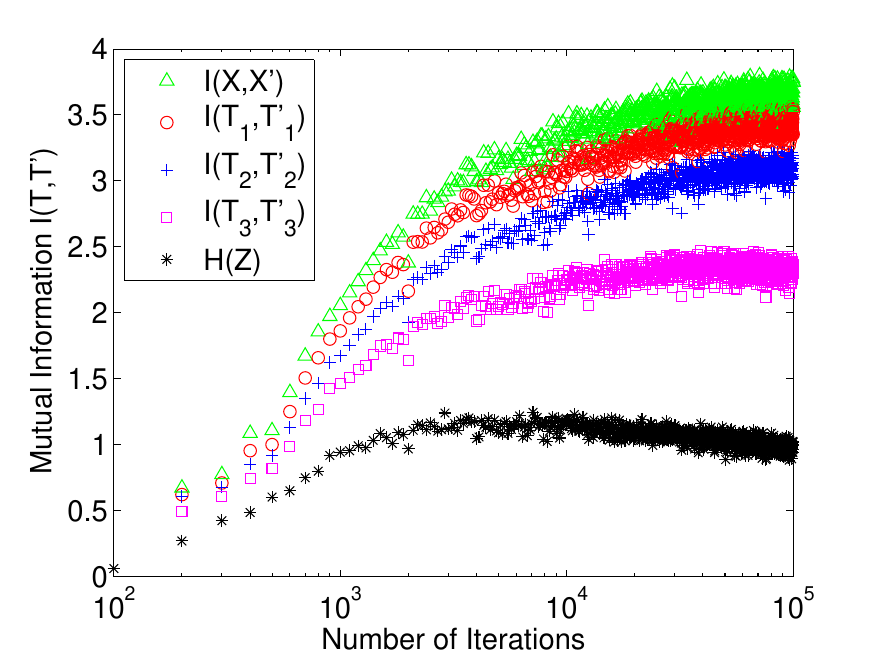}} & \\
\end{tabular}
\caption{The information plane and corresponding pair-wise mutual information with different values of $K$. (a), (c), (e) and (g) demonstrate the IP-I ($T$ is in the encoder module) with $K$ equals to $2$, $16$, $36$ and $81$ respectively, whereas (b), (d), (f) and (h) demonstrate the corresponding layer-wise mutual information to illustrate the DPI.\vspace{-0.0cm}}
\label{fig:FERG}
\end{figure*}

The network topology of SAEs on FERG-DB is selected to be ``1024-512-256-100-$K$-100-256-512-1024". We test the DPIs and the existence of bifurcation point together. We vary the value of $K$ from $2$ to $81$. Figs. \ref{fig:FERG}(a), \ref{fig:FERG}(c), \ref{fig:FERG}(e) and \ref{fig:FERG}(g) demonstrate the IP-I when $K$ equals to $2$, $16$, $36$ and $81$ respectively, whereas Figs. \ref{fig:FERG}(b), \ref{fig:FERG}(d), \ref{fig:FERG}(f) and \ref{fig:FERG}(h) demonstrate the layer-wise mutual information.

As can be seen, the two types of DPI is established no matter the value of $K$, i.e., $\mathbf{I}(X;T_1)\geq\mathbf{I}(X;T_2)\geq\dots\geq\mathbf{I}(X;T_S)$, $\mathbf{I}(X';T'_1)\geq\mathbf{I}(X';T'_2)\geq\dots\geq\mathbf{I}(X';T'_S)$ and $\mathbf{I}(X;X')\geq \mathbf{I}(T_1;T'_1)\geq\dots\geq \mathbf{I}(T_S;T'_S)=\mathbf{H}(Z)$). However, there is no distinct patterns in the IP-I, i.e., all the curves approaching the bisector monotonically for the values tried. This is different from the scenario in MNIST or Fashion-MNIST when $K>D$, in which all curves start to increase up to a point and then go back to approaching bisection. It is also different from the scenario when $K\leq D$, in which all the curves go away from bisection. One possible reason is that the effective dimensionality for FERG-DB is very small such that $K=2$ can already fit the data very well. Therefore, we expect the effective dimensionality to be $1$. Because the size of bottleneck layer is always larger than $1$, the curves in IPs should demonstrate the same pattern.

To support our argument, we estimate $D$ using the aforementioned benchmarking estimators. It is surprising to find that the intrinsic dimensionality estimated by the Maximum Likelihood Estimation (MLE) \cite{levina2005maximum}, the Minimum Neighbor Distance (MiND) Estimator \cite{lombardi2011minimum} and the Dimensionality from Angle and Norm Concentration (DANCo) \cite{ceruti2014danco} are $0$, $1$ and $1$, respectively. These results corroborate our argument, i.e., since the bifurcation point in IPs for FERG-DB is just $1$, all the curves for $K>1$ should demonstrate the same pattern no matter the value of $K$. Recall that this dataset is a representation of a face from a single cartoon character (or template), which can be sketched using just one line with different shapes, therefore the result is acceptable.

\section{Robustness analysis on kernel size $\sigma$} \label{appendix_C}
\begin{figure*}[!htbp]
\centering
\begin{tabular}{ccc}
\subfigure[IP ($h=0.5$, $K=2$)] {\includegraphics[width=.23\textwidth]{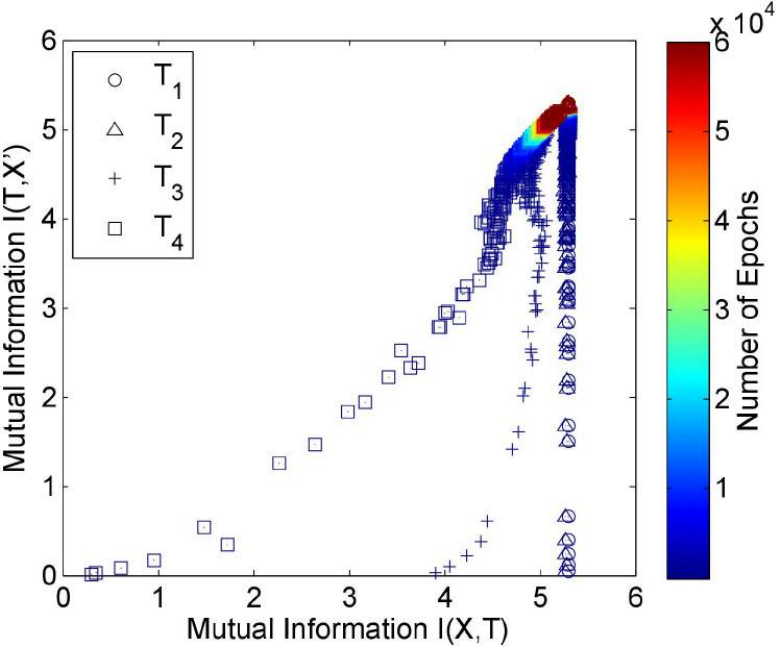}}
\subfigure[IP ($h=0.5$, $K=36$)] {\includegraphics[width=.23\textwidth]{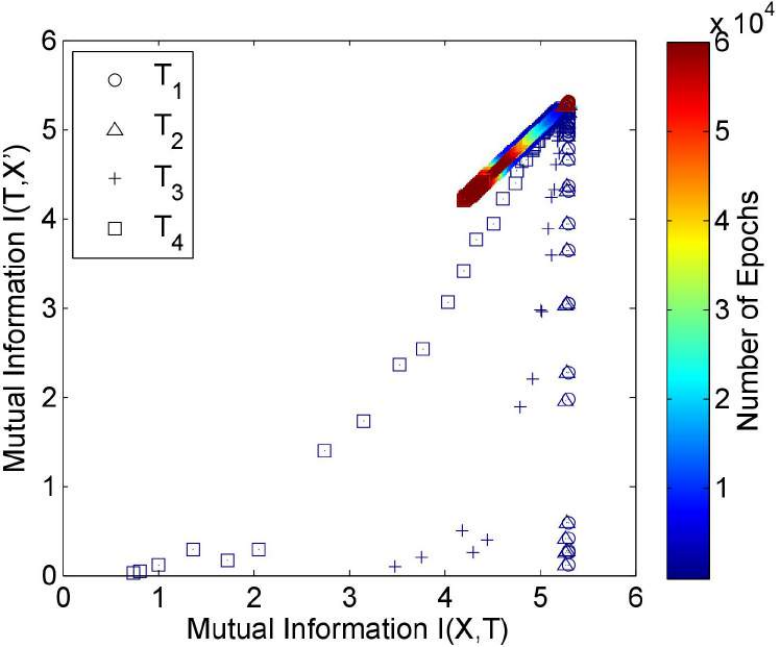}}
\subfigure[IP ($h=10$, $K=2$)] {\includegraphics[width=.23\textwidth]{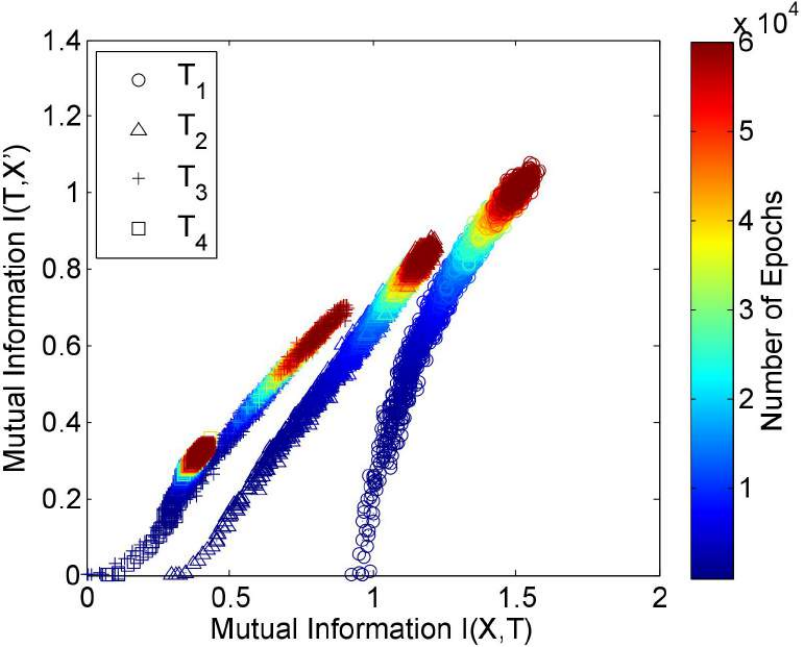}}
\subfigure[IP ($h=10$, $K=36$)] {\includegraphics[width=.23\textwidth]{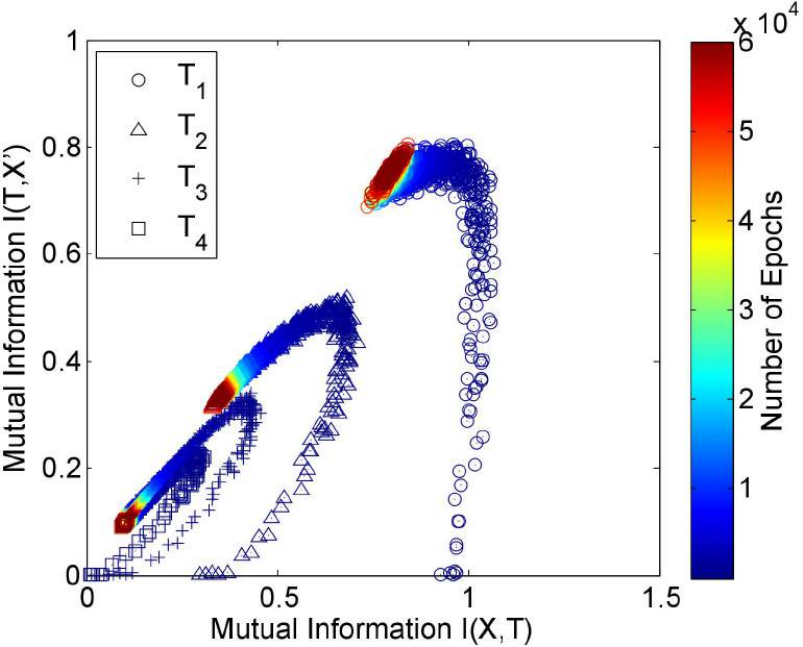}} & \\
\end{tabular}
\caption{The robustness analysis on the selection of parameter $h$ (hence $\sigma$). (a) and (b) demonstrate, when $h=1$, the IPs for $K=2$ and $K=36$ respectively. (c) and (d) demonstrate, when $h=11$, the IPs for $K=2$ and $K=36$ respectively. The SAE topology is fixed to ``784-1000-500-250-$K$-250-500-1000-784", where ``$K$" denotes the bottleneck layer size. The distinctions of curves in the IP remain the same if $h$ is in a large yet reasonable range. However, it should be acknowledged that if $h$ is too small (less than $0.1$), all the estimation values are saturated and approaching to their limit. On the other hand, if $h$ is large enough (more than $30$), it is hard to observe the distinctions due to the large estimation variance among different epoches.\vspace{-0.0cm}}
\label{fig:robust_silverman}
\end{figure*}

As emphasized in the main text, care must be taken to select an appropriate value for kernel size $\sigma=h\times n^{-1/(4+d)}$. This paper selects $h=6$. We show, in Fig.~\ref{fig:robust_silverman}, that even though $h$ (hence $\sigma$) is not optimized, we can still observe the same trends of general patterns of the curves in the IP that is controlled by the bottleneck size.

\begin{figure*}[!htbp]
\centering
\begin{tabular}{ccc}
\subfigure[epoch 1 ($a=-30$, $e=20$)] {\includegraphics[width=.3\textwidth]{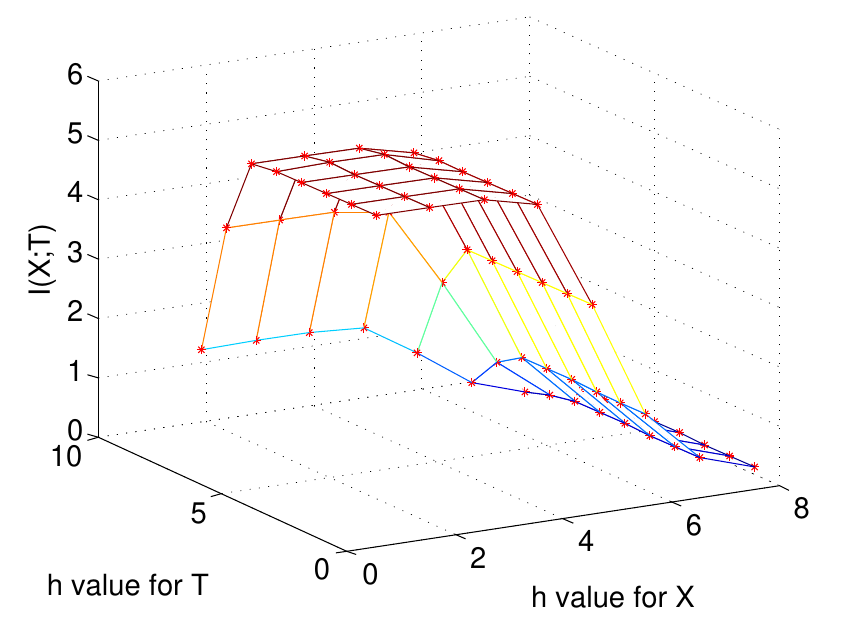}}
\subfigure[epoch 1 ($a=30$, $e=20$)] {\includegraphics[width=.3\textwidth]{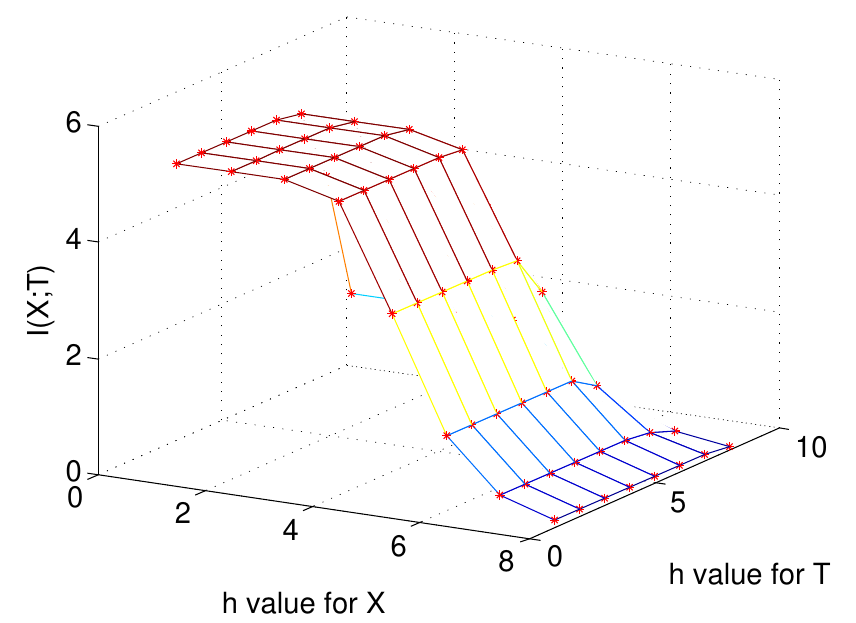}}
\subfigure[epoch 1 ($a=150$, $e=20$)] {\includegraphics[width=.3\textwidth]{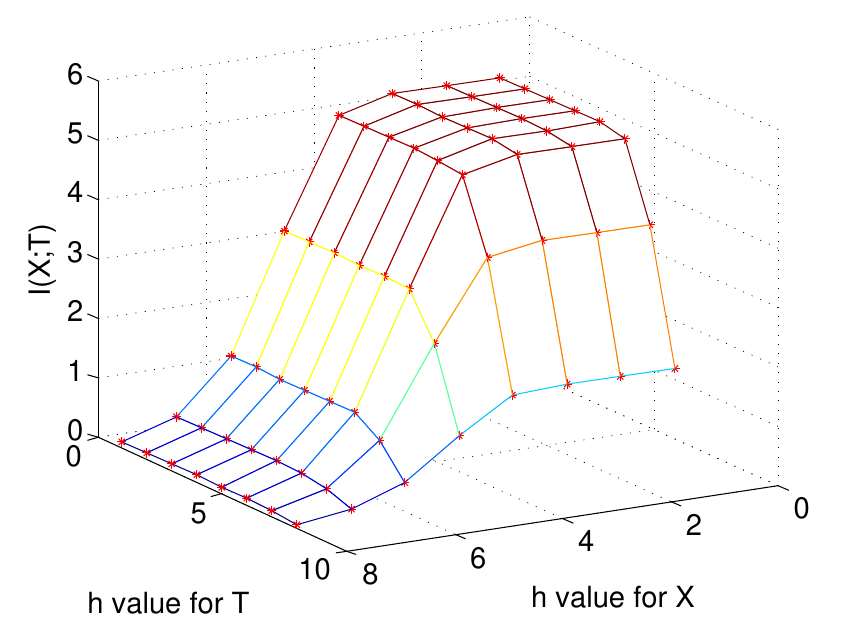}} & \\
\subfigure[epoch 30 ($a=-30$, $e=20$)] {\includegraphics[width=.3\textwidth]{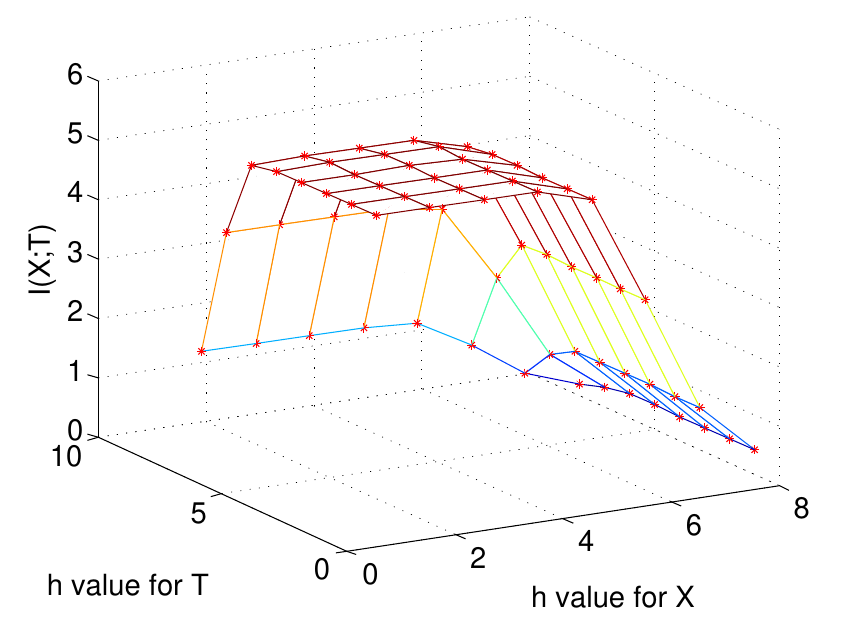}}
\subfigure[epoch 30 ($a=30$, $e=20$)] {\includegraphics[width=.3\textwidth]{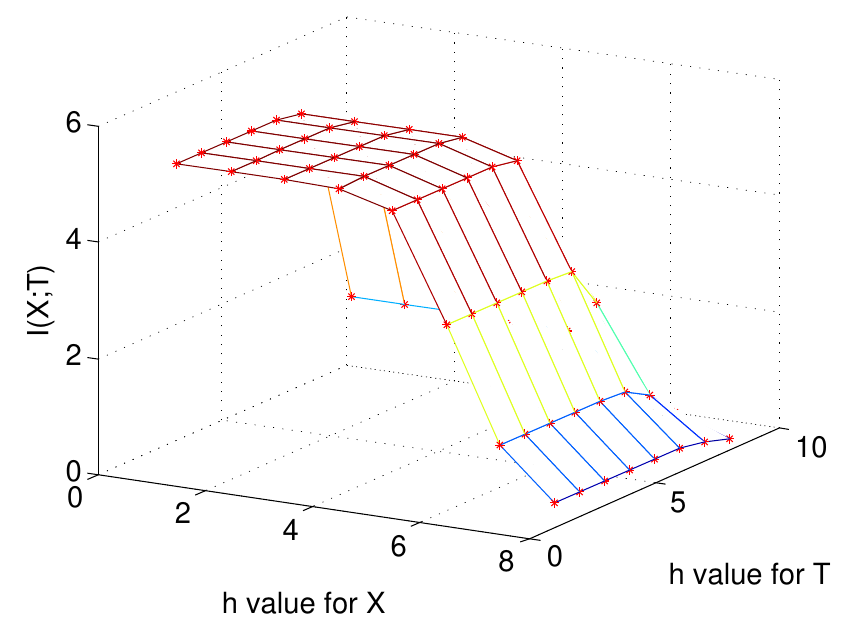}}
\subfigure[epoch 30 ($a=150$, $e=20$)] {\includegraphics[width=.3\textwidth]{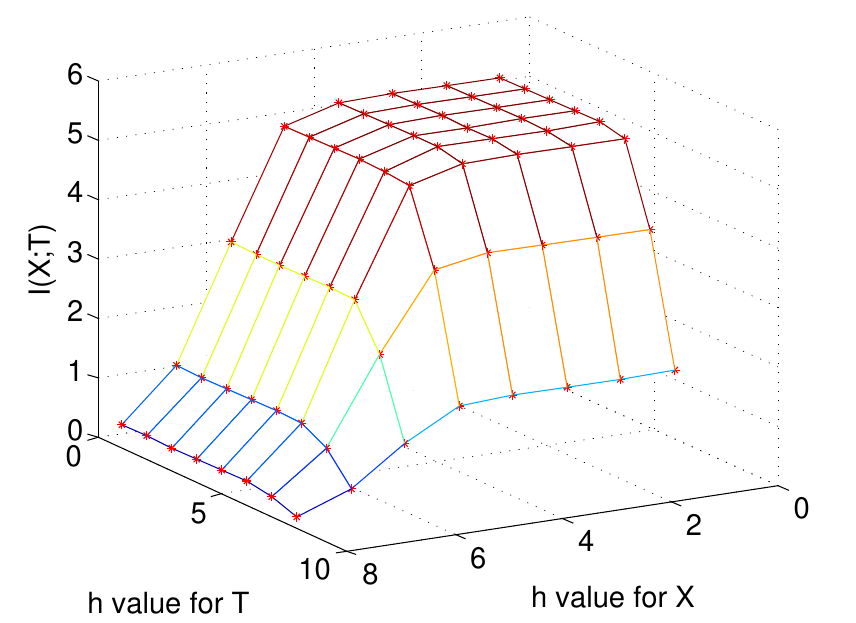}} & \\
\end{tabular}
\caption{(a)-(c) show the mutual information value $\mathbf{I}(X;T)$ (after $1$ epoch training) with respect to different kernel sizes in input layer and bottleneck layer at view $(a,e)=(-30,20)$, $(a,e)=(30,20)$, and $(a,e)=(150,20)$ respectively, where $a$ denotes azimuth and $e$ denotes elevation. (d)-(f) show the mutual information value $\mathbf{I}(X;T)$ (after $30$ epochs training) with respect to different kernel sizes in input layer and bottleneck layer at view $(a,e)=(-30,20)$, $(a,e)=(30,20)$, and $(a,e)=(150,20)$ respectively.\vspace{-0.5cm}}
\label{fig:robust_silverman_MI}
\end{figure*}

We also show that the value of mutual information between two variables is monotonically decreasing if the kernel size in any one of the variables increases. To this end, we train a basic autoencoder with topology ``$784$-$36$-$784$" on MNIST dataset with $30$ epochs, which, as has been observed, is sufficient to reliably converge. We estimate the mutual information $\mathbf{I}(X;T)$ with respect to $8$ different $h$ values ($0.1\times2^k$, $k=0,1,\cdots,7$) in both input layer and bottleneck layer. Fig.~\ref{fig:robust_silverman_MI} demonstrates the value of $\mathbf{I}(X;T)$ at the end of epoch $1$ and epoch $30$. As can be seen, $\mathbf{I}(X;T)$ is monotonically decreasing as any one of the $h$ increases. This suggests that the same trends for entropy also apply for mutual information.

\end{document}